\documentclass[10pt,twocolumn,letterpaper]{article}

\usepackage[pagenumbers]{wacv} % To force page numbers, e.g. for an arXiv version

\usepackage{graphicx}
\usepackage{amsmath}
\usepackage{amssymb}
\usepackage{booktabs}
\usepackage{algorithm}
\newtheorem{theorem}{Theorem}
\newtheorem{corollary}[theorem]{Corollary}
\newtheorem{lemma}[theorem]{Lemma}
\usepackage{tikz}
\usepackage{multirow}
\usepackage{subcaption}

\usepackage{xcolor}         % colors
\definecolor{mintbg}{rgb}{.63,.79,.95}
\definecolor{apricot}{rgb}{0.98, 0.81, 0.69}
\definecolor{asparagus}{rgb}{0.53, 0.66, 0.42}
\definecolor{bondiblue}{rgb}{0.0, 0.58, 0.71}
\colorlet{lightmintbg}{bondiblue!50}
\colorlet{lightapricot}{apricot!100}

\newcommand{\norm}[1]{\left\lVert#1\right\rVert}
\providecommand{\E}{{\mathbb E}}
\usepackage[pagebackref,breaklinks,colorlinks]{hyperref}
\usepackage[accsupp]{axessibility} % Improves PDF readability for those with disabilities.

\usepackage[capitalize]{cleveref}
\crefname{section}{Sec.}{Secs.}
\Crefname{section}{Section}{Sections}
\Crefname{table}{Table}{Tables}
\crefname{table}{Tab.}{Tabs.}

\def\wacvPaperID{2101} % *** Enter the WACV Paper ID here
\def\confName{WACV}
\def\confYear{2025}

\begin{document}
\title{SADDLe: Sharpness-Aware Decentralized Deep Learning with Heterogeneous Data}

\author{Sakshi Choudhary\\
Purdue University\\
West Lafayette, IN\\
{\tt\small choudh23@purdue.edu}
\and
Sai Aparna Aketi\\
Purdue University\\
West Lafayette, IN\\
{\tt\small aketiaparna@gmail.com}
\and
Kaushik Roy\\
Purdue University\\
West Lafayette, IN\\
{\tt\small kaushik@purdue.edu}
}

\maketitle

%%%%%%%%% ABSTRACT
\begin{abstract}
Decentralized training enables learning with distributed datasets generated at different locations without relying on a central server. In realistic scenarios, the data distribution across these sparsely connected learning agents can be significantly heterogeneous, leading to local model over-fitting and poor global model generalization. Another challenge is the high communication cost of training models in such a peer-to-peer fashion without any central coordination. In this paper, we jointly tackle these two-fold practical challenges by proposing SADDLe, a set of sharpness-aware decentralized deep learning algorithms.
SADDLe leverages Sharpness-Aware Minimization (SAM) to seek a flatter loss landscape during training, resulting in better model generalization as well as enhanced robustness to communication compression. We present two versions of our approach and demonstrate its effectiveness through extensive experiments on various Computer Vision datasets (CIFAR-10, CIFAR-100, Imagenette, and ImageNet), model architectures, and graph topologies. Our results show that SADDLe leads to 1-20\% improvement in test accuracy as compared to existing techniques while incurring a minimal accuracy drop ($\sim 1\%$) in the presence of up to 4$\times$ compression.
%along with an average drop of only 1\% in the presence of up to 4$\times$ compression.
%(with baselines incurring a drop of 4.3\% on average).
%conduct extensive experiments on various Computer Vision datasets (CIFAR-10, CIFAR-100, Imagenette, and ImageNet) to show that SADDLe leads to 1-20\% improvement in test accuracy compared to other existing techniques. Additionally, our proposed approach is robust to communication compression, with an average drop of only 1\% in the presence of up to 4$\times$ compression.
\end{abstract}

%%%%%%%%% BODY TEXT
\section{Introduction}
\label{sec:intro}

Federated learning enables training with distributed data across multiple agents under the orchestration of a central server \cite{federated}. 
%The server collects local model updates, processes them, and sends the global updates back to the agents. 
However, the presence of such a central entity can lead to a single point of failure and network bandwidth issues \cite{sgp}. To address these concerns, several decentralized learning algorithms have been proposed \cite{dpsgd,sgp,ngm,cga,qgm,gut,gt}.
Decentralized learning is a peer-to-peer learning paradigm in which agents connected in a fixed graph topology learn by communicating with their peers/neighbors without the need for a central server. 
Decentralized learning algorithms have been shown to perform comparably to centralized algorithms on image classification and natural language processing (NLP) tasks \cite{dpsgd}. 
The authors in \cite{dpsgd} present Decentralized Parallel Stochastic Gradient Descent (DPSGD), which combines SGD with gossip averaging algorithm \cite{gossip} and show that the convergence rate of DPSGD is similar to its centralized counterpart \cite{dean}. Decentralized Momentum Stochastic Gradient Descent \cite{dpsgdm} introduced momentum to DPSGD, while Stochastic Gradient Push (SGP) \cite{sgp} extends DPSGD to directed and time-varying graphs. 
%There are two major challenges associated with decentralized learning: data heterogeneity and high communication cost.
 %Second paragraph: Challenges in decentralized learning
    %data is generally non-IID
    %communication compression techniques like choco sgd and deep squeeze work well with IID data only, leading to further performance degradation due to heterogeneity

 The above-mentioned algorithms assume the data to be independently and identically distributed (IID) across the agents. This refers to a scenario where the training data is distributed uniformly and randomly.
 However, in real-world applications, the data distributions can be remarkably different, i.e. non-IID or heterogeneous \cite{niid}. 
 While several algorithms have been proposed to mitigate the impact of such data heterogeneity \cite{qgm,ngm,cga,gut,gt,d2}, these algorithms do not explicitly focus on the aspect of communication cost, which may account for about 70\% of energy consumption \cite{carbonfootprint, dadaquant}.
In decentralized learning, agents communicate the models with their neighbors after every mini-batch update, leading to high communication costs. Various communication compression techniques have been proposed to address this, but these algorithms primarily focus on the settings when the data distribution is IID \cite{choco-sgd, deepsqueeze, powersgd}.
%To tackle this, several communication compression techniques have been proposed \cite{dcd, choco-sgd, deepsqueeze, powersgd}. However, these techniques primarily focus on the settings when the data distribution is IID, and fail to give any insights on achieving communication efficiency in the presence of data heterogeneity.
%fail to give any insights on the possibility of achieving communication efficiency in the presence of data heterogeneity.
\begin{figure*}[htbp!]
\centering
	\begin{subfigure}{0.32\textwidth}
		\includegraphics[width=\textwidth]{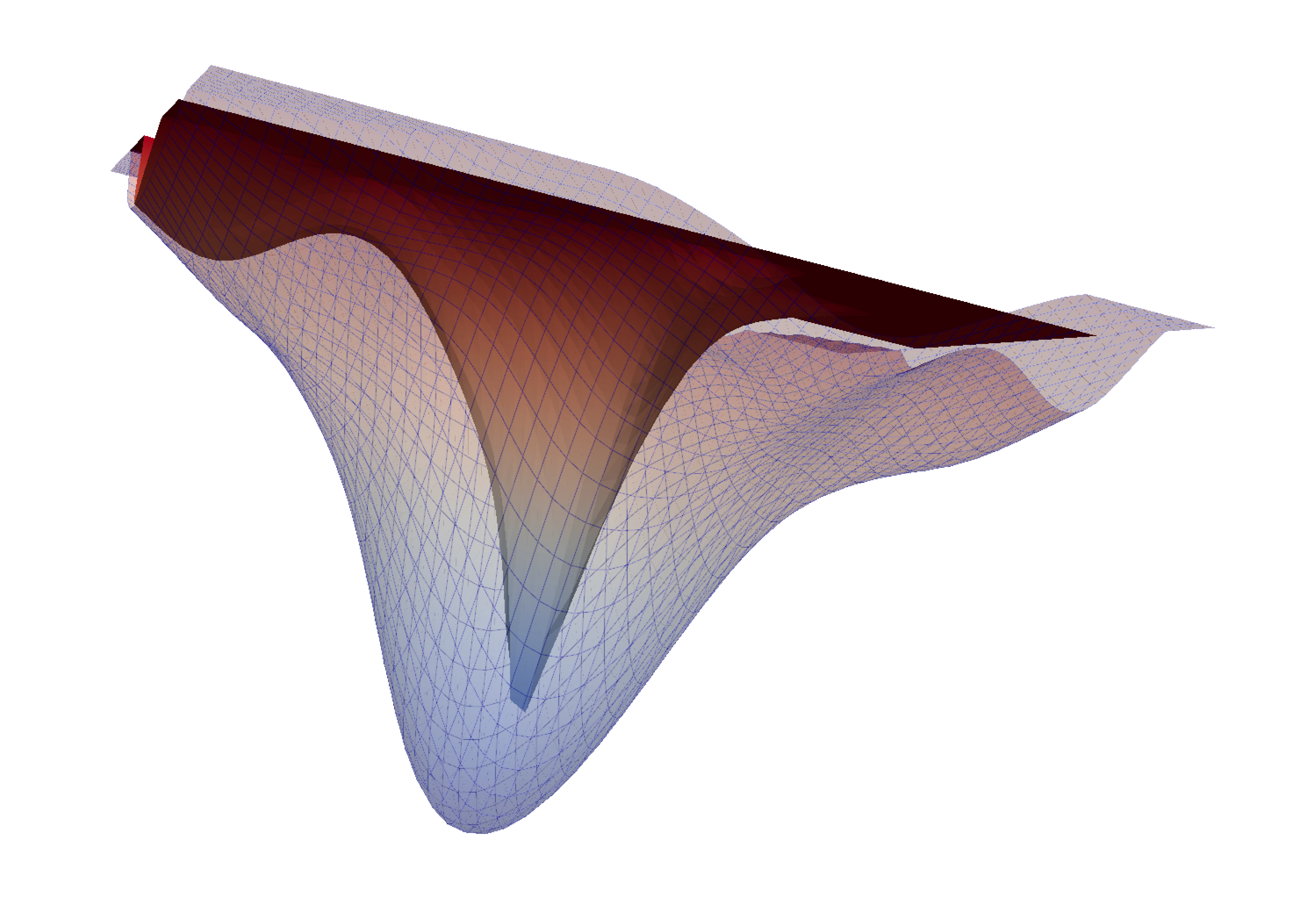}
		\caption{QGM vs Q-SADDLe}
	\end{subfigure}
	\begin{subfigure}{0.32\linewidth}
		\includegraphics[width=\textwidth]{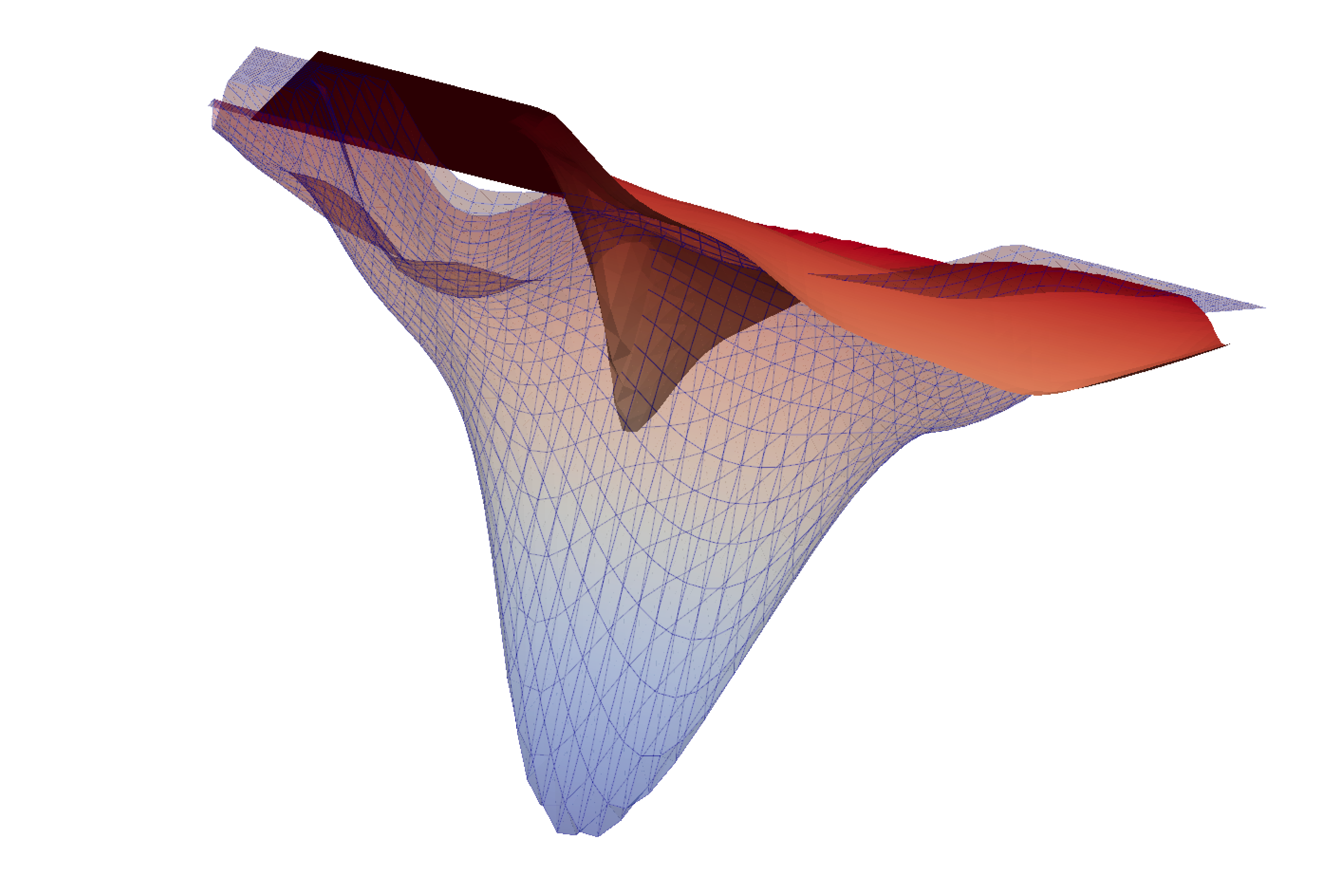}
		\caption{Comp QGM vs Comp Q-SADDLe}
	\end{subfigure}
	\caption{Loss landscape visualization for QGM (surface) vs Q-SADDLe (mesh) and Comp QGM (surface) vs Comp Q-SADDLe (mesh) for ResNet-20 trained on CIFAR-10 with non-IID data across 10 agents. Comp signifies communication compression through 8-bit stochastic quantization.}
	\label{fig:lossvisual}
\end{figure*}
%Introduce and hype up the problem more here, emphasizing how existing works fail to achieve both.
\newline In this paper, we aim to answer the following question: \textit{Can we improve the performance of decentralized learning on heterogeneous data in terms of test accuracy as well as robustness to communication compression?}
%\textit{Can we improve the performance of decentralized learning on heterogeneous data in terms of test accuracy as well as communication efficiency?} 
We put forward an orthogonal direction of enhancing the local training at each agent to positively impact the global model generalization. We propose that seeking a flatter loss landscape during training can alleviate the issue of local over-fitting, a common concern in decentralized learning scenarios with non-IID data. To achieve this, we propose SADDLe, a set of sharpness-aware decentralized deep learning algorithms. SADDLe improves generalization by simultaneously minimizing the loss value and the sharpness through gradient perturbation. This is enabled by utilizing Sharpness-Aware Minimization (SAM) \cite{sam} to seek parameters in neighborhoods with uniformly low loss values. 
%Further, in the presence of communication compression, we find that SAM leads to lower compression error, along with enhanced robustness to compression error. 
Furthermore, flatter loss landscapes are inherently more robust to perturbations in the training loss landscape \cite{sam}. Leveraging this potential, we observe that SAM leads to enhanced robustness against compression errors due to erroneous model updates resulting from communication compression. 
%Interestingly, training with the SAM optimizer at each agent also reduces compression errors, a phenomenon we attribute to the lower gradient norms produced by SAM updates (as discussed in Section \ref{sec:compsam}).
 Notably, training with SAM optimizer at each agent also reduces compression errors, a result we attribute to SAM producing lower gradient norms, which serve as an upper bound on compression errors (as discussed in Section \ref{sec:compsam}).
%Further, in the presence of communication compression, we find that SAM leads to enhanced robustness to compression error, along with lower compression error.  

To that effect, we demonstrate that SADDLe can be used in synergy with existing decentralized learning algorithms for non-IID data to attain better generalization and reduce the accuracy drop incurred due to compression. Specifically, we present two versions of our approach: Q-SADDLe, which incorporates a Quasi Global Momentum (QGM) buffer \cite{qgm}, and N-SADDLe, which utilizes cross-gradient information \cite{ngm}.
Figure \ref{fig:lossvisual} presents a visualization of the loss landscape \cite{visualizeloss} for QGM, Q-SADDLe, and their compressed counterparts. Clearly, Q-SADDLe has a much smoother loss landscape, resulting in better generalization and minimal performance loss due to communication compression.
Our detailed theoretical analysis highlights that the convergence rate of Q-SADDLe matches the well-known best result in decentralized learning \cite{dpsgd}. We also conduct extensive experiments to establish that Q-SADDLe and N-SADDLe achieve better accuracy than state-of-the-art decentralized algorithms \cite{qgm,ngm}, with a minimal accuracy drop due to communication compression. 
%on various datasets, model architectures, graph topologies, sizes and compression operators 
%like QG-DSGDm \cite{qgm} and NGM \cite{ngm} 

% Through an exhaustive set of experiments on various datasets, model architectures, graph topologies, sizes and compression operators, we show that incorporating SAM with existing decentralized learning algorithms leads to better test accuracy and better communication-efficiency with minimal performance degradation. SAM with QG-DSGDm \cite{qgm} results in an average improvement of --\% in test accuracy and aa-bb x compression with minimal performance degradation. 

%maybe this is more helpful to be written in the contribution bullet points...
%Utilizing SAM with NGM \cite{ngm} results in --\% better test accuracy on average, and it helps in compressing the second communication round by upto --x without significant drop in accuracy. In both scenarios, we observe that SAM generally leads to lower accuracy loss as compared to SGD optimizer in the presence of communication compression.

%Third paragraph: Overview of decentralized learning works for non-IID data. 1 communication round: QGM, GUT. 2 communication rounds: NGM, CGA, GT. Problems: focus only on accuracy improvements without paying heed to communication compression
%Fourth Paragraph: We propose that it is possible to improve both by utilizing flatness-seeking local optimizers like SAM. Cite DFL+SAM work, mention the different setting and lack of focus on communication costs.

%\subsection{Contributions}
In summary, we make the following contributions:
\begin{itemize}
 \setlength\itemsep{-0.1mm}
    \item We propose Sharpness-Aware Decentralized Deep Learning (SADDLe) to seek flatter loss landscapes in decentralized learning, alleviating the local over-fitting with non-IID data.
    % \item Leveraging the fact that flatter loss landscapes tend to be more robust to compression errors, we demonstrate that SADDLe improves communication efficiency in the presence of data heterogeneity. 
    \item Leveraging the fact that flatter loss landscapes tend to be more robust to perturbations, we demonstrate that SADDLe improves robustness to communication compression in the presence of data heterogeneity. 
    \item We theoretically establish that SADDLe leads to a convergence rate of $\mathcal{O}(1/\sqrt{nT})$, similar to existing decentralized learning algorithms \cite{dpsgd}.
    \item Through extensive experiments on various datasets, models, graphs, and compression schemes, we show that Q-SADDLe and N-SADDLe result in a 1-20\% improvement in test accuracy. Additionally, our proposed algorithms maintain a minimal accuracy drop of 1\% for up to 4$\times$ compression, in contrast to the 4.3\% average accuracy drop for the baselines.
\end{itemize}

\section{Related Work}
\textbf{Data Heterogeneity.}
The impact of data heterogeneity in decentralized learning is an active area of research \cite{qgm,ngm,cga,gut,gt,d2}. Quasi-Global Momentum (QGM) \cite{qgm} improves decentralized learning with non-IID data through a globally synchronized momentum buffer. Gradient Tracking \cite{gt} tracks average gradients but requires 2x communication overhead as compared to DPSGD \cite{dpsgd}, while Global Update Tracking \cite{gut} tracks the average model updates to enhance performance with heterogeneous data. Cross Gradient Aggregation (CGA) \cite{cga} and Neighborhood Gradient Mean (NGM) \cite{ngm} utilize cross-gradient information through an extra communication round, achieving state-of-the-art performance in terms of test accuracy. In this work, we take an orthogonal route and focus on improving local training with a flatness-seeking optimizer \cite{sam} to achieve better generalization. 

\textbf{Communication Compression.}
Several algorithms have been proposed for communication-restricted decentralized settings \cite{dcd, deepsqueeze, choco-sgd, beer}. %The authors in \cite{dcd} introduced two quantized decentralized algorithms: Difference Compression DPSGD (DCD) and Extrapolation Compression DPSGD (ECD). DCD quantizes the difference, while ECD quantizes the extrapolation between the last two local models. 
DeepSqueeze \cite{deepsqueeze} introduced error-compensated compression to decentralized learning.
%, and achieved higher compression ratios than DCD and ECD. 
Choco-SGD \cite{choco-sgd} communicates compressed model updates rather than parameters and achieves better accuracy than DeepSqueeze. Recently, BEER \cite{beer} adopted communication compression with gradient tracking \cite{gt}, resulting in a faster convergence rate than Choco-SGD \cite{choco-sgd}. However, as shown in QGM \cite{qgm}, gradient tracking doesn't scale well for deep learning models and requires further study.
%REWRITE: Focus on how directly applying these techniques in existing algorithms for non-IID data leads to degradation, which is fixed by incorporating SADDLe.
In this paper, we compress the first (and only) communication round in Q-SADDLe and the second round in N-SADDLe. In both cases, we observe that SADDLe aids communication efficiency by alleviating the severe accuracy degradation incurred due to compression in existing decentralized learning algorithms for non-IID data \cite{qgm, ngm}.
% To show how SAM aids communication compression in the presence of data heterogeneity, we compress the first (and only) communication round in QGM and the second in NGM. In both cases, we observe that incorporating SAM optimizer at each agent can alleviate the severe accuracy degradation incurred due to compression.

% We show that Choco-SGD \cite{choco-sgd} with QGM \cite{qgm} leads to severe accuracy degradation, which can be alleviated by incorporating SAM as a local optimizer. Similarly, we observe that it is possible to compress the second communication round in NGM \cite{ngm} with minimal accuracy drop, 

% Therefore, we use Choco-SGD with QGM \cite{qgm} and NGM \cite{ngm} and show that utilizing SAM optimizer significantly improves the test accuracy and reduces the accuracy loss incurred due to compression.
%Choco-SGD, Deep-Squeeze, BEER
%CoDeC: mention a different setting
\textbf{Sharpness-Aware Minimization.}
% The relationship between the flatness of minima and generalization has been studied extensively \cite{keskar, jiang}.
Sharpness-Aware Minimization (SAM) \cite{sam} explores the connection between the flatness of minima and generalization by simultaneously minimizing loss value and loss sharpness during training \cite{keskar, jiang}. The authors in \cite{understandsam} provide a theoretical understanding of SAM through convergence results.
%and discussing its link to better generalization. 
%The authors in \cite{practicalsam} analyze the convergence guarantees further, and prove that SAM can converge only up to the neighborhoods of optima. 
Several variants of SAM have been proposed for centralized learning \cite{penalizing, asam, esam, samfree, adaptivesam, mi2022make}. In addition, there have been several efforts to improve the generalization performance in federated learning using SAM \cite{qu2022generalized, caldarola, sun2023dynamic, fedgamma, dflsam}. The authors in \cite{avsam} provide some theoretical insights to establish an asymptotic equivalence between decentralized training and average-direction SAM.
%Recently, DFedSAM \cite{dflsam} was proposed to improve the model consistency in decentralized federated learning settings, where agents communicate with each other after multiple local iterations. 
In contrast, our work focuses on simultaneously improving test accuracy and robustness to communication compression for decentralized learning with extreme data heterogeneity. 
%Further, we focus on extreme data heterogeneity settings and simultaneously improve test accuracy and communication efficiency. 

%recent variants of SAM- ASAM, Efficient-SAM, Penalizing grad norms etc.
%SAM used to improve FL performance
%SAM used to improve DFL, clarify how this is a different setting
%To the best of our knowledge, we are the first one to analyze the impact of SAM in distributed setting on communication-compression

\section{Background}
% \subsection{Overview of SAM}
% \subsection{Decentralized Learning}
%basic stuff about decentralized learning setup, loss functions, non-IID dirichlet explanation, communication compression basics
A global model is learned in decentralized learning by aggregating models trained on locally stored data at $n$ agents connected in a sparse graph topology. This topology is modeled as a graph $G=([n], \mathbf{W})$, where $\mathbf{W}$ is the mixing matrix indicating the graph's connectivity. Each entry $w_{ij}$ in $\mathbf{W}$ encodes the effect of agent $j$ on agent $i$, and $w_{ij}=0$ implies that agents $i$ and $j$ are not connected directly. $\mathcal{N}(i)$ represents neighbors of $i$ including itself. We aim to minimize the global loss function $f(\mathbf{x})$ shown in equation \ref{eq:1}. Here, $F_i(\mathbf{x}; d_i)$ is the local loss function at agent $i$, and $f_i(\mathbf{x})$ is the expected value of $F_i(\mathbf{x}; d_i)$ over the dataset $D_i$.
\vspace{-0.5mm}
\begin{equation}
\begin{split}
\label{eq:1}
    & \min \limits_{\mathbf{x} \in \mathbb{R}^d} f(\mathbf{x}) = \frac{1}{n}\sum_{i=1}^n f_i(\mathbf{x}); \hspace{1mm} f_i(\mathbf{x}) = \mathbb{E}_{d_i \sim D_i}[F_i(\mathbf{x}; d_i)]\\
    %& \text{where} \hspace{2mm} f_i(\mathbf{x}) = \mathbb{E}_{d_i \sim D_i}[F_i(\mathbf{x}; d_i)] \hspace{1mm} \forall I
\end{split}
\end{equation}

DPSGD\cite{dpsgd} tackles this by combining Stochastic Gradient Descent (SGD) with gossip averaging algorithms \cite{gossip}. Each agent maintains model parameters $\mathbf{x}_i^t$, computes local gradient $\mathbf{g}_i^t$ through SGD, and incorporates neighborhood information as shown in the following update rule:
% \vspace{-0.7mm}
\begin{equation*}
\label{eq:dsgd}
\begin{split}
   \text{DPSGD:} \hspace{1mm} \mathbf{x}_i^{t+1} = \sum_{j \in \mathcal{N}(i)}w_{ij}\mathbf{x}_j^t - \eta \mathbf{g}_j^t;\mathbf{g}_j^t=\nabla F_j(\mathbf{x}_j^t, d_j^t).  
\end{split}
\vspace{-1mm}
\end{equation*}
% Traditional decentralized learning algorithms like DPSGD assume the data distribution across the agents to be IID. However, in the presence of data heterogeneity, DPSGD leads to performance degradation due to local over-fitting. 
DPSGD assumes the data distribution across the agents to be IID and results in significant performance degradation in the presence of data heterogeneity.
To handle this, QGM \cite{qgm} incorporates a globally synchronized momentum buffer within DPSGD. 
%This mitigates the impact of non-IID data by maintaining a form of global information through the momentum buffer.QGM improves the test accuracy on non-IID data without introducing any communication overhead. 
This mitigates the impact of non-IID data by maintaining a form of global information through the momentum buffer, resulting in better test accuracy without any extra communication overhead. To further improve the performance with extreme heterogeneity, NGM\cite{ngm} and CGA\cite{cga} utilize cross-gradients obtained through an additional communication round.
In the first communication round, the agents exchange models with each other (similar to DPSGD). However, in the second round, the agents communicate cross-gradients computed over the neighbors' models and their local data. Each gradient update is a weighted average of the self and received cross-gradients \cite{ngm}. Note that these algorithms represent two distinct variants proposed to enhance decentralized learning with non-IID data based on the available communication budget. For additional details, refer to Appendix Section \ref{apex:bg}.

\section{Methodology}
%%Add a couple of lines here to explain what is 4.1 and 4.2
This section presents the two variants of SADDLe and their communication-compressed versions.
\subsection{SADDLe}
In the presence of data heterogeneity, models in decentralized training tend to overfit the local data at each agent. Aggregating such models adversely impacts the global model's generalization ability. To circumvent this, we propose SADDLe, which purposely seeks a flatter loss landscape in each training iteration through Sharpness-Aware Minimization (SAM) \cite{sam}.
Instead of focusing on finding parameters with low loss values like SGD, SAM searches for parameters whose neighborhoods have uniformly low loss. This is achieved by adding a perturbation $\xi_i$ to the model parameters, which is obtained through a scaled gradient ascent step. To summarize, SADDLe aims to solve the following optimization problem:
%Incorporating SAM in decentralized training results in the following optimization formulation:
\vspace{-0.7mm}
\begin{equation}\label{eq:samupdate}
\begin{split}
    &f(\mathbf{x})= \frac{1}{n}\sum_{i=1}^n \mathbb{E}_{d_i \sim D_i} \max_{\|\xi_i\| \leq \rho} [F_i(\mathbf{x}_i^t+\xi_i; d_i)] \hspace{2mm} \forall i,\\
    &\text{where} \hspace{2mm} \xi_i= \rho\frac{\mathbf{g}_i}{\|\mathbf{g}_i\|}
\end{split}
\end{equation}
Here, $\rho$ is a tunable hyperparameter, defining the perturbation radius. Since SADDLe modifies the local optimizer at each agent, it is orthogonal to existing techniques and can be used in synergy to improve performance with non-IID data. We employ a QGM buffer \cite{qgm} and cross-gradients similar to NGM \cite{ngm} with SADDLe and present two versions: Q-SADDLe and N-SADDLe. 

\begin{algorithm}[h]
%\small
\textbf{Input:} Each agent $i \in [1,n]$ initializes model parameters $\mathbf{x}_i$, $\hat{\mathbf{m}}_i^{(0)}$= 0, step size $\eta$, momentum coefficients $\beta, \mu$, mixing matrix $\mathbf{W}=[w_{ij}]_{i,j \in [1,n]}$, $\mathcal{N}(i)$ represents neighbors of $i$ including itself.\\

  \textbf{procedure} T\text{\scriptsize RAIN}( ) for $\forall i$\\
1.  \hspace{4mm}\textbf{for} t = $1,2,\hdots,T$ \textbf{do}\\
2.  \hspace*{8mm}$\mathbf{g}_{i}^{t}=\nabla F_i(\mathbf{x}_i^{t}; d_i^t)$ for $d_i^{t} \sim D_i$\\
3.  \hspace*{8mm}\colorbox{lightapricot}{$\mathbf{m}_i^{(t)}= \beta \hat{\mathbf{m}}^{(t-1)}_i + \mathbf{g}_i^{(t)}$} \\
4.  \hspace*{8mm}\colorbox{lightmintbg}{$\widetilde{\mathbf{g}}_{i}^{t}=\nabla F_i(\mathbf{x}_i^{t}+ \xi(\mathbf{x}_i^{t}); d_i^t)$, where $\xi(\mathbf{x}_i^{t})= \rho \frac{\mathbf{g}_i^t}{\|\mathbf{g}_i^t\|}$} \\

5.  \hspace*{8mm}\colorbox{lightmintbg}{$\mathbf{m}_i^{(t)}= \beta \hat{\mathbf{m}}^{(t-1)}_i + \widetilde{\mathbf{g}}_i^{(t)}$} \\
6.  \hspace*{8mm} $\mathbf{x}_i^{(t+1/2)}= \mathbf{x}_i^{(t)}- \eta \mathbf{m}_i^{(t)}$\\
7.  \hspace*{8mm}S\text{\scriptsize END}R\text{\scriptsize ECEIVE}($ \mathbf{x}_i^{(t+1/2)}$) \\
8.  \hspace*{8mm}$\mathbf{x}_i^{t+1}= \sum_{j \in \mathcal{N}_i^{(t)}}w_{ij}\mathbf{x}_j^{(t+1/2)}$ \\
9.  \hspace*{7mm}$\mathbf{d}_i^{(t)}= \frac{\mathbf{x}_i^{(t)}-\mathbf{x}_i^{(t+1)}}{\eta}$\\
10. \hspace*{7mm}$ \hat{\mathbf{m}}^{(t)}= \mu \hat{\mathbf{m}}_i^{(t-1)} + (1-\mu)\mathbf{d}_i^{(t)}$\\
11. \hspace{4mm}\textbf{end}\\
  \textbf{return $\mathbf{x}_i^{T}$}
\caption{\colorbox{lightapricot}{QGM} v.s. \colorbox{lightmintbg}{Q-SADDLe}}
\label{alg:QGSAM}
\end{algorithm}

The differences between QGM and Q-SADDLe are highlighted in Algorithm \ref{alg:QGSAM}. In particular, Q-SADDLe utilizes the SAM-based gradient update $\widetilde{\mathbf{g}}_i$ shown in line 4 instead of the gradient update $\mathbf{g}_i$. N-SADDLe employs SAM-based self and cross-gradients to further improve the performance of NGM. Algorithm \ref{alg:NGM} in the Appendix summarizes the difference in training procedures for NGM and N-SADDLe.

\subsection{SADDLe with Compressed Communication} \label{sec:compsam}
A major concern in decentralized learning is the high communication cost of training. Hence, we also investigate the impact of a flatter loss landscape on generalization performance in the presence of communication compression. We present compressed versions of QGM and Q-SADDLe in Algorithm \ref{alg:CompQGSAM}. Instead of sharing models $\mathbf{x}_i$, the agents exchange compressed model updates $\mathbf{q}_i$ (similar to Choco-SGD \cite{choco-sgd}). Each agent maintains compressed copies $\hat{\mathbf{x}}_j$ of their neighbors and employs a modified gossip averaging step as shown on line 7 (Algorithm \ref{alg:CompQGSAM}). Similarly, we implement Comp NGM and Comp N-SADDLe to compress the second communication round, which involves sharing cross-gradients (Algorithm \ref{apx_alg:compNGM} in Appendix). In addition to robustness to data heterogeneity, seeking flatter models also results in higher resiliency to compression error (as indicated by our results in Tables \ref{tab:qgmcf10_100}-\ref{tab:effqsaddle}).

\begin{figure*}[h]
\centering
	\begin{subfigure}{0.4\textwidth}
		\includegraphics[width=\textwidth]{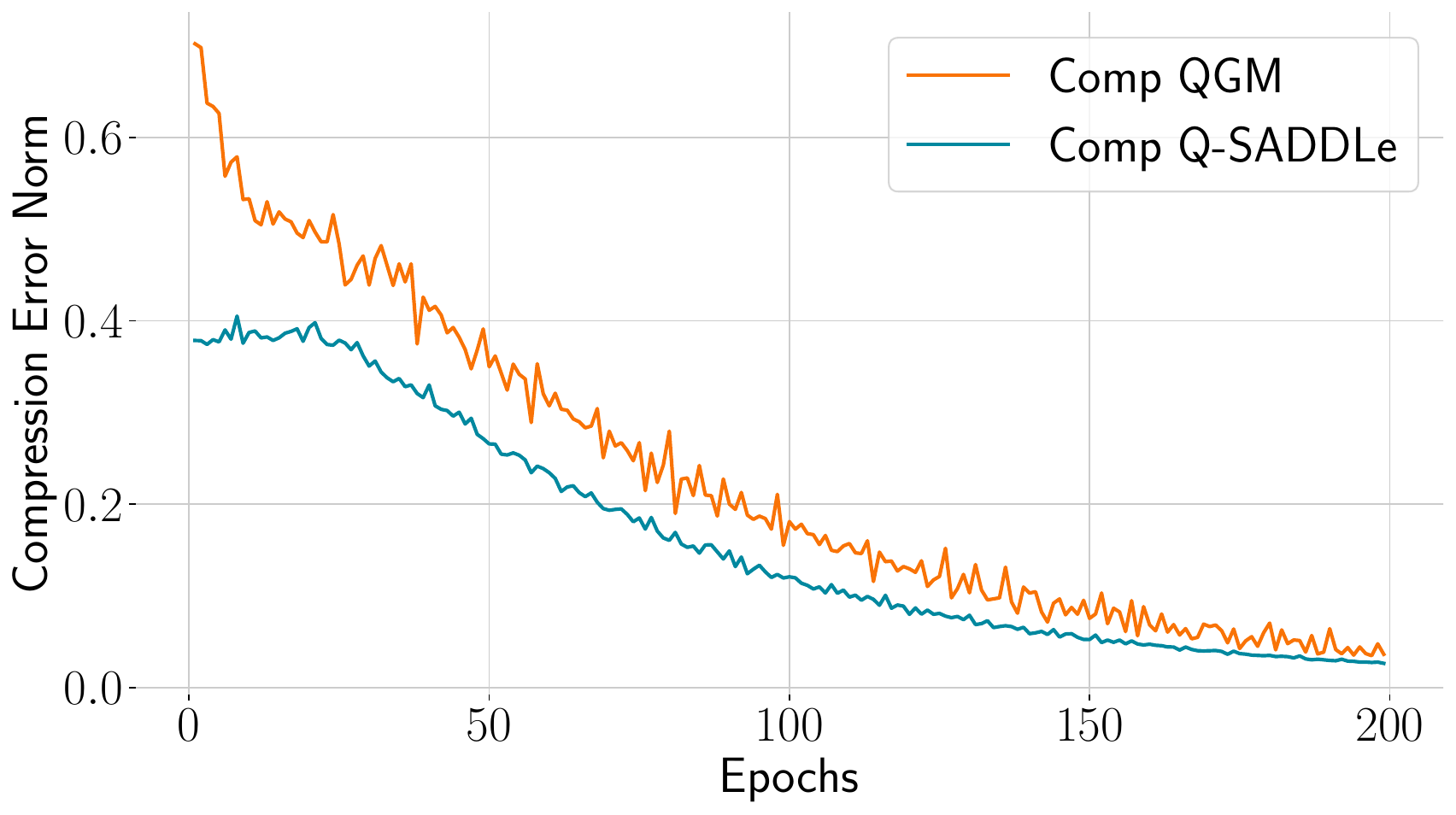}
		\caption{Compression Error $(\|Q(\mathbf{\theta})- \theta \|)$}
	\end{subfigure}
	\begin{subfigure}{0.4\linewidth}
		\includegraphics[width=\textwidth]{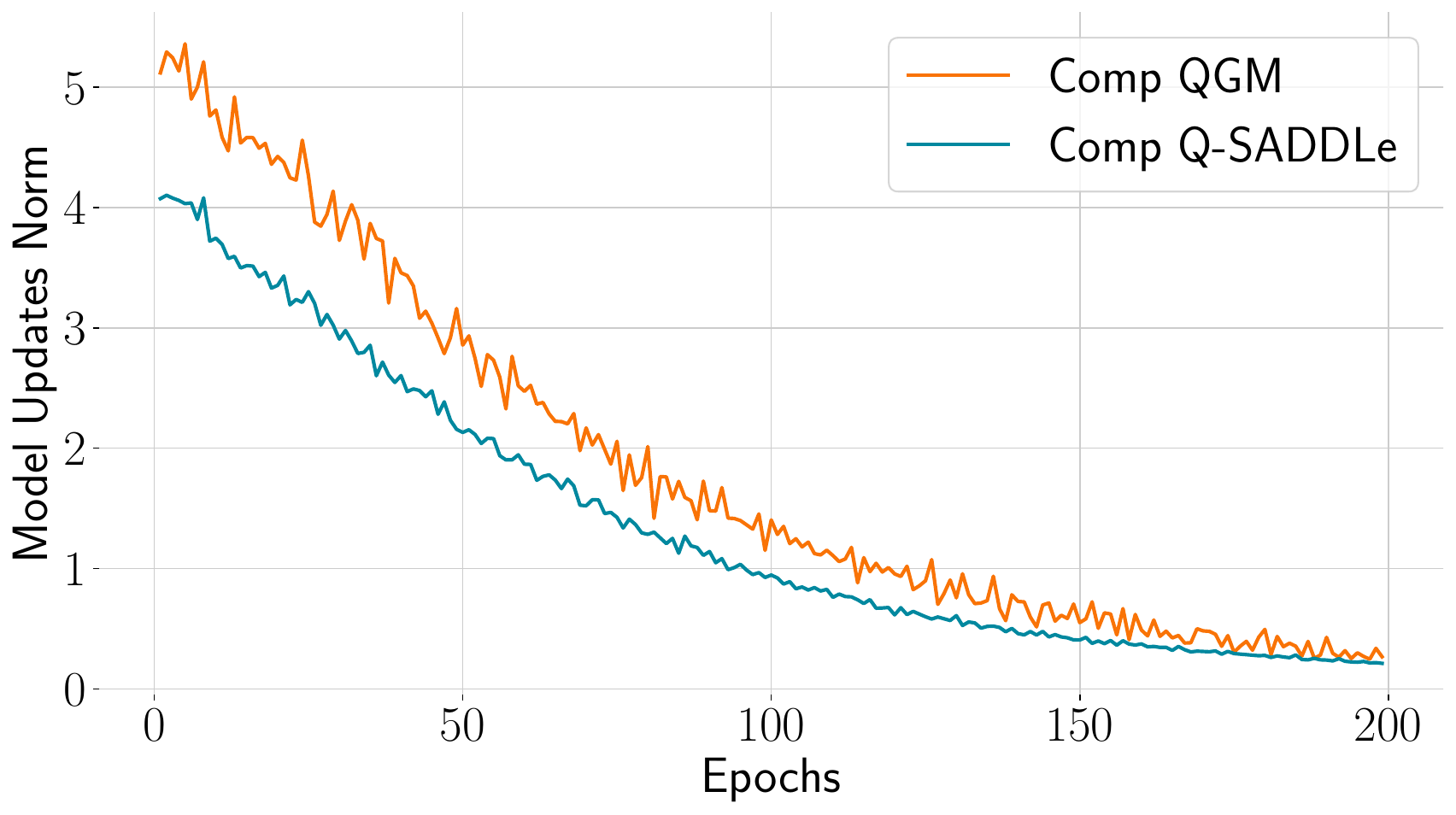}
		\caption{Model Updates $(\|\mathbf{x}_i-\hat{\mathbf{x}}_i\|)$}
	\end{subfigure}
	\caption{Impact of flatness on (a) Compression Error and (b) Model Updates for ResNet-20 trained on CIFAR-10 distributed in a non-IID manner across a 10 agent ring topology.}
	\label{fig:comperror}
\end{figure*}

\begin{algorithm}[h]
%\small
\textbf{Input:} Each agent $i \in [1,n]$ initializes model parameters $\mathbf{x}_i$ and $\hat{\mathbf{x}}_i^{1}=0$, step size $\eta$,  momentum coefficients $\beta, \mu$, global averaging rate $\gamma$, mixing matrix $\mathbf{W}=[w_{ij}]_{i,j \in [1,n]}$.\\

  \textbf{procedure} T\text{\scriptsize RAIN}( ) for $\forall i$\\
1.  \hspace{4mm}\textbf{for} t = $1,2,\hdots,T$ \textbf{do}\\
2.  \hspace{8mm}$\mathbf{g}_{i}^{t}=\nabla F_i(\mathbf{x}_i^{t}; d_i^t)$ for $d_i^{t} \sim D_i$\\
3.  \hspace{8mm}\colorbox{lightapricot}{$\mathbf{m}_i^{(t)}= \beta \hat{\mathbf{m}}^{(t-1)}_i + \mathbf{g}_i^{(t)}$} \\
4.  \hspace*{8mm}\colorbox{lightmintbg}{$\widetilde{\mathbf{g}}_{i}^{t}=\nabla F_i(\mathbf{x}_i^{t}+ \xi(\mathbf{x}_i^{t}); d_i^t)$, where $\xi(\mathbf{x}_i^{t})= \rho \frac{\mathbf{g}_i^t}{\|\mathbf{g}_i^t\|}$} \\

5.  \hspace*{8mm}\colorbox{lightmintbg}{$\mathbf{m}_i^{(t)}= \beta \hat{\mathbf{m}}^{(t-1)}_i + \widetilde{\mathbf{g}}_i^{(t)}$} \\
6.  \hspace*{7mm} $\mathbf{x}_i^{(t+1/2)}= \mathbf{x}_i^{(t)}- \eta \mathbf{m}_i^{(t)}$\\
7.  \hspace*{7mm} $\mathbf{x}_i^{(t+1)}= \mathbf{x}_i^{(t+1/2)}+ \gamma \sum_{j \in \mathcal{N}(i)}w_{ij} (\hat{\mathbf{x}}_j^{(t)}-\hat{\mathbf{x}}_i^{(t)})$\\
8.  \hspace*{8mm}$\mathbf{d}_i^{(t)}= \frac{\mathbf{x}_i^{(t)}-\mathbf{x}_i^{(t+1)}}{\eta}$\\
9. \hspace*{8mm}$ \hat{\mathbf{m}}^{(t)}= \mu \hat{\mathbf{m}}_i^{(t-1)} + (1-\mu)\mathbf{d}_i^{(t)}$\\
10. \hspace*{6mm} $\mathbf{q}_i^{(t)}= Q(\mathbf{x}_i^{(t+1)}-\hat{\mathbf{x}}_i^{(t)})$\\
11.  \hspace*{7mm}S\text{\scriptsize END}R\text{\scriptsize ECEIVE}($\mathbf{q}_i^{(t)}$) \\
12.  \hspace*{6mm} $\hat{\mathbf{x}}_j^{(t+1)}= \mathbf{q}_j^{(t)} + \hat{\mathbf{x}}_j^{(t)}$ for all $j \in \mathcal{N}(i)$\\
13. \hspace{4mm}\textbf{end}\\
  \textbf{return $\mathbf{x}_i^{T}$}
\caption{\colorbox{lightapricot}{Comp QGM} v.s. \colorbox{lightmintbg}{Comp Q-SADDLe}}
\label{alg:CompQGSAM}
\end{algorithm}

Interestingly, Comp Q-SADDLe and Comp N-SADDLe incur less compression error than Comp QGM and Comp NGM respectively, leading to a lower accuracy drop due to compression. We investigate this with the aid of a well-known bound on the compression error \cite{qsgd}. For a compression operator $Q(.)$, the expectation of error $\|Q(\mathbf{\theta})- \theta\|$ is bounded as:
\vspace{-0.5mm}
\begin{equation}\label{eq:compbound}
    \mathbb{E}_Q \|Q(\mathbf{\theta})- \theta\|^2 \leq (1-\zeta) \|\theta\|^2, \hspace{1mm} \text{where} \hspace{1mm} \zeta>0
\end{equation}
% We investigate this in the context of stochastic quantization-based compressor \cite{qsgd} (i.e., $Q(.)$ at line 11 in Algorithm \ref{alg:CompQGSAM}). Let $\theta_{max}-\theta_{min}$ be the range of input $\theta$ to be quantized, and $b$ be the quantization precision. Under the assumption that quantization error is uncorrelated with the original signal, we have \cite{fit, sdq}:
% \begin{equation}
%     \mathbb{E}[(\theta-Q(\theta))^2]= \frac{1}{12}\bigg(\frac{\theta_{max}-\theta_{min}}{2^b-1}\bigg)^2
% \end{equation}
% The above equation establishes that for a fixed quantization precision $b$, a lower range implies a lower quantization error. 

In our setup, $\theta$ corresponds to model updates $(\mathbf{x}_i-\hat{\mathbf{x}}_i)$ for Comp QGM and Comp Q-SADDLe, and gradients for Comp NGM and Comp N-SADDLe. Note that a wide range of compression operators (with some $\zeta$) have been shown to adhere to this bound \cite{choco-sgd, pmlr-v97-koloskova19a, qsgd, topk, topk_2}. Figure \ref{fig:comperror} shows the norm of compression error (i.e. $\|Q(\mathbf{\theta})- \theta\|$) and the norm of model updates (i.e. $\|\theta\|= \|\mathbf{x}_i-\hat{\mathbf{x}}_i\|$) for Comp QGM and Comp Q-SADDLe for ResNet-20 trained on CIFAR-10 in a 10 agent ring with extreme data heterogeneity. It can be observed that Q-SADDLe leads to a lower norm of model updates ($\|\theta\|$) and lower compression error. In essence, the bound in Equation \ref{eq:compbound} is tighter for Q-SADDLe than QGM in the presence of compression. We observe similar trends for N-SADDLe (refer to Figure \ref{fig:comperror_ngm} in Appendix). Note that our observation regarding lower gradient norms (and hence model updates) in the presence of SAM aligns with the fact that SAM optimization is a special form of penalizing the gradient norm \cite{penalizing}.

\section{Convergence Rate Analysis}\label{section:convergence}
In this section, we provide a convergence analysis for Q-SADDLe.
Similar to prior works in decentralized learning \cite{dpsgd,qgm}, we make the following standard assumptions:

\textbf{Assumption 1 - Lipschitz Gradients:} Each function $f_i(\mathbf{x})$ is L-smooth i.e., $||\nabla f_i(\mathbf{y}) - \nabla f_i(\mathbf{x})||\leq L ||\mathbf{y}-\mathbf{x}||$.

\textbf{Assumption 2 - Bounded Variance:} The variance of the stochastic gradients is assumed to be bounded. There exist constants $\sigma$ and $\delta$ such that
\begin{equation}
\begin{split}
\label{eq:variance}
&\mathbb{E}_{ d \sim \mathcal{D}_i} || \nabla F_i(\mathbf{x}; d) - \nabla f_i(\mathbf{x})||^2 \leq \sigma^2\\
&\hspace{1mm} \frac{1}{n} \sum_{i=1}^n || \nabla f_i(\mathbf{x}) - \nabla f(\mathbf{x})||^2 \leq \delta^2 \hspace*{2mm} \forall i
\end{split}
\end{equation}
\textbf{Assumption 3 - Doubly Stochastic Mixing Matrix:} $\mathbf{W}$ is a real doubly stochastic matrix which satisfies $\mathbb{E}_{\mathbf{W}} \norm{\mathbf{Z} \mathbf{W} - \bar{\mathbf{Z}}}^2 \leq (1 - \lambda) \norm{\mathbf{Z} - \bar{\mathbf{Z}}}^2$ for any matrix $\mathbf{Z} \in  \mathbb{R}^{d \times n}$ and
$\bar{\mathbf{Z}} = \mathbf{Z} \frac{1}{n}^T$. 
%Q-SADDLe --- repeat all assumptions in DPSGD, show how the 'rho' terms only show up in the higher order terms

Theorem \ref{theorem_1} presents convergence of the proposed Q-SADDLe algorithm (proof in Appendix Section \ref{apex:proof}).
\begin{theorem}\label{theorem_1}
Given Assumptions 1-3, for a momentum coefficients $\beta$ and $\mu$, let the learning rate satisfy $\eta \leq \min \left(\frac{\lambda}{7 L}, \frac{1-\beta}{4L} ,  \frac{(1-\beta)^2(1-\mu)}{\sqrt{12}L\beta}\right)$.
% \begin{equation} \label{eq:eta}
% \eta \leq \frac{\sqrt{(1-\sqrt{\lambda})^2+12(1-\sqrt{\lambda})}-(1-\sqrt{\lambda})}{6L}
% \end{equation}
For all $T \geq 1$, we have
\begin{equation} \label{eq:theorem}
    \begin{split}
&\frac{1}{T}\sum_{t=0}^{T-1} \mathbb{E}\left[\left\|\nabla f\left(\bar{\mathbf{x}}_{t}\right)\right\|^{2}\right] \leq
\frac{4}{\tilde\eta T}(\mathbb{E}[f(\Bar{x}^0)-f^*])+
\bigg(\frac{6\tilde\eta L}{n}+\\
&18\tilde\eta^2 L^2C_2+\frac{768\tilde\eta^2L^2C_1}{\lambda^2}\bigg)\sigma^2+\bigg(\frac{832L^2\tilde\eta^2(1-\beta)^2}{\lambda^2}\bigg)\delta^2\\
&+\bigg(8 L^2+\frac{12\tilde\eta L^3}{n}+36\tilde\eta^2L^4C_2+\frac{3136L^4\tilde\eta^2C_1}{\lambda^2}\bigg)\rho^2
    \end{split}
\end{equation} 
where $C_1= \frac{(2-\beta-\mu)(1-\beta)^2}{(1-\mu)}$, $C_2=\frac{\beta^2}{(1-\mu)(1-\beta)}$, $\Bar{x}$ is the average/consensus model and $\tilde\eta=\frac{\eta}{(1-\beta)}$.
\end{theorem}

We observe that the convergence rate includes three main terms related to the suboptimality gap $f(\Bar{x}^0)-f^*$, the sampling variance $\sigma$ and the gradient variance $\delta$ representing data heterogeneity, followed by an additional term compared to existing state-of-the-art decentralized convergence bounds \cite{dpsgd,qgm}. This term includes the perturbation radius $\rho$, signifying the impact of leveraging gradient perturbation to improve generalization in decentralized learning. We present a corollary to show the convergence rate in terms of training iterations (proof in Appendix Section \ref{apex:proof_corol}).
\vspace{-2mm}
\begin{corollary}
\label{corol}
Suppose that the learning rate satisfies $\eta=\mathcal{O}\Big(\sqrt{\frac{n}{T}}\Big)$ and $\rho= \mathcal{O}\Big(\sqrt{\frac{1}{T}}\Big)$.  
For a sufficiently large $T$, 
\begin{equation}
\begin{split}
&\frac{1}{T}\sum_{t=0}^{T-1} \mathbb{E}\left[\left\|\nabla f\left(\bar{\mathbf{x}}_{t}\right)\right\|^{2}\right] \leq \mathcal{O}(\frac{1}{\sqrt{nT}}+\frac{1}{T}+\frac{1}{ T^{3/2}}+\frac{1}{T^2})
\end{split}
\end{equation}
\end{corollary}
\vspace{-2mm}
Note that the dominant term here is $(1/\sqrt{nT})$, and the terms introduced because of the additional SGD step for flatness (i.e., $1/T^{3/2}$ and $1/T^2$) can be ignored due to their higher order (similar to \cite{qu2022generalized, dflsam, practicalsam}). This convergence rate matches the well-known best result in existing decentralized learning algorithms \cite{dpsgd}.

\section{Experiments}

\begin{table*}[htbp]
\vspace{-2.5mm}
\caption{Test accuracy of QGM, Q-SADDLe, and their compressed versions evaluated on CIFAR-10 and CIFAR-100 over ResNet-20, distributed over ring topologies. Comp implies stochastic quantization \cite{qsgd}  with 8 bits, which leads to 4$\times$ lower communication cost.}
\vspace{-2mm}
\label{tab:qgmcf10_100}
\small
\begin{center}
%\resizebox{1.0\columnwidth}{!}{
\begin{tabular*} {\textwidth}{ccl @{\extracolsep{\fill}}*{4}{c}}
\hline
\multirow{ 2}{*}{Agents} & \multirow{2}{*}{Comp} &\multirow{2}{*}{Method}& \multicolumn{2}{c}{CIFAR-10} & \multicolumn{2}{c}{CIFAR-100}\\
\cline{4-5}  
\cline{6-7}
& & & $\alpha=0.01$ & $\alpha=0.001$ & $\alpha=0.01$ & $\alpha=0.001$\\
 \hline
 \multirow{4}{*}{$5$} & & QGM &  88.44 $\pm$ 0.39 & 88.72 $\pm$ 0.64 & 56.84 $\pm$ 2.01 & 59.58 $\pm$ 1.22\\
 & \checkmark & QGM & 86.85 $\pm$ 0.73 & 86.82 $\pm$ 0.99 & 48.80 $\pm$ 8.58 & 51.99 $\pm$ 4.23\\
 & & \textit {Q-SADDLe (ours)} & \textbf{90.66 $\pm$ 0.08} & \textbf{90.56 $\pm$ 0.33} & \textbf{61.96 $\pm$ 1.00} & \textbf{61.64 $\pm$ 0.70}\\
 & \checkmark & \textit{Q-SADDLe (ours)} & \textbf{89.70 $\pm$ 0.15} & \textbf{90.02 $\pm$ 0.08} & \textbf{60.11 $\pm$ 0.99} & \textbf{60.53 $\pm$ 0.25}\\
 \hline
 \multirow{4}{*}{$10$} & & QGM & 77.41 $\pm$ 8.00 & 79.48 $\pm$ 2.76 & 48.06 $\pm$ 4.36 & 44.16 $\pm$ 6.71 \\
 & \checkmark & QGM & 76.59 $\pm$ 5.95 & 73.03 $\pm$ 4.63 & 46.14 $\pm$ 6.88 & 43.00 $\pm$ 6.55 \\
 & & \textit {Q-SADDLe (ours)} & \textbf{87.72 $\pm$ 1.59} & \textbf{86.33 $\pm$ 0.24} & \textbf{58.06 $\pm$ 0.68} & \textbf{56.76 $\pm$ 0.86}\\
 & \checkmark & \textit{Q-SADDLe (ours)} & \textbf{87.82 $\pm$ 1.42} & \textbf{85.57 $\pm$ 1.33} & \textbf{57.90 $\pm$ 0.82} & \textbf{56.27 $\pm$ 0.84} \\
  \hline
%\multirow{5}{*}{$15$} & DSGDm (IID) & \multicolumn{2}{c}{-}  \\
 \multirow{4}{*}{$20$} & & QGM &  72.20 $\pm$ 0.77 & 62.48 $\pm$ 8.56 & 45.23 $\pm$ 3.26 & 44.48 $\pm$ 4.53 \\
 & \checkmark & QGM & 66.61 $\pm$ 6.68 & 60.30 $\pm$ 6.60 & 43.64 $\pm$ 3.68 & 42.75 $\pm$ 1.82\\
 & & \textit {Q-SADDLe (ours)} & \textbf{78.41 $\pm$ 2.13} & \textbf{82.81 $\pm$ 0.89} & \textbf{52.59 $\pm$ 0.48} & \textbf{48.20 $\pm$ 0.93} \\
 & \checkmark & \textit{Q-SADDLe (ours)} & \textbf{80.80 $\pm$ 2.20} & \textbf{82.18 $\pm$ 0.56} & \textbf{52.64 $\pm$ 1.09} & \textbf{48.00 $\pm$ 0.85} \\
  \hline
%\multirow{5}{*}{$20$} & DSGDm (IID) & \multicolumn{2}{c}{-}  \\
 \multirow{4}{*}{$40$} & & QGM & 70.46 $\pm$ 4.14 & 60.86 $\pm$ 0.98 & 40.15 $\pm$ 0.90 & 38.73 $\pm$ 1.47 \\
 & \checkmark & QGM & 67.81 $\pm$ 2.62 & 57.01 $\pm$ 1.88 & 35.36 $\pm$ 1.50 & 36.04 $\pm$ 1.31\\
 & & \textit {Q-SADDLe (ours)} & \textbf{77.49 $\pm$ 0.83} & \textbf{73.54 $\pm$ 2.04} & \textbf{43.25 $\pm$ 1.71} & \textbf{41.99 $\pm$ 1.27} \\
 & \checkmark & \textit{Q-SADDLe (ours)} & \textbf{76.35 $\pm$ 0.42} & \textbf{72.03 $\pm$ 2.12} & \textbf{41.75 $\pm$ 2.14} & \textbf{41.03 $\pm$ 0.67} \\
 %%%%%%%%%%%%%%%%%%%
 \hline
\end{tabular*}
\end{center}
\vspace{-3mm}
\end{table*}

\begin{table}[ht]
\caption{Test accuracy of QGM, Q-SADDLe, and their compressed versions evaluated on Imagenette, distributed over a ring topology. Comp implies stochastic quantization \cite{qsgd}  with 10-bits.}
\vspace{-2mm}
\label{tab:qgmimagenette}
\small
\begin{center}
% \begin{tabular*} {\textwidth}{ccl @{\extracolsep{\fill}}*{2}{c}}
\addtolength{\tabcolsep}{-0.2em}
 \begin{tabular}{ccccc}
\hline
\multirow{ 2}{*}{Agents} & \multirow{2}{*}{Comp} & \multirow{ 2}{*}{Method} & \multicolumn{2}{c}{Imagenette (MobileNet-V2)}\\
\cline{4-5}  
& & & $\alpha=0.01$ & $\alpha=0.001$\\
 \hline
 \multirow{4}{*}{$5$} & & QGM &  64.25 $\pm$ 11.53 & 57.67 $\pm$ 6.32\\
 & \checkmark & QGM & 59.86 $\pm$ 17.05 & 50.14 $\pm$ 9.39 \\
 & & \textit {Q-SADDLe} & \textbf{73.34 $\pm$ 0.80} & \textbf{72.50 $\pm$ 0.21}\\
 & \checkmark & \textit{Q-SADDLe} & \textbf{72.64 $\pm$ 1.66} & \textbf{72.77 $\pm$ 0.43} \\
 \hline
 \multirow{4}{*}{$10$} & & QGM &  56.30 $\pm$ 4.03 & 45.82 $\pm$ 5.99 \\
 & \checkmark & QGM & 53.50 $\pm$ 5.66 & 36.71 $\pm$ 3.65 \\
 & & \textit {Q-SADDLe} & \textbf{62.35 $\pm$ 3.64} & \textbf{63.18 $\pm$ 1.59} \\
 & \checkmark & \textit{Q-SADDLe} & \textbf{63.35 $\pm$ 2.61} & \textbf{61.14 $\pm$ 0.71} \\
 %%%%%%%%%%%%%%%%%%%
 \hline
\end{tabular}
\end{center}
\vspace{-4mm}
\end{table}

\subsection{Experimental Setup}
\begin{figure}[ht!]
	\hfill
	\resizebox{0.5\linewidth}{!}{
		\resizebox{.2\linewidth}{!}{
			\begin{tikzpicture}[scale=1]
			\def \n {16}
			\def \radius {3.0cm}
			\def \margin {-4} %
			
			\foreach \s in {1,...,\n}
			{
				\node[draw, circle] at ({-360/\n * (\s - 1) + 90}:\radius) {};
				\draw[-, >=latex] ({-360/\n * (\s - 1)+\margin + 90}:\radius)
				arc ({-360/\n * (\s - 1)+\margin + 90}:{-360/\n * (\s)-\margin + 90}:\radius);
			}
			\end{tikzpicture}
   
		}
        \vspace{8mm}
		\hfill
		\resizebox{.2\linewidth}{!}{
			\begin{tikzpicture}[scale=2.5]
            \node[shape=circle,draw=black] (1) at (-1,-1.5) {};
            \node[shape=circle,draw=black] (2) at (0,-1.5) {};
			\node[shape=circle,draw=black] (3) at (1, -1.5) {};
			\node[shape=circle,draw=black] (4) at (2,-1.5) {};
            \node[shape=circle,draw=black] (5) at (-1,-1.0) {};
            \node[shape=circle,draw=black] (6) at (0,-1.0) {};
			\node[shape=circle,draw=black] (7) at (1, -1.0) {};
			\node[shape=circle,draw=black] (8) at (2,-1.0) {};
            \node[shape=circle,draw=black] (9) at (-1,-0.5) {};
            \node[shape=circle,draw=black] (10) at (0,-0.5) {};
			\node[shape=circle,draw=black] (11) at (1, -0.5) {};
			\node[shape=circle,draw=black] (12) at (2,-0.5) {};
			\node[shape=circle,draw=black] (13) at (-1,0) {};
			\node[shape=circle,draw=black] (14) at (0,0) {};
			\node[shape=circle,draw=black] (15) at (1, 0) {};
			\node[shape=circle,draw=black] (16) at (2,0) {};

            \node[shape=circle,draw=black] (17) at (-1,0.5) {};
			\node[shape=circle,draw=black] (18) at (0,0.5) {};
			\node[shape=circle,draw=black] (19) at (1,0.5) {};
			\node[shape=circle,draw=black] (20) at (2,0.5) {};
            \node[shape=circle,draw=black] (21) at (-1,1) {};
			\node[shape=circle,draw=black] (22) at (0,1) {};
			\node[shape=circle,draw=black] (23) at (1,1) {};
			\node[shape=circle,draw=black] (24) at (2,1) {};
			\node[shape=circle,draw=black] (25) at (-1,1.5) {};
			\node[shape=circle,draw=black] (26) at (0,1.5) {};
			\node[shape=circle,draw=black] (27) at (1, 1.5) {};
			\node[shape=circle,draw=black] (28) at (2,1.5) {};
			\node[shape=circle,draw=black] (29) at (-1,2) {};
            \node[shape=circle,draw=black] (30) at (0,2) {};
			\node[shape=circle,draw=black] (31) at (1, 2) {};
			\node[shape=circle,draw=black] (32) at (2,2) {};
            
			%\begin{scope}[>={Stealth[black]},
			every edge/.style={draw=black}]
            \path [dashed,-] (1) edge[bend left=30] node[left] {} (29);
            \path [dashed,-] (2) edge[bend left=30] node[left] {} (30);
             \path [dashed,-] (3) edge[bend left=30] node[left] {} (31);
             \path [dashed,-] (4) edge[bend left=30] node[left] {} (32);
            
			\path [-] (1) edge node[left] {} (2);
			\path [-] (2) edge node[left] {} (3);
            \path [-] (3) edge node[left] {} (4);
            \path [-] (1) edge node[left] {} (5);
            \path [-] (2) edge node[left] {} (6);
            \path [-] (3) edge node[left] {} (7);
            \path [-] (4) edge node[left] {} (8);
			\path [-] (5) edge node[left] {} (6);
			\path [-] (6) edge node[left] {} (7);
            \path [-] (7) edge node[left] {} (8);
			\path [-] (5) edge node[left] {} (9);
            \path [-] (6) edge node[left] {} (10);
			\path [-] (7) edge node[left] {} (11);
			\path [-] (8) edge node[left] {} (12);
			\path [-] (9) edge node[left] {} (10);
			\path [-] (10) edge node[left] {} (11);
            \path [-] (11) edge node[left] {} (12);
            \path [-] (9) edge node[left] {} (13);
            \path [-] (10) edge node[left] {} (14);
			\path [-] (11) edge node[left] {} (15);
			\path [-] (12) edge node[left] {} (16);
            \path [-] (13) edge node[left] {} (14);
			\path [-] (14) edge node[left] {} (15);
            \path [-] (15) edge node[left] {} (16);
            \path [-] (13) edge node[left] {} (17);
			\path [-] (14) edge node[left] {} (18);
            \path [-] (15) edge node[left] {} (19);
            \path [-] (16) edge node[left] {} (20);
            \path [-] (17) edge node[left] {} (18);
            \path [-] (18) edge node[left] {} (19);
			\path [-] (19) edge node[left] {} (20);
            \path [-] (17) edge node[left] {} (21);
            \path [-] (18) edge node[left] {} (22);
			\path [-] (19) edge node[left] {} (23);
            \path [-] (20) edge node[left] {} (24);
            \path [-] (21) edge node[left] {} (22);
            \path [-] (22) edge node[left] {} (23);
			\path [-] (23) edge node[left] {} (24);
            \path [-] (21) edge node[left] {} (25);
            \path [-] (22) edge node[left] {} (26);
			\path [-] (23) edge node[left] {} (27);
            \path [-] (24) edge node[left] {} (28);
            \path [-] (25) edge node[left] {} (26);
            \path [-] (26) edge node[left] {} (27);
			\path [-] (27) edge node[left] {} (28);
            \path [-] (25) edge node[left] {} (29);
            \path [-] (26) edge node[left] {} (30);
			\path [-] (27) edge node[left] {} (31);
            \path [-] (28) edge node[left] {} (32);
            \path [-] (29) edge node[left] {} (30);
			\path [-] (30) edge node[left] {} (31);
            \path [-] (31) edge node[left] {} (32);
            \path [dashed,-] (4) edge[bend right=20] node[left] {} (1);
            \path [dashed,-] (8) edge[bend right=20] node[left] {} (5);
            \path [dashed,-] (12) edge[bend right=20] node[left] {} (9);
            \path [dashed,-] (16) edge[bend right=20] node[left] {} (13);
            \path [dashed,-] (20) edge[bend right=20] node[left] {} (17);
            \path [dashed,-] (24) edge[bend right=20] node[left] {} (21);
            \path [dashed,-] (28) edge[bend right=20] node[left] {} (25);
            \path [dashed,-] (32) edge[bend right=20] node[left] {} (29);
			%\end{scope}
			\end{tikzpicture}
		}
	}
	\hfill\null
	\caption{Ring Graph (left), and Torus Graph (right). }\label{fig:topologies}
\end{figure}
We analyze the test accuracy and communication efficiency of the proposed Q-SADDLe and N-SADDLe and compare them with the state-of-the-art QGM and NGM.
We evaluate the proposed algorithms across diverse datasets, model architectures, graph topologies, graph sizes, and compression operators, with all models using Evonorm \cite{evonorm} as it is better suited for non-IID data \cite{evonorm2}. The analysis is presented on - (a) \textbf{Datasets}: CIFAR-10 \cite{cifar}, CIFAR-100 \cite{cifar}, Imagenette \cite{imagenette} and ImageNet \cite{imagenet}, (b) \textbf{Model architectures}: ResNet-20, ResNet-18 and MobileNet-v2, (c) \textbf{Graph topologies}: ring with 2 peers/agent and torus with 4 peers/agent (visualization in Figure \ref{fig:topologies}), (d) \textbf{Graph sizes}: 5 to 40 agents, (e) \textbf{Compression operators}: Stochastic quantization \cite{qsgd}, Top-k sparsification \cite{topk, topk_2} and Sign SGD \cite{efsgd}. For QGM, stochastic quantization diverges beyond 8-10 bits, and Top-k diverges beyond 30\% sparsification, likely due to erroneous compressed updates affecting both gossip and momentum buffers (lines 7-8, Algorithm \ref{alg:CompQGSAM}). However, in NGM, the second communication round is more compressible as it only affects gradient updates, making 1-bit Sign SGD viable for NGM and N-SADDLe.

We focus on non-IID data partitions generated by Dirichlet distribution \cite{dirichlet}, varying the concentration parameter $\alpha$—smaller $\alpha$ increases non-IIDness (see Figure \ref{fig:alphavisual} in Appendix). These partitions are non-overlapping, with no shuffling across the agents during training. Training hyperparameters are detailed in Section \ref{apx:hyperparams} of the Appendix.

\subsection{Results}\label{results}
\begin{table*}[ht]
\vspace{-2mm}
\caption{Test accuracy of NGM, N-SADDLe, and their compressed versions evaluated on CIFAR-10 and CIFAR-100 over ResNet-20, distributed with different degrees of heterogeneity over ring topologies. Comp implies 1-bit Sign SGD \cite{efsgd} based compression, which reduces the communication cost of the second round by 32$\times$ and the total communication cost by 1.94$\times$.}
\vspace{-2mm}
\label{tab:ngmcf10_100}
\small
\begin{center}
%\resizebox{1.0\columnwidth}{!}{
\begin{tabular*} {\textwidth}{ccl @{\extracolsep{\fill}}*{4}{c}}
\hline
\multirow{ 2}{*}{Agents} & \multirow{2}{*}{Comp} &\multirow{ 2}{*}{Method}& \multicolumn{2}{c}{CIFAR-10} & \multicolumn{2}{c}{CIFAR-100}\\
\cline{4-5}  
\cline{6-7}
& & & $\alpha=0.01$ & $\alpha=0.001$ & $\alpha=0.01$ & $\alpha=0.001$\\
 \hline
 \multirow{4}{*}{$5$} & & NGM &  90.87 $\pm$ 0.39 & 90.73 $\pm$ 0.46 & 59.00 $\pm$ 4.26 & 54.78 $\pm$ 4.68\\
 & \checkmark & NGM & 89.50 $\pm$ 0.68 & 87.69 $\pm$ 1.98 & 56.91 $\pm$ 1.82 & 50.65 $\pm$ 2.67 \\
 & & \textit {N-SADDLe (ours)} & \textbf{91.96 $\pm$ 0.19} & \textbf{91.69 $\pm$ 0.15} & \textbf{63.87 $\pm$ 0.45} & \textbf{64.10 $\pm$ 0.48}\\
 & \checkmark & \textit{N-SADDLe (ours)} & \textbf{91.88 $\pm$ 0.36} & \textbf{91.77 $\pm$ 0.19} & \textbf{62.35 $\pm$ 0.87} & \textbf{62.43 $\pm$ 0.36} \\
 \hline
 \multirow{4}{*}{$10$} & & NGM &  85.08 $\pm$ 2.73 & 83.43 $\pm$ 0.95 & 55.2 $\pm$ 1.41 & 54.70 $\pm$ 1.36\\
 & \checkmark & NGM & 76.85 $\pm$ 15.10 & 76.67 $\pm$ 3.67 & 43.41 $\pm$ 4.50 & 43.17 $\pm$ 4.65 \\
 & & \textit {N-SADDLe (ours)} & \textbf{88.43 $\pm$ 1.38} & \textbf{87.29 $\pm$ 1.23} & \textbf{59.31 $\pm$ 0.61} & \textbf{58.37 $\pm$ 0.30} \\
 & \checkmark & \textit{N-SADDLe (ours)} & \textbf{88.11 $\pm$ 1.54} & \textbf{87.14 $\pm$ 1.45} & \textbf{58.39 $\pm$ 0.89} & \textbf{58.33 $\pm$ 0.45} \\
  \hline
 \multirow{4}{*}{$20$} & & NGM &  84.84 $\pm$ 0.43 & 83.58 $\pm$ 0.89 & 53.98 $\pm$ 0.31 & 53.37 $\pm$ 0.53 \\
 & \checkmark & NGM & 83.91 $\pm$ 0.96 & 78.90 $\pm$ 0.11 & 50.07 $\pm$ 2.79 & 46.73 $\pm$ 4.35 \\
 & & \textit {N-SADDLe (ours)} & \textbf{86.26 $\pm$ 0.29} & \textbf{86.61 $\pm$ 0.20} & \textbf{55.77 $\pm$ 0.53} & \textbf{55.14 $\pm$ 0.49}\\
 & \checkmark & \textit{N-SADDLe (ours)} & \textbf{86.34 $\pm$ 0.24} & \textbf{87.41 $\pm$ 0.52} & \textbf{56.65 $\pm$ 0.17} & \textbf{55.11 $\pm$ 1.16} \\
  \hline
 %%%%%%%%%%%%%%%%%%%
\end{tabular*}
\end{center}
\vspace{-4mm}
\end{table*}

\begin{table*}[ht]
\caption{Test accuracy of NGM, N-SADDLe, and their compressed versions evaluated on ImageNette and ImageNet, distributed over a ring topology. Comp implies 1-bit Sign SGD based compression.}
\vspace{-2mm}
\label{tab:ngmimnet}
\small
\begin{center}
%\resizebox{1.0\columnwidth}{!}{
\begin{tabular*} {\textwidth}{ccl @{\extracolsep{\fill}}*{4}{c}}
\hline
\multirow{ 2}{*}{Agents} &\multirow{2}{*}{Comp} &\multirow{ 2}{*}{Method}& \multicolumn{2}{c}{Imagenette (MobileNet-V2)} & \multicolumn{2}{c}{ImageNet (ResNet-18)}\\
\cline{4-5}  
\cline{6-7}
& & & $\alpha=0.01$ & $\alpha=0.001$ & $\alpha=0.01$ & $\alpha=0.001$\\
 \hline
 % \multirow{4}{*}{$5$} & NGM &  73.72 $\pm$ 0.90 & 73.64 $\pm$ 1.28 & - & -\\
 % & Comp NGM & 74.60 $\pm$ 1.13 & 72.72 $\pm$ 1.02 & - & -\\
 % & \textit {NG-SAM (ours)} & 77.15 $\pm$ 0.17 & 75.51 $\pm$ 0.28 & - & -\\
 % & \textit{Comp NG-SAM (ours)} & 75.20 $\pm$ 0.60 & 75.46 $\pm$ 0.21 & - & - \\
 % \hline
 \multirow{4}{*}{$10$} & & NGM & 67.68 $\pm$ 1.23 & 67.10 $\pm$ 0.71 & 49.30 & 46.38\\
 & \checkmark & NGM & 63.65 $\pm$ 2.70 & 63.70 $\pm$ 0.87 & 34.66 & 34.40 \\
 & & \textit {N-SADDLe (ours)} & \textbf{69.54 $\pm$ 0.33} & \textbf{68.70 $\pm$ 0.79} & \textbf{52.20} & \textbf{51.94} \\
 & \checkmark & \textit{N-SADDLe (ours)} & \textbf{68.09 $\pm$ 0.57} & \textbf{67.37 $\pm$ 0.68} & \textbf{51.44} & \textbf{49.61} \\
  \hline
 %%%%%%%%%%%%%%%%%%%
\end{tabular*}
\end{center}
\vspace{-3mm}
\end{table*}

\vspace{-2mm}

\begin{table}[ht]
\caption{Test accuracy of different decentralized algorithms on CIFAR-10, distributed with $\alpha=0.001$ over torus topology.}
\vspace{-3mm}
\label{tab:torus}
\small
\begin{center}
\begin{tabular}{cccc}
\hline
Agents & Comp & Method & Accuracy(\%)\\ 
 \hline
% \multirow{8}{*}{$20$} & & QGM & 56.07 $\pm$ 3.88 \\
% & \checkmark & QGM & 47.62 $\pm$ 6.75 \\
% & & \textit{Q-SADDLe (ours)} & \textbf{77.10 $\pm$ 1.18} \\
% & \checkmark & \textit{Q-SADDLe (ours)} & \textbf{74.50 $\pm$ 1.31} \\
% \cline{2-4}
% & & NGM & 88.18 $\pm$ 0.17 \\
% & \checkmark & NGM & 87.71 $\pm$ 0.54 \\
% & & \textit{N-SADDLe (ours)} & \textbf{89.21 $\pm$ 0.26} \\
% & \checkmark & \textit{N-SADDLe (ours)} & \textbf{89.30 $\pm$ 0.36} \\
%  \hline

\multirow{8}{*}{$40$} & & QGM & 57.96 $\pm$ 3.90 \\
& \checkmark & QGM & 47.08 $\pm$ 7.72 \\
& & \textit{Q-SADDLe (ours)} & \textbf{70.05 $\pm$ 3.35} \\
& \checkmark & \textit{Q-SADDLe (ours)} & \textbf{65.84 $\pm$ 2.46} \\
\cline{2-4}
& & NGM & 86.00 $\pm$ 0.34 \\
& \checkmark & NGM & 86.30 $\pm$ 0.52 \\
& & \textit{N-SADDLe (ours)} & \textbf{86.67 $\pm$ 0.32} \\
& \checkmark & \textit{N-SADDLe (ours)} & \textbf{87.00 $\pm$ 0.18} \\
\hline
 %%%%%%%%%%%%%%%%%%%
\end{tabular}
\end{center}
\end{table}
\textbf{Performance Comparison:}
As shown in Table \ref{tab:qgmcf10_100}, for CIFAR-10 Q-SADDLe results in 8.4\% better accuracy on average as compared to QGM\cite{qgm} across a range of graph sizes and two different degrees of non-IIDness ($\alpha=0.01, 0.001$). QGM suffers a 1-6\% accuracy drop in the presence of a stochastic quantization-based compression scheme, whereas, for Q-SADDLe, this drop is only 0-1.5\%.
%Comp QGM suffers a 1-6\% accuracy drop as compared to QGM, while Comp Q-SADDLe is only 0-1.5\% lower in test accuracy than Q-SADDLe. 
For a challenging dataset such as CIFAR-100, Table \ref{tab:qgmcf10_100} shows that Q-SADDLe outperforms QGM by $\sim$6\% on average. The accuracy drop due to compression is 1-8\% for QGM, while Q-SADDLe proves to be more resilient to compression error with only a 0-1.8\% drop in accuracy.
We present additional results on Imagenette, a subset of ImageNet trained on MobileNet-v2 in Table \ref{tab:qgmimagenette}. Q-SADDLe leads to an average improvement of $\sim$12\% over QGM, with only a 0-2\% drop in accuracy due to compression. In contrast, QGM incurs a significant drop of 4-9\% in the presence of communication compression. Furthermore, as the degree of non-IIDness is increased from $\alpha=0.01$ to $\alpha=0.001$, QGM suffers from an 8.5\% average drop in accuracy, whereas Q-SADDLe nearly retains the performance. We present additional results for Top-30\% Sparsification in Table \ref{tab:qgmcf10_100_topk} in Appendix.

Table \ref{tab:ngmcf10_100} and \ref{tab:ngmimnet} demonstrate the significant improvements in test accuracy and communication efficiency achieved by N-SADDLe over NGM\cite{ngm}. As shown in Table \ref{tab:ngmcf10_100}, N-SADDLe outperforms NGM by 2.3\% and 4.2\% on average for CIFAR-10 and CIFAR-100, respectively. For CIFAR-10, the accuracy drop due to compression for NGM is $\sim$1-8\%, while it is only about 0-0.3\% for N-SADDLe. Similarly, for CIFAR-100, the drop due to compression for NGM is $\sim$2-11\%, whereas for N-SADDLe, it is only 0-1.7\%. These performance trends are maintained for ImageNette in Table \ref{tab:ngmimnet}, where N-SADDLe outperforms NGM by 1.7\%, with a minimal drop of 1.4\% with compression (as compared to 3.7\% drop in case of NGM). To demonstrate the scalability of our approach, we present additional results on ImageNet distributed over a ring topology of 10 agents with varying degrees of heterogeneity. Our results in Table \ref{tab:ngmimnet} show that N-SADDLe outperforms NGM by 4.2\% while also being much more robust to communication compression. Specifically, NGM incurs a significant drop of 13\% in accuracy, compared to about a 1.5\% drop for N-SADDLe. 

We also evaluate our techniques on a torus graph with 40 agents, and the results are presented in Table \ref{tab:torus}. Q-SADDLe outperforms QGM by 12\%, with only a $\sim$5\% drop in accuracy with compression, whereas QGM experiences a significantly larger drop of $\sim$11\%. N-SADDLe achieves 0.7\% better accuracy than NGM, with both methods maintaining their performance even under 1-bit Sign SGD-based communication compression. Additionally, please refer to Table \ref{tab:ngm_2bitcf10_100} in the Appendix for results on stochastic quantization-based compression for NGM and N-SADDLe. For the exact communication cost of all our presented experiments, please refer to Section \ref{commcost} in the Appendix.
\begin{figure*}[htbp]
\centering
	\begin{subfigure}{0.32\textwidth}
		\includegraphics[width=\textwidth]{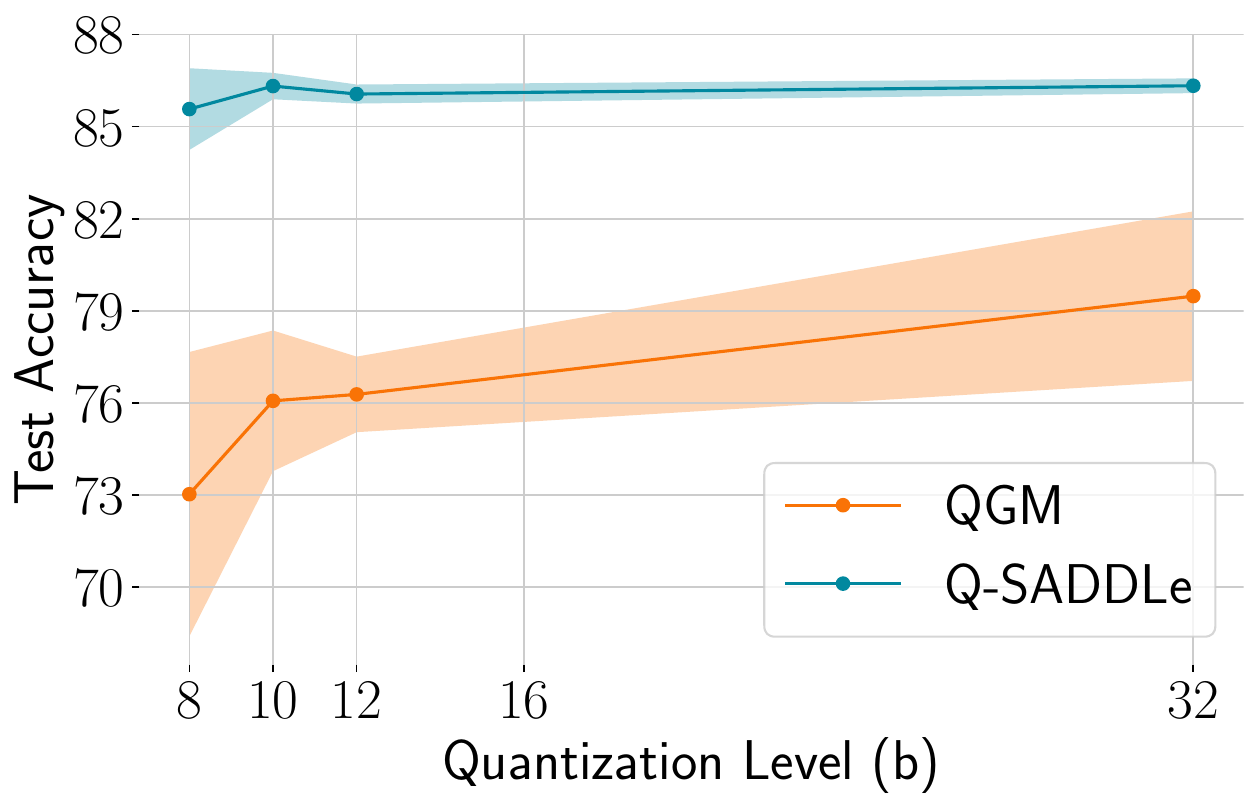}
		\caption{QGM vs Q-SADDLe}
	\end{subfigure}
	\begin{subfigure}{0.32\linewidth}
		\includegraphics[width=\textwidth]{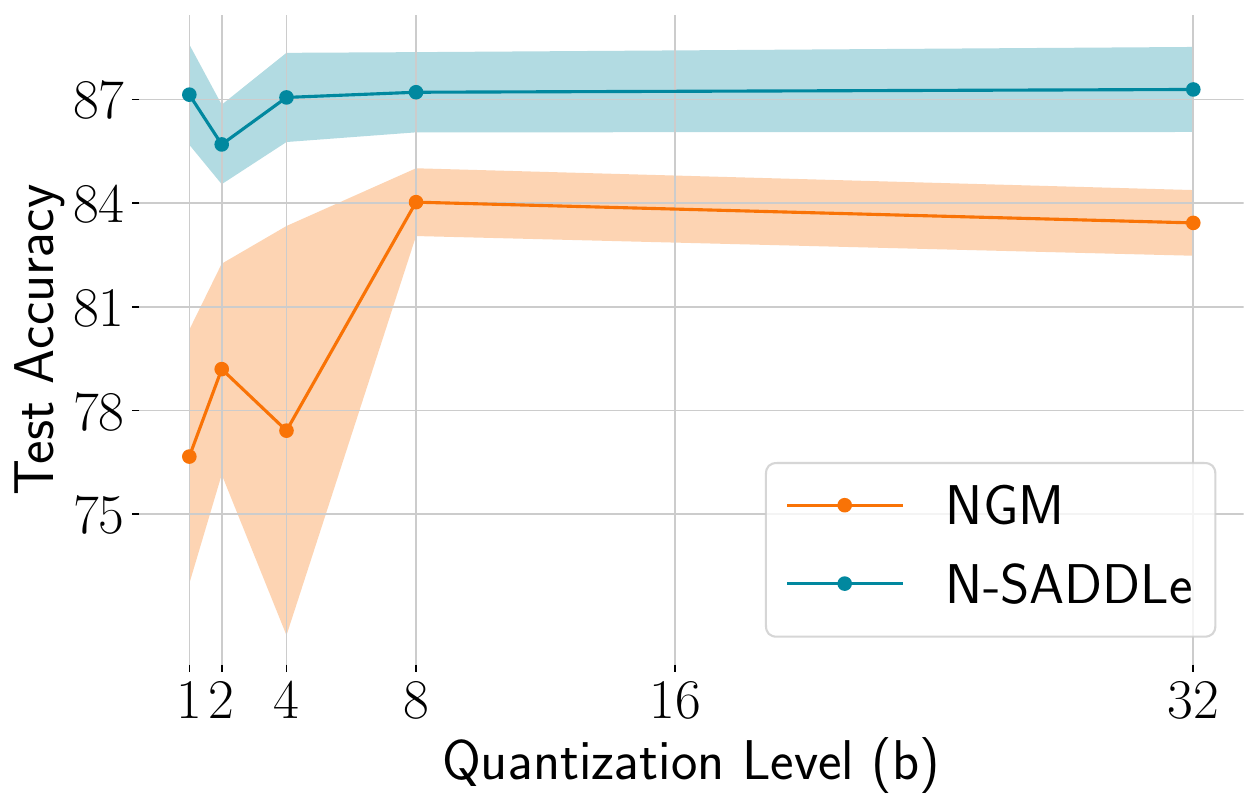}
		\caption{NGM vs N-SADDLe}
	\end{subfigure}
    \vspace{-1mm}
	\caption{Test accuracy for different levels of quantization-based compression scheme for CIFAR-10 over a 10 agent ring topology.}
    %Note that $b=32$ represents the full communication baseline.}
	\label{fig:changingcomp}
\end{figure*}

\textbf{Impact of Varying Compression Levels:} To understand the impact of the degree of compression, we evaluate QGM, Q-SADDLe, NGM, and N-SADDLe for a range of quantization levels and present the results in Figure \ref{fig:changingcomp}. Test accuracy for QGM drops from about 79\% to 73\% as the compression becomes more extreme, while Q-SADDLe retains its performance with a minimal drop of $\sim$ 0.7\%. Similarly, NGM incurs an accuracy drop of about $\sim$7\% due to compression, while N-SADDLe maintains its performance.

\begin{figure}[H]
\centering
    \includegraphics[width=0.47\textwidth]{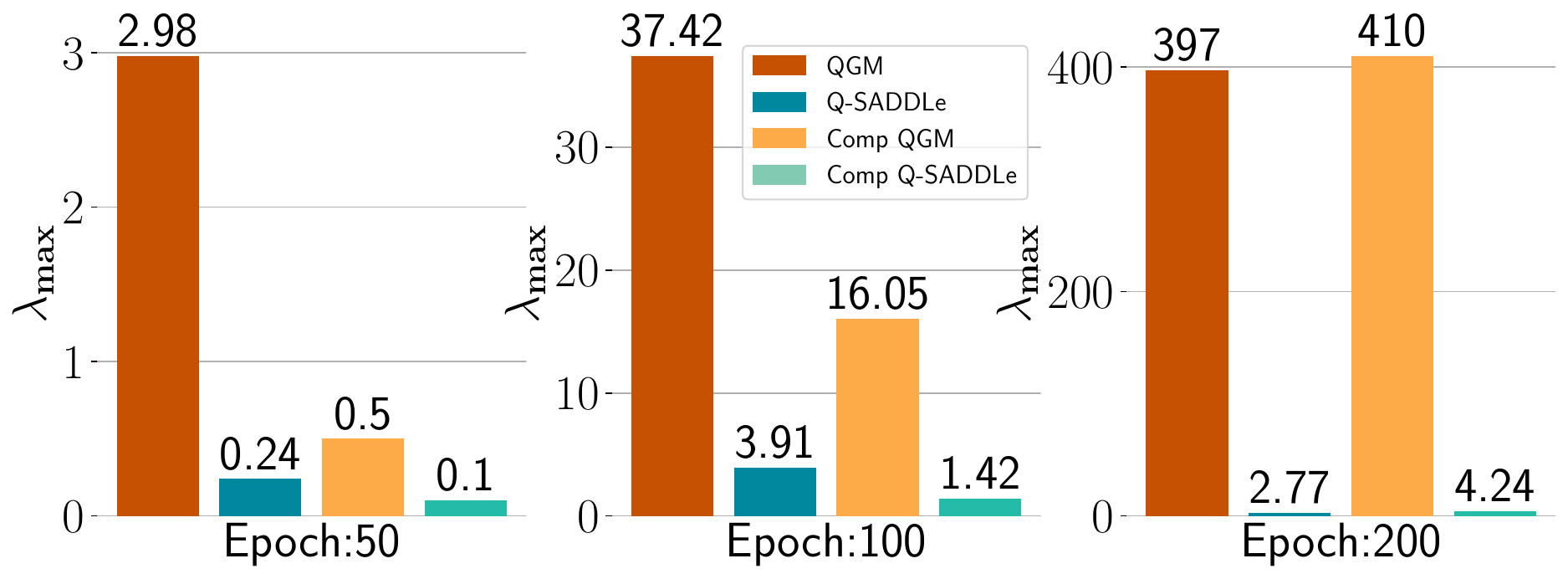}
    \vspace{-0.8mm}
	\caption{Largest Eigenvalue of the Hessian $(\mathbf{\lambda_{max}})$ at 3 stages of training for ResNet-20 trained on CIFAR-10 in a 10 agent ring topology with $\alpha$= 0.001.}
     \label{fig:lambdamax}
     \vspace{-2mm}
\end{figure}
\textbf{Evaluating Flatness Measures:} To confirm our hypothesis that the presence of SAM in decentralized training leads to a flatter loss landscape, we compute the highest eigenvalue $\mathbf{\lambda_{max}}$ of the Hessian at different epochs during the training \cite{hessian-eigenthings}. Note that lower the $\mathbf{\lambda_{max}}$, flatter the loss landscape \cite{jastrzebski,sam,samfree}. As shown in Figure \ref{fig:lambdamax}, Q-SADDLe and Comp Q-SADDLe have consistently lower $\mathbf{\lambda_{max}}$ as compared to QGM and Comp QGM, respectively. This enhances the robustness of Q-SADDLe to erroneous updates due to communication compression. The difference in the eigenvalues is remarkably high towards the end of the training, indicating that models trained with SAM converge to a flatter minimum as expected. Please refer to Figures \ref{fig:apexlossvisualqgm} and \ref{fig:apexlossvisualngm} in the Appendix for loss landscape visualization.

\textbf{Compute-Efficient Variant:} SADDLe seeks flatter loss landscapes through a gradient ascent step, requiring an additional backward pass to compute the perturbation $\xi_i$ (Equation \ref{eq:samupdate}). To reduce this computational overhead, we implement a more efficient variant of SADDLe, where the gradient ascent step is calculated once in every 5 training iterations \cite{esam}, leading to 1.66$\times$ lower compute. As shown in Table \ref{tab:effqsaddle}, this compute-efficient Q-SADDLe variant achieves only 1.75\% lower accuracy compared to the original version but still outperforms QGM by approximately 10\% across two different graph sizes. Even with communication compression, Q-SADDLe nearly maintains its performance, unlike QGM, which incurs a 1-6\% drop (Table \ref{tab:qgmcf10_100}).

\begin{table}[ht]
\caption{Test accuracy of compute-efficient \textit{Q-SADDLe} evaluated on CIFAR-10 distributed over ring topology.}
\vspace{-3mm}
\label{tab:effqsaddle}
\small
\begin{center}
\begin{tabular}{cccc}
\hline
\multirow{ 2}{*}{Agents} & \multirow{ 2}{*}{Comp} & \multicolumn{2}{c}{Accuracy(\%)}\\
\cline{3-4}  
& &  $\alpha=0.01$ & $\alpha=0.001$\\
 \hline
\multirow{2}{*}{$10$} & & 85.50 $\pm$ 0.26 & 84.27 $\pm$ 0.15\\
& \checkmark & 84.79 $\pm$ 0.34 & 84.61 $\pm$ 0.23\\
 \hline

\multirow{2}{*}{$20$} & & 82.77 $\pm$ 1.38 & 80.07 $\pm$ 1.34\\
& \checkmark & 82.41 $\pm$ 0.86 & 80.14 $\pm$ 1.67\\
\hline
 %%%%%%%%%%%%%%%%%%%
 \vspace{-6mm}
\end{tabular}
\end{center}
\end{table}

\section{Conclusion}
Communication-efficient decentralized learning on heterogeneous data is crucial for enabling on-device learning to leverage vast amounts of user-generated data. In this work, we propose Sharpness-Aware Decentralized Deep Learning (SADDLe) to improve generalization and robustness to communication compression in the presence of data heterogeneity. SADDLe aims to seek a flatter loss landscape during training through gradient perturbation via SAM\cite{sam}. Our theoretical analysis shows that SADDLe achieves a convergence rate comparable to well-known decentralized convergence bounds \cite{dpsgd}. The proposed technique is complementary to existing decentralized learning algorithms and can be used synergistically to improve performance. We present two versions of our approach, Q-SADDLe, and N-SADDLe, and conduct exhaustive experiments to evaluate these techniques over various datasets, models, graphs, and compression schemes. Our results show that SADDLe leads to 1-20\% better accuracy than existing decentralized algorithms for non-IID data, with a minimal drop of $\sim$1\% in the presence of up to 4$\times$ communication compression.

\textbf{Acknowledgements.} This work was supported by the Center for the Co-Design of Cognitive Systems (COCOSYS), an SRC/ DARPA-sponsored JUMP 2.0 center.

%%%%%%%%% REFERENCES
{\small
\bibliographystyle{ieee_fullname}
\bibliography{egbib}
}
\clearpage
\maketitle  % Ensure this is AFTER \begin{document}
% \documentclass[10pt,twocolumn,letterpaper]{article}

%%%%%%%% PAPER TYPE  - PLEASE UPDATE FOR FINAL VERSION
% \usepackage[review,algorithms]{wacv}      % To produce the REVIEW version for the algorithms track
% \usepackage[review,applications]{wacv}      % To produce the REVIEW version for the applications track
% \usepackage{wacv}              % To produce the CAMERA-READY version
% \usepackage[pagenumbers]{wacv} % To force page numbers, e.g. for an arXiv version

% Include other packages here, before hyperref.
% \usepackage{graphicx}
% \usepackage{amsmath}
% \usepackage{amssymb}
% \usepackage{booktabs}
% \usepackage{algorithm}
% \newtheorem{theorem}{Theorem}
% \newtheorem{corollary}[theorem]{Corollary}
% \newtheorem{lemma}[theorem]{Lemma}

% \usepackage{multirow}
% \usepackage{subcaption}
% \usepackage{tikz}

% \usepackage{xcolor}         % colors
% \definecolor{mintbg}{rgb}{.63,.79,.95}
% \definecolor{apricot}{rgb}{0.98, 0.81, 0.69}
% \definecolor{asparagus}{rgb}{0.53, 0.66, 0.42}
% \definecolor{bondiblue}{rgb}{0.0, 0.58, 0.71}
% \colorlet{lightmintbg}{bondiblue!50}
% \colorlet{lightapricot}{apricot!100}

% \newcommand{\norm}[1]{\left\lVert#1\right\rVert}
\providecommand{\E}{{\mathbb E}}

\providecommand{\lin}[1]{\ensuremath{\left\langle #1 \right\rangle}}
\providecommand{\abs}[1]{\left\lvert#1\right\rvert}
\providecommand{\norm}[1]{\left\lVert#1\right\rVert}

\providecommand{\refLE}[1]{\ensuremath{\stackrel{(\ref{#1})}{\leq}}}
\providecommand{\refEQ}[1]{\ensuremath{\stackrel{(\ref{#1})}{=}}}
\providecommand{\refGE}[1]{\ensuremath{\stackrel{(\ref{#1})}{\geq}}}
\providecommand{\refID}[1]{\ensuremath{\stackrel{(\ref{#1})}{\equiv}}}

\providecommand{\R}{\mathbb{R}} %
\providecommand{\N}{\mathbb{N}} %

\providecommand{\E}{{\mathbb E}}
\providecommand{\E}[1]{{\mathbb E}\left.#1\right. }        %
\providecommand{\Eb}[1]{{\mathbb E}\left[#1\right] }       %
\providecommand{\EE}[2]{{\mathbb E}_{#1}\left.#2\right. }  %
\providecommand{\EEb}[2]{{\mathbb E}_{#1}\left[#2\right] } %
\providecommand{\prob}[1]{{\rm Pr}\left[#1\right] }
\providecommand{\Prob}[2]{{\rm Pr}_{#1}\left[#2\right] }
\providecommand{\P}[1]{{\rm Pr}\left.#1\right. }
\providecommand{\Pb}[1]{{\rm Pr}\left[#1\right] }
\providecommand{\PP}[2]{{\rm Pr}_{#1}\left[#2\right] }
\providecommand{\PPb}[2]{{\rm Pr}_{#1}\left[#2\right] }
\providecommand{\derive}[2]{\frac{{\partial}{#1}}{\partial #2} }  %

\providecommand{\0}{\mathbf{0}}
\providecommand{\1}{\mathbf{1}}
\renewcommand{\aa}{\mathbf{a}}
\providecommand{\bb}{\mathbf{b}}
\providecommand{\cc}{\mathbf{c}}
\providecommand{\dd}{\mathbf{d}}
\providecommand{\ee}{\mathbf{e}}
\providecommand{\ff}{\mathbf{f}}
\let\ggg\gg
\renewcommand{\gg}{\mathbf{g}}
\providecommand{\gv}{\mathbf{g}}
\providecommand{\hh}{\mathbf{h}}
\providecommand{\ii}{\mathbf{i}}
\providecommand{\jj}{\mathbf{j}}
\providecommand{\kk}{\mathbf{k}}
\let\lll\ll
\renewcommand{\ll}{\mathbf{l}}
\providecommand{\mm}{\mathbf{m}}
\providecommand{\nn}{\mathbf{n}}
\providecommand{\oo}{\mathbf{o}}
\providecommand{\pp}{\mathbf{p}}
\providecommand{\qq}{\mathbf{q}}
\providecommand{\rr}{\mathbf{r}}
\renewcommand{\ss}{\mathbf{s}}
\providecommand{\tt}{\mathbf{t}}
\providecommand{\uu}{\mathbf{u}}
\providecommand{\vv}{\mathbf{v}}
\providecommand{\ww}{\mathbf{w}}
\providecommand{\xx}{\mathbf{x}}
\providecommand{\yy}{\mathbf{y}}
\providecommand{\zz}{\mathbf{z}}

\providecommand{\mA}{\mathbf{A}}
\providecommand{\mB}{\mathbf{B}}
\providecommand{\mC}{\mathbf{C}}
\providecommand{\mD}{\mathbf{D}}
\providecommand{\mE}{\mathbf{E}}
\providecommand{\mF}{\mathbf{F}}
\providecommand{\mG}{\mathbf{G}}
\providecommand{\mH}{\mathbf{H}}
\providecommand{\mI}{\mathbf{I}}
\providecommand{\mJ}{\mathbf{J}}
\providecommand{\mK}{\mathbf{K}}
\providecommand{\mL}{\mathbf{L}}
\providecommand{\mM}{\mathbf{M}}
\providecommand{\mN}{\mathbf{N}}
\providecommand{\mO}{\mathbf{O}}
\providecommand{\mP}{\mathbf{P}}
\providecommand{\mQ}{\mathbf{Q}}
\providecommand{\mR}{\mathbf{R}}
\providecommand{\mS}{\mathbf{S}}
\providecommand{\mT}{\mathbf{T}}
\providecommand{\mU}{\mathbf{U}}
\providecommand{\mV}{\mathbf{V}}
\providecommand{\mW}{\mathbf{W}}
\providecommand{\mX}{\mathbf{X}}
\providecommand{\mY}{\mathbf{Y}}
\providecommand{\mZ}{\mathbf{Z}}
\providecommand{\mz}{\mathbf{z}}
\providecommand{\mLambda}{\boldsymbol{\Lambda}}
\providecommand{\mpi}{\boldsymbol{\pi}}
\providecommand{\malpha}{\boldsymbol{\alpha}}
\providecommand{\meta}{\boldsymbol{\eta}}
\providecommand{\mxi}{\boldsymbol{\xi}}
\providecommand{\mepsilon}{\boldsymbol{\epsilon}}
\providecommand{\mphi}{\boldsymbol{\phi}}

\providecommand{\cA}{\mathcal{A}}
\providecommand{\cB}{\mathcal{B}}
\providecommand{\cC}{\mathcal{C}}
\providecommand{\cD}{\mathcal{D}}
\providecommand{\cE}{\mathcal{E}}
\providecommand{\cF}{\mathcal{F}}
\providecommand{\cG}{\mathcal{G}}
\providecommand{\cH}{\mathcal{H}}
\providecommand{\cI}{\mathcal{I}}
\providecommand{\cJ}{\mathcal{J}}
\providecommand{\cK}{\mathcal{K}}
\providecommand{\cL}{\mathcal{L}}
\providecommand{\cM}{\mathcal{M}}
\providecommand{\cN}{\mathcal{N}}
\providecommand{\cO}{\mathcal{O}}
\providecommand{\cP}{\mathcal{P}}
\providecommand{\cQ}{\mathcal{Q}}
\providecommand{\cR}{\mathcal{R}}
\providecommand{\cS}{\mathcal{S}}
\providecommand{\cT}{\mathcal{T}}
\providecommand{\cU}{\mathcal{U}}
\providecommand{\cV}{\mathcal{V}}
\providecommand{\cX}{\mathcal{X}}
\providecommand{\cY}{\mathcal{Y}}
\providecommand{\cW}{\mathcal{W}}
\providecommand{\cZ}{\mathcal{Z}}
% Support for easy cross-referencing
% \usepackage[capitalize]{cleveref}
% \crefname{section}{Sec.}{Secs.}
% \Crefname{section}{Section}{Sections}
% \Crefname{table}{Table}{Tables}
% \crefname{table}{Tab.}{Tabs.}

%%%%%%%%% PAPER ID  - PLEASE UPDATE
% \def\wacvPaperID{2101} % *** Enter the WACV Paper ID here
% \def\confName{WACV}
% \def\confYear{2025}
\setcounter{section}{0}  % Restart numbering
% \setcounter{equation}{0}  % Restart numbering
% \setcounter{lemma}{0}  % Restart lemma numbering

% \begin{document}
% \title{Supplementary Material - SADDLe: Sharpness-Aware Decentralized Deep Learning with Heterogeneous Data}
% \maketitle
\section*{Appendix}
\section{Theoretical Analysis}\label{appendix:dsam}
The update rule for Q-SADDLe with SAM-based gradient $\widetilde{\mathbf{G}}$ is as follows:
\begin{equation} \label{eq:appendix_our_scheme_matrix_form}
	\begin{split}
		\mathbf{X}^{(t+1)} 	&= \mathbf{W} \left( \mX^{(t)} - \eta \left( \beta \mathbf{M}^{(t)} + \widetilde{\mathbf{G}}^{(t)} \right) \right) \\
		\mathbf{M}^{(t + 1)}   &= \mu \mathbf{M}^{(t)} + (1 - \mu) \frac{ \mathbf{X}^{(t)} - \mathbf{X}^{(t + 1)} }{\eta} \\
		&= \left( \mu + (1 - \mu) \beta \mW \right) \mM^{(t)} + (1-\mu)\mW\widetilde{\mG}^{(t)}\\
        &+ \frac{1 - \mu}{\eta}(\mI-\mW)\mX^{(t)} \,,
	\end{split}
\end{equation}

For a doubly stochastic mixing matrix $\mW$, we can simplify the updates as follows:
\begin{equation} \label{eq:appendix_our_scheme_averaged_form}
	\begin{split}
		\bar\xx^{(t+1)} 	&= \bar\xx^{(t)} - \eta \left( \beta \bar\mm^{(t)} + \frac{1}{n}\sum_{i=1}^n \widetilde{\mathbf{g}}_i^t \right) \text{,} \\
		\bar \mm^{(t + 1)}   &= \mu \bar \mm^{(t)} + (1 - \mu) \frac{ \bar \xx^{(t)} - \bar\xx^{(t + 1)} }{\eta}\\
        &= (1 - (1 - \mu)(1- \beta)) \bar \mm^{(t)} + (1 - \mu)\frac{1}{n}\sum_{i=1}^n \widetilde{\mathbf{g}}_i^t \,.
	\end{split}
\end{equation}

Here, $\widetilde{\mathbf{g}}_i^t$ is the SAM-based gradient update, which we reiterate for ease of understanding :
\begin{equation}
    \widetilde{\mathbf{g}}_i^{t} =\nabla F_i(\mathbf{x}_i^{t}+ \xi(\mathbf{x}_i^{t}); d_i^t), \hspace{2mm} \text{where} \hspace{2mm} \xi(\mathbf{x}_i^{t})= \rho \frac{\mathbf{g}_i^t}{\|\mathbf{g}_i^t\|}
\end{equation}
For the rest of the analysis, we use $\xi(\mathbf{x}_i^{t})= \xi_i^t$ for simplicity of notation. We introduce the following lemma to define an upper bound on the stochastic variance of SAM-based updates.
\begin{lemma}\label{lemma:stochasvar}
    Given assumptions 1-3, the stochastic variance of local gradients with perturbation can be bounded as
    \begin{equation}
        \|\nabla F_i(\mathbf{x}_i+\xi_i)- \nabla f_i(\mathbf{x}_i+\xi_i)\|^2 \leq 3\sigma^2+6L^2\rho^2
    \end{equation}
\end{lemma}
\textit{Proof:} \begin{equation}\label{eq:stochasvar_bound}
\begin{split}
    &\|\nabla F_i(\mathbf{x}_i+ \xi_i)- \nabla f_i(\mathbf{x}_i+ \xi_i)\|^2= \\
    &\| \nabla F_i(\mathbf{x}_i+ \xi_i)- \nabla F_i(\mathbf{x}_i)+ \nabla F_i(\mathbf{x}_i)- \nabla f_i(\mathbf{x}_i) +\nabla f_i(\mathbf{x}_i)\\
    &- \nabla f_i(\mathbf{x}_i+ \xi_i)\|^2
    \overset{a}{\leq} 3 \| \nabla F_i(\mathbf{x}_i+ \xi_i)- \nabla F_i(\mathbf{x}_i) \|^2\\
    &+ 3 \| \nabla F_i(\mathbf{x}_i)- \nabla f_i(\mathbf{x}_i)\|^2 + 3 \|  \nabla f_i(\mathbf{x}_i)- \nabla f_i(\mathbf{x}_i+ \xi_i) \|^2\\
    & \overset{b}{\leq} 3 \| \nabla F_i(\mathbf{x}_i+ \xi_i)- \nabla F_i(\mathbf{x}_i) \|^2 + 3 \sigma^2\\
    &+ 3 \|  \nabla f_i(\mathbf{x}_i)- \nabla f_i(\mathbf{x}_i+ \xi_i) \|^2 \overset{c}{\leq} 3\sigma^2 + 6L^2\rho^2
\end{split}
\end{equation}

(a) follows from the property $\| x_1+x_2+...x_n\|^2 \leq n[\|x_1\|^2+\|x_2\|^2...\|x_n\|^2]$ for random variables $x_1,x_2,...x_n$. (b) follows from Assumption 2 in the main paper. (c) follows from Assumption 1 and the perturbation $\xi_i$ being bounded by the perturbation radius $\rho$.

\begin{lemma}\label{lemma:gtilde_bound}
Given assumptions 1-3 and $\widetilde{\mathbf{g}}_i= \nabla F_i(\mathbf{x}_i+ \xi_i)$, the following relationship holds
\begin{equation}
    \mathbb{E} \| [\frac{1}{n} \sum_{i=1}^{n} \widetilde{\mathbf{g}}_i\|^2] \leq \frac{3\sigma^2}{n}+ \frac{6L^2\rho^2}{n}+
    \mathbb{E}[\| \frac{1}{n} \sum_{i=1}^{n} \nabla f_i(\mathbf{x}_i+\xi_i)\|^2]
\end{equation}
\end{lemma}

\textit{Proof:}
\begin{equation}
    \begin{split}
        &\mathbb{E} [\|\frac{1}{n} \sum_{i=1}^{n} \widetilde{\mathbf{g}}_i\|^2] = \mathbb{E} [\|\frac{1}{n} \sum_{i=1}^{n} \widetilde{\mathbf{g}}_i- \nabla f_i(\mathbf{x}_i+\xi_i)\|^2]\\
        &+ \mathbb{E} [\|\frac{1}{n} \sum_{i=1}^{n} \nabla f_i(\mathbf{x}_i+\xi_i)\|^2]\\
        &= \frac{1}{n^2}\|\sum_{i=1}^{n} \mathbb{E} [\| \widetilde{\mathbf{g}}_i- \nabla f_i(\mathbf{x}_i+\xi_i)\|^2]\\
        &+ \mathbb{E} [\|\frac{1}{n} \sum_{i=1}^{n} \nabla f_i(\mathbf{x}_i+\xi_i)\|^2]
        \leq \frac{3\sigma^2}{n}+\frac{6L^2\rho^2}{n}\\
        &+ \mathbb{E} [\|\frac{1}{n} \sum_{i=1}^{n} \nabla f_i(\mathbf{x}_i+\xi_i)\|^2]
    \end{split}
\end{equation}

As a first step, we simplify our convergence analysis by defining another sequence of parameters ${\mz^{(t)}}$ with the following update rule:
\begin{equation}\label{eq:appendix_virtual_sequence}
	\begin{split}
		\mz^{(t+1)} &= \mz^{(t)} - \bigg(\frac{\eta}{1-\beta}\bigg)\frac{1}{n}\sum_{i=1}^n \widetilde{\mathbf{g}}_i^t
	\end{split}
\end{equation}

Inspired by QGM \cite{qgm}, this sequence has a simpler SAM update rule, while our parameters ${\bar \xx^{(t)}}$ follow SAM-based gradient updates along with a momentum buffer $\mathbf{m}_i^t$.
We use $\bar\gg^{(t)}= \frac{1}{n}\sum_{i=1}^n \widetilde{\mathbf{g}}_i^t$ and $\tilde \eta=\frac{\eta}{1-\beta}$ for rest of the analysis. We begin by proving that the error $\ee^{(t)} = \mz^{(t)} - \bar\xx^{(t)}$ remains bounded.

\begin{lemma}\label{lem:error-sequence}
	Given Assumptions 1-3, the sequence of iterates generated by Q-SADDLe satisfy
	\begin{equation*}
    \begin{split}
		&\E\norm{\ee^{(t+1)}}^2 \leq (1- (1-\mu)(1-\beta))\E\norm{\ee^{(t)}}^2 + \\
  &\frac{2\tilde\eta^2 \beta^2}{(1-\beta) (1-\mu)} \E\norm{\frac{1}{n}\sum_{i=1}^n \nabla f_i(\mathbf{x}_i+\xi_i)}^2\\
       &+ 3\tilde\eta^2\beta^2 \sigma^2 + 6\tilde \eta^2\beta^2L^2\rho^2\,.
       \end{split}
	\end{equation*}
\end{lemma}

\textit{Proof}: For $\ee^{(0)}=0$, specifying $\ee^{(t+1)}$ in terms of update sequences $\mz^{(t+1)}$ and $\bar\xx^{(t+1)}$:
\begin{equation}
\begin{split}
    &\ee^{(t+1)} = \mz^{(t+1)} - \bar\xx^{(t+1)}= \left( \mz^{(t)} - \frac{\eta}{1 - \beta}\bar\gg^{(t)} \right) - (\bar\xx^{(t)} - \\
    &\eta (\beta\bar\mm^{(t)} +\bar\gg^{(t)}))= \ee^{(t)} - \eta\beta(\frac{1}{1-\beta}\bar\gg^{(t)} -\bar\mm^{(t)})\\
    &= \sum_{k=0}^{t} -\eta\beta(\frac{1}{1-\beta}\bar\gg^{(k)} - \bar\mm^{(k)})\,.
\end{split}
\end{equation}

Using equation \eqref{eq:appendix_our_scheme_averaged_form}, we have \cite{qgm}:
	\begin{equation}
    \begin{split}
		&\ee^{(t+1)} = \sum_{k=0}^{(t)} -\eta\beta ( \frac{1}{1-\beta}\bar\gg^{(k)} - ( (1- (1-\mu)(1-\beta))\\
        &\bar\mm^{(k-1)}+(1-\mu)\bar\gg^{(k-1)})) = (1- (1-\mu)(1-\beta))\\
        &\sum_{k=0}^t -\eta\beta \left( \frac{1}{1-\beta}\bar\gg^{(k-1)} - \bar\mm^{(k-1)} \right)
		+ \sum_{k=0}^{(t)} -\frac{\eta\beta}{1-\beta} (\bar\gg^{(k)} -\\
        &\bar\gg^{(k-1)})= (1- (1-\mu)(1-\beta))\ee^{(t)} - \frac{\eta\beta}{1-\beta}\bar\gg^{(t)} \,.
    \end{split}
	\end{equation}

Taking expectation of $\|\ee^{(t+1)}\|^2$:
	\begin{equation}
    \begin{split}
		&\E\norm{\ee^{(t+1)}}^2 = \E\norm{(1- (1-\mu)(1-\beta))\ee^{(t)} - \frac{\eta\beta}{1-\beta}\bar\gg^{(t)}}^2\\
	  & \overset{a}{\leq} \E\norm{(1- (1-\mu)(1-\beta))\ee^{(t)} - \frac{\eta\beta}{1-\beta}\E_t[\bar\gg^{(t)}]}^2 + \\
     &\bigg(\frac{\eta^2\beta^2}{(1-\beta)^2}\bigg)(3\sigma^2+6L^2\rho^2)\leq (1- (1-\mu)(1-\beta))\E\norm{\ee^{(t)}}^2\\
     &+\frac{2\tilde \eta^2 \beta^2}{(1-\beta) (1-\mu)}\E\norm{\frac{1}{n}\sum_{i=1}^n \nabla f_i(\mathbf{x}_i+\xi_i)}^2
    + 3\tilde \eta^2\beta^2\sigma^2+\\
    &6\tilde \eta^2\beta^2L^2\rho^2 \,. 
    \end{split}
	\end{equation}
(a) is the result of Lemma~\ref{lemma:stochasvar}.

We now proceed to bound the consensus error.
\begin{lemma}\label{lem:consensus-change}
	Given Assumptions 1-3, the sequence of iterates generated by Q-SADDLe satisfy,
	\begin{equation}
    \begin{split}
		&\frac{1}{n}\E\norm{\mX^{t+1} - \bar\mX^{t+1}}^2 \leq \frac{(1 - \lambda/4)}{n}\E\norm{\mX^{t} - \bar\mX^{t}}^2
		+\\
        &\frac{24\eta^2L^2\rho^2}{\lambda}+\frac{12\eta^2\delta^2}{\lambda}+12\eta^2(1-\lambda)(\sigma^2 + 2L^2\rho^2)+\\
        &\frac{6\eta^2\beta^2}{\lambda n}\E\norm{\mM^{(t)} - \bar\mM^{(t)}}^2 \,.
 \end{split}
 \end{equation}
\end{lemma}

\textit{Proof:} We start by describing $\mX^{t+1}$ and $\bar\mX^{t+1}$ in terms of the update rule in equation \ref{eq:appendix_our_scheme_matrix_form}:
	\begin{equation}\label{eq:lemma2main}
    \begin{split}
		&\frac{1}{n}\E\norm{\mX^{t+1} - \bar\mX^{t+1}}^2=\frac{1}{n} \E\|\!\mW ( \mX^{(t)} - \eta ( \beta \mM^{(t)} + \widetilde{\mG}^{(t)}))\\
        &- ( \bar\mX^{(t)} - \eta (\beta \bar\mM^{(t)} + \bar\mG^{(t)}) )\!\|^2 \overset{a}{\leq} \frac{1 - \lambda}{n}\E\|\!( \mX^{(t)}- \eta ( \beta \mM^{(t)} \\
        &+ \widetilde{\mG}^{(t)} ) ) - ( \bar\mX^{(t)} - \eta ( \beta \bar\mM^{(t)} + \bar\mG^{(t)}))\!\|^2 \overset{b}
        {\leq} \frac{1 - \lambda}{n}\E\|\!( \mX^{(t)}\\
        &- \eta( \beta \mM^{(t)} + \E[\widetilde{\mG}^{(t)}])) -( \bar\mX^{(t)} - \eta ( \beta \bar\mM^{(t)} + \E[\bar\mG^{(t)}]))\!\|^2 \\
        &+ 12\eta^2(1-\lambda)(\sigma^2 + 2L^2\rho^2)\leq \frac{(1 - \lambda)(1+\lambda/2)}{n}\\
        &\E\| \mX^{(t)} - \bar\mX^{(t)}\|^2
        + \frac{6\eta^2\beta^2}{\lambda n}\E\|\mM^{(t)} - \bar\mM^{(t)}\|^2 + \frac{6\eta^2}{\lambda n}\\
        &\underbrace{\E\| \E_t[\widetilde{\mG}^{(t)}] - \E_t[\bar\mG^{(t)}]\|^2}_{\star} + 12\eta^2(1-\lambda)(\sigma^2 + 2L^2\rho^2)\,.
    \end{split}
	\end{equation}
 (a) comes from Assumption 3 on the Mixing matrix.
 (b) results from $\widetilde{\mG}^{(t)}= \E_t[\widetilde{\mG}^{(t)}]+\widetilde{\mG}^{(t)}-\E_t[\widetilde{\mG}^{(t)}]$ and Lemma~\ref{lemma:stochasvar}.

 We first analyze $\star$:
 \begin{equation}\label{eq:star}
     \begin{split}
         &\E\| \E_t[\widetilde{\mG}^{(t)}] - \E_t[\bar\mG^{(t)}]\|^2= \sum_{i=1}^{n} \E\|\! \nabla f_i(\xx_i^{(t)}+\xi_i^{(t)}) \pm \\
         &\nabla f_i(\bar{\mathbf{x}}^{(t)})- \nabla f(\bar{\mathbf{x}}^{(t)})\!\|^2
         \leq 2\sum_{i=1}^{n} \E \|\!\nabla f_i(\xx_i^{(t)}+\xi_i^{(t)})-\\
         &\nabla f_i(\bar{\mathbf{x}}^{(t)})\!\|^2
         + 2\sum_{i=1}^{n} \E \norm{\nabla f_i(\bar{\mathbf{x}}^{(t)})- \nabla f(\bar{\mathbf{x}}^{(t)})}^2\\
         &\overset{a}{\leq} 2L^2\sum_{i=1}^{n} \E \norm{\xx_i^{(t)}+\xi_i^{(t)}-\bar{\mathbf{x}}^{(t)}}^2+2n\delta^2\\
         &\overset{b}{\leq} 4L^2 \sum_{i=1}^{n} \E \norm{\xx_i^{(t)}-\bar{\mathbf{x}}^{(t)}}^2+4nL^2\rho^2+2n\delta^2
     \end{split}
 \end{equation}

(a) follows from Assumption 1, and (b) is the result of perturbation being bounded by the perturbation radius $\rho$.

Substituting the result of equation \ref{eq:star} in \ref{eq:lemma2main}:
 % \begin{equation}
 %     \begin{split}
 %     &\frac{1}{n}\E\norm{\mX^{t+1} - \bar\mX^{t+1}}^2 \leq \frac{(1 - \lambda)(1+\lambda/2)}{n}\E\norm{\mX^{(t)} - \bar\mX^{(t)}}^2 + \frac{6\eta^2\beta^2}{\lambda n}\E\norm{\mM^{(t)} - \bar\mM^{(t)}}^2\\                                                                        &+\frac{6\eta^2}{\lambda n} \bigg(4L^2 \sum_{i=1}^{n} \E \norm{\xx_i^{(t)}-\bar{\mathbf{x}}^{(t)}}^2+4nL^2\rho^2+2n\delta^2\bigg)+ 12\eta^2(1-\lambda)(\sigma^2 + 2L^2\rho^2)
 %    \end{split}
 % \end{equation}

\begin{equation}
 \begin{split}
 &\frac{1}{n}\E\norm{\mX^{t+1} - \bar\mX^{t+1}}^2 \leq \frac{(1 - \lambda/2)}{n}\E\norm{\mX^{(t)} - \bar\mX^{(t)}}^2 +\\
 &\frac{6\eta^2\beta^2}{\lambda n}\E\norm{\mM^{(t)} - \bar\mM^{(t)}}^2+\frac{24\eta^2L^2}{\lambda n} \bigg(\sum_{i=1}^{n} \E \norm{\xx_i^{(t)}-\bar{\mathbf{x}}^{(t)}}^2\bigg)\\
 &+ \frac{24\eta^2L^2\rho^2}{\lambda}+\frac{12\eta^2\delta^2}{\lambda}
 +12\eta^2(1-\lambda)(\sigma^2 + 2L^2\rho^2)
\end{split}
\end{equation}

The assumption that learning rate $\eta \leq \frac{\lambda}{10L}$ ensures that $24\eta^2 L^2 \leq \lambda^2/4$. Modifying the above equation through this and rearranging the terms we have:
\begin{equation}
 \begin{split}
 &\frac{1}{n}\E\norm{\mX^{t+1} - \bar\mX^{t+1}}^2 \leq \frac{(1 - \lambda/4)}{n}\E\norm{\mX^{(t)} - \bar\mX^{(t)}}^2\\
 &+ \frac{6\eta^2\beta^2}{\lambda n}\E\norm{\mM^{(t)} - \bar\mM^{(t)}}^2 + \frac{24\eta^2L^2\rho^2}{\lambda}+\frac{12\eta^2\delta^2}{\lambda}\\
 &+12\eta^2(1-\lambda)(\sigma^2 + 2L^2\rho^2)
\end{split}
\end{equation}

In the above bound on the consensus error, we have a momentum error term $\E\norm{\mM^{(t)} - \bar\mM^{(t)}}^2$. We present the following lemma to provide an upper bound on this error :
%Momentum update bounds
% \vspace{-1.8mm}
\begin{lemma}\label{lem:momentum}
	Given Assumptions 1-3, the sequence of iterates generated by Q-SADDLe
	for $\frac{\beta}{1 - \beta} \leq \frac{\lambda}{21}$,
	\begin{equation*}
    \begin{split}
		&\frac{6\eta^2\beta^2}{n\lambda(1-\mu)(1-\beta)}\E\norm{\mM^{t+1} - \bar\mM^{t+1}}^2\\ &\leq \left(\! \frac{6\eta^2\beta^2}{n\lambda(1-\mu)(1-\beta)} - \frac{6\eta^2\beta^2}{n\lambda} \!\right)\E\norm{(\mM^{(t)} -\bar\mM^{(t)})}^2 \\ +  &\frac{\lambda}{8n}\E\norm{\mX^{(t)} - \bar\mX^{(t)}}^2 + \frac{\lambda\eta^2 \delta^2 }{8} +
        \bigg(\frac{3(1-\beta)}{(1-\mu)}+\frac{1}{2}\bigg)\\
        &\frac{\lambda\eta^2L^2\rho^2}{4}+\frac{\lambda\eta^2 \sigma^2 (1-\beta)}{8(1-\mu)}\,.
    \end{split}
	\end{equation*}
\end{lemma}
% \vspace{-2mm}
\textit{Proof:} Starting from the update \eqref{eq:appendix_our_scheme_matrix_form}, we have:
% \vspace{-10mm}
\begin{equation}
    \begin{split}
        &\frac{1}{n}\E\norm{\mM^{(t+1)} - \bar\mM^{(t+1)}}^2= \frac{1}{n}\E\| \left( \mu\mI + (1 - \mu) \beta \mW \right)  (\mM^{(t)}\\
        &-\bar\mM^{(t)}) + (1-\mu)\mW(\widetilde{\mG}^{(t)}- \bar\mG^{(t)})+ \frac{1 - \mu}{\eta}(\mI-\mW)\mX^{(t)}\|^2\\
        &=\frac{1}{n}\E\|\!\left( \mu\mI + (1 - \mu) \beta \mW \right)  (\mM^{(t)} -\bar\mM^{(t)})+ \frac{1 - \mu}{\eta}(\mI - \\
        &\mW) \mX^{(t)}+ (1 - \mu) \mW( \E[\widetilde{\mG}^{(t)} - \bar\mG^{(t)}] ) \!\|^2 \\
		  & + \frac{1}{n} \E \|\!
			(1-\mu) \mW \left( \widetilde{\mG}^{(t)} - \E[\widetilde{\mG}^{(t)}]
   - ( \bar\mG^{(t)} - \E[\bar\mG^{(t)}] ) \right)
		\!\|^2\\
		  & \overset{a}{\leq} \frac{1}{n}\E\|\!
			\left( \mu\mI + (1 - \mu) \beta \mW \right)  (\mM^{(t)} -\bar\mM^{(t)})
			+ \frac{1 - \mu}{\eta}(\mI -\\
            &\mW) \mX^{(t)}
			+ (1 - \mu) \mW( \E[\widetilde{\mG}^{(t)} - \bar\mG^{(t)}] )\!\|^2+ 4 (3\sigma^2 + 6L^2\rho^2)\\
            &\overset{b}{\leq} \frac{1}{n}
		\left( 1 + \frac{ (1-\mu)(1-\beta) }{1 - (1-\mu)(1-\beta)} \right)
		\E\|\!\left( \mu\mI + (1 - \mu) \beta \mW \right)\\
        &(\mM^{(t)} -\bar\mM^{(t)})\!\|^2+ 12\sigma^2+24L^2\rho^2 +\\
        &\frac{1}{n} \left( 1 + \frac{ 1 - (1-\mu)(1-\beta) }{ (1-\mu)(1-\beta) } \right)\E \|
			\frac{1 - \mu}{\eta}(\mI-\mW)\mX^{(t)} +\\
   &(1-\mu) \mW ( \E[ \widetilde{\mG}^{(t)} - \bar\mG^{(t)}])\|^2 \,.
    \end{split}
    % \vspace{-15mm}
\end{equation}

(a) follows from Lemma~\ref{lemma:stochasvar}, and (b) follows from the inequality $\|x_i+x_j\|^2 \leq (1+a)\|x_i\|^2+(1+\frac{1}{a})\|x_j\|^2$ for any $a>0$. Since $\mW < \mI$, we have $\left( \mu\mI + (1 - \mu) \beta \mW \right) < ( \mu + (1 - \mu) \beta)\mI = (1 - (1-\beta)(1-\mu))\mI$. Further, we have $\mI -\mW < 2\mI$ \cite{qgm}. With these observations:
\vspace{-14mm}
	\begin{equation}
    \begin{split}
		&\frac{1}{n}\E\norm{\mM^{(t+1)} - \bar\mM^{(t+1)}}^2
		  \leq \frac{1}{n}
		\left( 1 + \frac{ (1-\mu)(1-\beta) }{1 - (1-\mu)(1-\beta)} \right)\\
		&\E\norm{\left( 1- (1 - \mu)(1 - \beta)\right)(\mM^{(t)} -\bar\mM^{(t)}) }^2
		+ 12\sigma^2+24L^2\rho^2 \\
		  & + \frac{1}{n} \left( 1 + \frac{ 1 - (1-\mu)(1-\beta) }{ (1-\mu)(1-\beta) } \right) \E \|\!\frac{1 - \mu}{\eta}(\mI-\mW)\mX^{(t)}\\
        &+ (1-\mu) \mW \left( \Eb{ \widetilde{\mG}^{(t)} - \bar\mG^{(t)}} \right)\!\|^2\\
		  & \leq \frac{1}{n}\left(1 - (1-\mu)(1-\beta) \right)\E\norm{\mM^{(t)} -\bar\mM^{(t)}}^2
		+ 12\sigma^2\\
       &+24L^2\rho^2 + \frac{1}{ (1-\mu)(1-\beta) n }
		\E \|\frac{1 - \mu}{\eta}(\mI-\mW)\mX^{(t)}\\
        &+ (1-\mu) \mW( \E[\widetilde{\mG}^{(t)} - \bar\mG^{(t)}])
		\|^2 \leq \frac{1}{n}
		\left(1 - (1-\mu)(1-\beta) \right)\\
		&\E\norm{\mM^{(t)} -\bar\mM^{(t)}}^2+ 12\sigma^2+24L^2\rho^2  + \frac{ 4(1-\mu)}{ (1-\beta) n \eta^2}\\
		&\E \norm{ \mX^{(t)} - \bar \mX^{(t)} }^2
		+ \frac{ 2(1-\mu) }{ (1-\beta) n } \E \norm{ \Eb{ \widetilde\mG^{(t)} } - \Eb{\bar\mG^{(t)}} }^2\,.
   \end{split}
	\end{equation}
% \vspace{-3mm}
Substituting equation \ref{eq:star} in the above equation:
\begin{equation}
    \begin{split}
    &\frac{1}{n}\E\norm{\mM^{(t+1)} - \bar\mM^{(t+1)}}^2 \leq \frac{1}{n}
		\left(1 - (1-\mu)(1-\beta) \right)\\
		&\E\norm{\mM^{(t)} -\bar\mM^{(t)}}^2
		+ 12\sigma^2+24L^2\rho^2+ \frac{ 4(1-\mu)}{ (1-\beta) n \eta^2}\\
		&\E \norm{ \mX^{(t)} - \bar \mX^{(t)} }^2
		+ \frac{ 8(1-\mu)L^2 }{ (1-\beta) n } \bigg(\sum_{i=1}^{n} \E \norm{\xx_i^{(t)}-\bar{\mathbf{x}}^{(t)}}^2\bigg)\\
        &+\frac{8(1-\mu)L^2\rho^2 }{(1-\beta)}+\frac{4\delta^2(1-\mu)}{(1-\beta)} \leq \frac{1}{n}\left(1 - (1-\mu)(1-\beta) \right)\\
		&\E\norm{\mM^{(t)} -\bar\mM^{(t)}}^2
		+ 12\sigma^2+24L^2\rho^2+\frac{ 4(1-\mu)(1+2\eta^2L^2)}{ (1-\beta) n \eta^2 }\\
        &\E \norm{ \mX^{(t)} - \bar \mX^{(t)} }^2+\frac{8(1-\mu)L^2\rho^2 }{(1-\beta)}+\frac{4\delta^2(1-\mu)}{(1-\beta)}
    \end{split}
\end{equation}

Multiplying both sides by $\frac{6\eta^2\beta^2}{\lambda(1-\mu)(1-\beta)}$ yields
	\begin{equation}
    \begin{split}
    &\frac{6\eta^2\beta^2}{\lambda n(1-\mu)(1-\beta)}\E\norm{\mM^{(t+1)} - \bar\mM^{(t+1)}}^2 \leq \frac{6\eta^2\beta^2}{\lambda n}\\
    &\bigg( \frac{1}{(1-\mu)(1-\beta)}-1\bigg)\E\norm{\mM^{(t)} - \bar\mM^{(t)}}^2\\
    &+\bigg( \frac{24\beta^2(1+2\eta^2L^2)}{n\lambda(1-\beta)^2}\bigg)\E \norm{ \mX^{(t)} - \bar \mX^{(t)} }^2+\frac{72\eta^2\beta^2\sigma^2}{\lambda(1-\mu)(1-\beta)}\\
    &+\frac{144\eta^2\beta^2L^2\rho^2}{\lambda(1-\mu)(1-\beta)}+\frac{48 L^2\rho^2\eta^2\beta^2}{\lambda(1-\beta)^2}+\frac{24\eta^2\beta^2\delta^2}{\lambda(1-\beta)^2}\\
    &\overset{a}{\leq} \frac{6\eta^2\beta^2}{\lambda n}\bigg( \frac{1}{(1-\mu)(1-\beta)}-1\bigg)\E\norm{\mM^{(t)} - \bar\mM^{(t)}}^2+\frac{\lambda}{8n}\\
    &\E \norm{ \mX^{(t)} - \bar \mX^{(t)} }^2+ \frac{\lambda\eta^2\sigma^2(1-\beta)}{6(1-\mu)}+\bigg(\frac{(1-\beta)}{3(1-\mu)}+\frac{1}{9}\bigg)\\
    &\lambda\eta^2L^2\rho^2+\frac{\lambda\eta^2\delta^2}{18}
    \end{split}
	\end{equation}
(a) follows from our assumption that the momentum parameter satisfies $\frac{\beta}{1 - \beta} \leq \frac{\lambda}{21}$ and $\eta\leq \frac{1}{7L}$.

Adding the results of Lemma \ref{lem:consensus-change} and \ref{lem:momentum} and simplifying the coefficients, we describe the progress made in each gossip averaging consensus round:
\begin{equation}
\begin{split}\label{eq:totalchange}
    &\frac{1}{n}\E\norm{\mX^{t+1} - \bar\mX^{t+1}}^2 + \frac{6\eta^2\beta^2}{n\lambda(1-\mu)(1-\beta)}\E\norm{\mM^{t+1} - \bar\mM^{t+1}}^2          \\& \leq \frac{1 - \lambda/8}{n}\E\norm{\mX^{t} - \bar\mX^{t}}^2 + \frac{6\eta^2\beta^2}{n\lambda(1-\mu)(1-\beta)}\E\norm{\mM^{t} - \bar\mM^{t}}^2\\
    &+ \frac{13\eta^2\delta^2}{\lambda}+ \frac{12\eta^2\sigma^2(2 -\beta -\mu)}{(1-\mu)\lambda}+ \frac{49\eta^2L^2\rho^2(2-\beta-\mu)}{(1-\mu)\lambda}
\end{split}
\end{equation}
\subsection{Proof for Theorem 1} \label{apex:proof}
We start with the following property for a $L$-smooth function $f(\mathbf{x})$:
\begin{equation}
    \begin{split}\label{eq:smooth}
        &\E f(\mz^{(t+1)}) \leq \E f(\mz^{(t)})+\E\lin{\nabla f(\mz^{(t)}) , \mz^{(t+1)} - \mz^{(t)}}+\\
        &\frac{L}{2}\E\norm{\mz^{(t+1)} - \mz^{(t)}}^2
        = \E f(\mz^{(t)})-\\
        &\tilde\eta\underbrace{\E\lin{\nabla f(\mz^{(t)}) , \frac{1}{n} \sum_{i=1}^n \nabla f_i(\xx_i^t+\xi_i^t)}}_{I}+\frac{L}{2}\underbrace{\E\norm{\mz^{(t+1)} - \mz^{(t)}}^2}_{II}
    \end{split}
\end{equation}

We first focus on finding an upper bound for $I$:
\begin{equation}
    \begin{split}\label{eq:I_bound}
        &I: \frac{1}{2}\bigg( \E\norm{\nabla f(\mz^t)}^2+\E\norm{\frac{1}{n} \sum_{i=1}^n \nabla f_i(\xx_i^t+\xi_i^t)}^2\bigg)\\
        &-\frac{1}{2}\underbrace{\bigg( \E\norm{\nabla f(\mz^t))-\frac{1}{n} \sum_{i=1}^n \nabla f_i(\xx_i^t+\xi_i^t)}^2\bigg)}_{\star}
    \end{split}
\end{equation}
To bound $\star$:
\begin{equation}
    \begin{split}\label{eq:star_2}
        &\star: \bigg( \E\norm{\frac{1}{n} \sum_{i=1}^n \nabla f_i(\mz^t)-\frac{1}{n} \sum_{i=1}^n \nabla f_i(\xx_i^t+\xi_i^t)}^2\bigg)\\
        &\leq \frac{1}{n} \sum_{i=1}^{n} \E\norm{\nabla f_i(\mz^t)-\nabla f_i(\xx_i^t+\xi_i^t)}^2\\
        %\overset{a}{\leq} \frac{L^2}{n} \sum_{i=1}^n \norm{\mz^t-\xx_i^t-\xi_i^t}^2 \leq \frac{2L^2}{n} \sum_{i=1}^n (\norm{\mz^t-\xx_i^t}^2+\rho^2) \leq \frac{2L^2}{n} \sum_{i=1}^n \norm{\mz^t-\xx_i^t}^2+2L^2\rho^2
    \end{split}
\end{equation}

Substituting equation \ref{eq:star_2} in \ref{eq:I_bound}:
\begin{equation}
    \begin{split}\label{eq:I_bound_final}
        &I \geq \frac{1}{2}\bigg( \E\norm{\nabla f(\mz^t)}^2+\E\norm{\frac{1}{n} \sum_{i=1}^n \nabla f_i(\xx_i^t+\xi_i^t)}^2\bigg)-\\
        &\frac{1}{2n}\sum_{i=1}^{n} \E\norm{\nabla f_i(\mz^t)-\nabla f_i(\xx_i^t+\xi_i^t)}^2
    \end{split}
    % \vspace{-2mm}
\end{equation}

Now, we find an upper bound for $II$:
\begin{equation}
    \begin{split}\label{eq:II_bound}
        &\E\norm{\mz^{(t+1)} - \mz^{(t)}}^2 = \tilde\eta^2\E\norm{\bar{\mathbf{g}}}^2\\
        &\overset{a}{\leq} \tilde\eta^2 \bigg(\frac{3\sigma^2}{n} +\frac{6L^2\rho^2}{n}+\E\norm{\frac{1}{n}\sum_{i=1}^n \nabla f_i(\xx_i^t+\xi_i^t)}^2\bigg)
    \end{split}
\end{equation}

Here, (a) is the result of Lemma~\ref{lemma:gtilde_bound}. Putting equation \ref{eq:I_bound_final} and \ref{eq:II_bound} in \ref{eq:smooth}:
\begin{equation}
    \begin{split}
        &\E f(\mz^{(t+1)}) \leq \E f(\mz^{(t)})-\frac{\tilde\eta}{2} \E\norm{\nabla f(\mz^t)}^2-\\
        &\frac{\tilde\eta}{2} \E\|\!\frac{1}{n} \sum_{i=1}^n \nabla f_i(\xx_i^t+\xi_i^t)\!\|^2
        +\frac{\tilde\eta}{2n} \sum_{i=1}^{n} \E\|\!\nabla f_i(\mz^t)-\\
        &\nabla f_i(\xx_i^t+\xi_i^t)\!\|^2+\frac{\tilde\eta^2 L}{2}(\frac{3\sigma^2}{n} +\frac{6L^2\rho^2}{n}+\\
        &\E\norm{\frac{1}{n}\sum_{i=1}^n \nabla f_i(\xx_i^t+\xi_i^t)}^2)
    \end{split}
\end{equation}

Rearranging the above terms we get:
\begin{equation}
    \begin{split}\label{eq:rearrange}
        &\E f(\mz^{(t+1)}) \leq \E f(\mz^{(t)})-\frac{\tilde\eta}{2} \E\norm{\nabla f(\mz^t)}^2+\bigg(\frac{\tilde\eta^2 L}{2}-\frac{\tilde\eta}{2}\bigg)\\
        &\E\norm{\frac{1}{n} \sum_{i=1}^n \nabla f_i(\xx_i^t+\xi_i^t)}^2+\frac{\tilde\eta}{2n} \sum_{i=1}^{n} \E\|\!\nabla f_i(\mz^t)-\\
        &\nabla f_i(\xx_i^t+\xi_i^t)\!\|^2+ \frac{3\tilde\eta^2 L^3 \rho^2}{n}+\frac{3\tilde\eta^2 L\sigma^2}{2n} \leq \E f(\mz^{(t)})-\\
        &\frac{\tilde\eta}{4}\E\norm{\nabla f(\bar\xx^t)}^2
        \underbrace{+\frac{\tilde\eta}{2}\E\norm{\nabla f(\mz^t)-\nabla f(\bar\xx^t)}^2}_{\star}\\
        &\underbrace{+\frac{\tilde\eta}{2n} \sum_{i=1}^{n} \E\norm{\nabla f_i(\mz^t)-\nabla f_i(\xx_i^t+\xi_i^t)}^2}_{\star}+\bigg(\frac{\tilde\eta^2 L}{2}-\frac{\tilde\eta}{2}\bigg)\\
        &\E\norm{\frac{1}{n} \sum_{i=1}^n \nabla f_i(\xx_i^t+\xi_i^t)}^2+\frac{3\tilde\eta^2 L^3 \rho^2}{n}+\frac{3\tilde\eta^2 L\sigma^2}{2n}
    \end{split}
\end{equation}
 Now we simplify $\star$:
\begin{equation}
    \begin{split}
        &\star: \frac{\tilde\eta}{2n} \sum_{i=1}^{n} \E\norm{\nabla f_i(\mz^t)-\nabla f_i(\bar\xx^t)}^2+\frac{\tilde\eta}{2n} \sum_{i=1}^{n} \E\|\!\nabla f_i(\mz^t)-\\
        &\nabla f_i(\xx_i^t+\xi_i^t)\!\|^2\leq \frac{\tilde\eta}{2n} \sum_{i=1}^{n} \E\norm{\nabla f_i(\mz^t)-\nabla f_i(\bar\xx^t)}^2\\
        &+\frac{\tilde\eta}{n} \sum_{i=1}^{n} \E\norm{\nabla f_i(\mz^t)-\nabla f_i(\bar\xx^t)}^2+ \frac{\tilde\eta}{n} \sum_{i=1}^{n}\E\|\nabla f_i(\bar\xx^t)-\\
        &\nabla f_i(\xx_i^t+\xi_i^t)\|^2=  \frac{3\tilde\eta}{2n} \sum_{i=1}^{n} \E\norm{\nabla f_i(\mz^t)-\nabla f_i(\bar\xx^t)}^2\\
        &+\frac{\tilde\eta}{n} \sum_{i=1}^{n} \E\norm{\nabla f_i(\bar\xx^t)-\nabla f_i(\xx_i^t+\xi_i^t)}^2 
    \end{split}
\end{equation}
Putting this back into equation \ref{eq:rearrange}:
\begin{equation}
    \begin{split}\label{eq:etacondition}
    &\E f(\mz^{(t+1)}) \leq \E f(\mz^{(t)})-\frac{\tilde\eta}{4}\E\norm{\nabla f(\bar\xx^t)}^2+\bigg(\frac{\tilde\eta^2 L}{2}-\frac{\tilde\eta}{2}\bigg)\\
    &\E\norm{\frac{1}{n} \sum_{i=1}^n \nabla f_i(\xx_i^t+\xi_i^t)}^2+\frac{3\tilde\eta}{2} \sum_{i=1}^{n} \E\norm{\nabla f_i(\mz^t)-\nabla f_i(\bar\xx^t)}^2\\
    &\hspace{2mm} +\frac{\tilde\eta}{n} \sum_{i=1}^{n} \E\norm{\nabla f_i(\bar\xx^t)-\nabla f_i(\xx_i^t+\xi_i^t)}^2+ \frac{3\tilde\eta^2 L^3 \rho^2}{n}+\\
    &\frac{3\tilde\eta^2 L\sigma^2}{2n}
    \end{split}
\end{equation}
Using our assumption $\tilde\eta \leq \frac{1}{4L}$ and Assumption 1, we have:
\begin{equation}
    \begin{split}\label{eq:smooth_prefinal}
    &\E f(\mz^{(t+1)}) \leq \E f(\mz^{(t)})-\frac{\tilde\eta}{4}\E\norm{\nabla f(\bar\xx^t)}^2\\
    &-\frac{\tilde\eta}{4} \E\norm{\frac{1}{n} \sum_{i=1}^n \nabla f_i(\xx_i^t+\xi_i^t)}^2+
    \frac{3\tilde\eta L^2}{2} \sum_{i=1}^{n} \E\norm{\mz^t-\bar\xx^t}^2\\
    &+\frac{\tilde\eta L^2}{n} \sum_{i=1}^{n} \E\norm{\bar\xx^t-\xx_i^t-\xi_i^t}^2+ \frac{3\tilde\eta^2 L^3 \rho^2}{n}+\frac{3\tilde\eta^2 L\sigma^2}{2n} \overset{a}{\leq}\\
    &\E f(\mz^{(t)})-\frac{\tilde\eta}{4}\E\norm{\nabla f(\bar\xx^t)}^2
    -\frac{\tilde\eta}{4} \E\norm{\frac{1}{n} \sum_{i=1}^n \nabla f_i(\xx_i^t+\xi_i^t)}^2\\
    &+\frac{3\tilde\eta L^2}{2} \sum_{i=1}^{n} \E\norm{\mz^t-\bar\xx^t}^2+\frac{2\tilde\eta L^2}{n} \sum_{i=1}^{n} \E\norm{\bar\xx^t-\xx_i^t}^2\\
    &+2\tilde\eta L^2\rho^2+ \frac{3\tilde\eta^2 L^3 \rho^2}{n}+\frac{3\tilde\eta^2 L\sigma^2}{2n}
    \end{split}
\end{equation}

(a) follows from the perturbation $\xi_i$ being bounded by the perturbation radius $\rho$. Now we see that the terms $\|\mz^{(t)} - \bar\xx^{(t)}\|^2$ and $\|\bar\xx^t-\xx_i^t\|^2$, which we bound in Lemma~\ref{lem:error-sequence} and \ref{lem:consensus-change} respectively, appear in the above equation. We start by Lemma~\ref{lem:error-sequence}, and scale both sides by $\frac{3L^2\tilde\eta}{2(1-\mu)(1-\beta)}$:

\begin{equation}
    \begin{split}\label{eq:scaleerror}
		&\frac{3L^2\tilde\eta}{2(1-\mu)(1-\beta)}\E\norm{\ee^{(t+1)}}^2 \leq \bigg(\frac{3L^2\tilde\eta}{2(1-\mu)(1-\beta)}-\frac{3L^2\tilde\eta}{2}\bigg)\\
        &\E\norm{\ee^{(t)}}^2+ \frac{3L^2\tilde \eta^3 \beta^2}{(1-\beta)^2 (1-\mu)^2}\E\norm{\E_t[\bar\gg^{(t)}]}^2 +\\
        &\frac{9L^2\tilde\eta^3\beta^2\sigma^2}{2(1-\mu)(1-\beta)}+\frac{9L^4\tilde\eta^3\beta^2\rho^2}{(1-\mu)(1-\beta)}\\
    \end{split}
\end{equation}

Next, we take the total consensus change from equation \ref{eq:totalchange}, and scale it with $\frac{16L^2\tilde\eta}{\lambda}$:
\begin{equation}
    \begin{split}\label{eq:scaleconsen}
    	&\frac{16L^2\tilde\eta}{\lambda n}\E\norm{\mX^{t+1} - \bar\mX^{t+1}}^2 + \frac{96\tilde\eta^3\beta^2(1-\beta)}{n\lambda^2(1-\mu)}\E\|\!\mM^{t+1} -\\
     &\bar\mM^{t+1}\!\|^2 \leq \frac{16L^2\tilde \eta}{\lambda n}\E\norm{\mX^{t} - \bar\mX^{t}}^2 + \frac{96\tilde\eta^3\beta^2(1-\beta)}{n\lambda^2(1-\mu)}\\
     &\E\norm{\mM^{t} - \bar\mM^{t}}^2- \frac{2L^2\tilde \eta}{n}\E\norm{\mX^{t} - \bar\mX^{t}}^2+ \frac{208L^2\tilde\eta^3(1-\beta)^2\delta^2}{\lambda^2}\\
    &+\frac{192L^2\tilde\eta^3(2 -\beta -\mu)(1-\beta)^2\sigma^2}{(1-\mu)\lambda^2}\\
    &+ \frac{784L^4\tilde\eta^3(1-\beta)^2(2-\beta-\mu)\rho^2}{(1-\mu)\lambda^2}
    \end{split}
\end{equation}

Through equation \ref{eq:scaleerror} and \ref{eq:scaleconsen}, we define another sequence $\phi^t\geq0$ such that $\phi^0=\E[f(\bar\xx^0)-f^*]$:
	\begin{equation*}
    \begin{split}
		&\phi^t : \frac{16L^2 \tilde\eta}{\lambda n}\E\norm{\mX^{t} - \bar\mX^{t}}^2 + \frac{96 L^2 \tilde\eta^3\beta^2 (1-\beta)}{n\lambda^2 (1-\mu)}\E\norm{\mM^{t} - \bar\mM^{t}}^2\\
        &+ \frac{3L^2\tilde\eta}{2(1-\mu)(1-\beta)}\E\norm{\ee^{(t)}}^2 + \E[f(\bar\xx^{t}) - f^\star]
    \end{split}
	\end{equation*}

Adding the right hand sides of equation \ref{eq:smooth_prefinal}, \ref{eq:scaleerror} and \ref{eq:scaleconsen}, and bounding $\phi^{t+1}$ in terms of $\phi^t$:
\begin{equation}
    \begin{split}
        &\phi^{t+1} \leq \phi^t - \frac{\tilde\eta}{4}\norm{\nabla f(\bar\xx^{(t)})}^2+ \left(  \frac{3L^2\tilde\eta^3 \beta^2}{(1-\beta)^2 (1-\mu)^2} - \frac{\tilde\eta}{4}\right)\\
        &\E\norm{\frac{1}{n}\sum_{i=1}^n \nabla f_i(\xx_i^{t}+\xi_i^t)}^2+2\tilde\eta L^2\rho^2+ \frac{3\tilde\eta^2 L^3 \rho^2}{n}+\frac{3\tilde\eta^2 L\sigma^2}{2n}\\
        &+ \frac{9L^2\tilde\eta^3\beta^2\sigma^2}{2(1-\mu)(1-\beta)}+\frac{9L^4\tilde\eta^3\beta^2\rho^2}{(1-\mu)(1-\beta)}+
        \frac{208L^2\tilde\eta^3(1-\beta)^2\delta^2}{\lambda^2}\\
        &+\frac{192L^2\tilde\eta^3(2 -\beta -\mu)(1-\beta)^2\sigma^2}{(1-\mu)\lambda^2}\\
        &+ \frac{784L^4\tilde\eta^3(1-\beta)^2(2-\beta-\mu)\rho^2}{(1-\mu)\lambda^2}
    \end{split}
\end{equation}
Simplifying the above equation by rearranging terms and approximating some coefficients:
\begin{equation}
    \begin{split}\label{eq:etacond2}
        &\norm{\nabla f(\bar\xx^{(t)})}^2 \leq \frac{4}{\tilde\eta}(\phi^t - \phi^{t+1}) + \left(\frac{12L^2\tilde\eta^2 \beta^2}{(1-\beta)^2 (1-\mu)^2} - 1\right)\\ &\E\norm{\frac{1}{n}\sum_{i=1}^n \nabla f_i(\xx_i^{t}+\xi_i^t)}^2+\underbrace{\frac{6\tilde\eta L}{n}+18\tilde\eta^2 L^2C_2+\frac{768\tilde\eta^2L^2C_1}{\lambda^2}}_{C_{\sigma}}\\
        &\sigma^2+\underbrace{\frac{832L^2\tilde\eta^2(1-\beta)^2}{\lambda^2}}_{C_{\delta}}\delta^2\\
        &+\underbrace{8 L^2+\frac{12\tilde\eta L^3}{n}+36\tilde\eta^2L^4C_2+\frac{3136L^4\tilde\eta^2C_1}{\lambda^2}}_{C_{\rho}}\rho^2
    \end{split}
\end{equation}
Here, $C_1= \frac{(2-\beta-\mu)(1-\beta)^2}{(1-\mu)}$ and $C_2=\frac{\beta^2}{(1-\mu)(1-\beta)}$.

For $\frac{12L^2\tilde\eta^2 \beta^2}{(1-\beta)^2 (1-\mu)^2}-1 \leq 0$:
\begin{equation}
    \begin{split}
        &\norm{\nabla f(\bar\xx^{(t)})}^2 \leq \frac{4}{\tilde\eta}(\phi^t - \phi^{t+1}) + C_{\sigma}\sigma^2+C_{\delta}\delta^2+C_{\rho}\rho^2
    \end{split}
\end{equation}

Averaging over $T$, we have:
\begin{equation}
    \begin{split}\label{eq:final_pre}
        &\frac{1}{T}\sum_{t=0}^{T-1}\E\norm{\nabla f(\bar\xx^{(t)})}^2 \leq \frac{4}{\tilde\eta T}(f(\bar \xx^0)-f^*) + C_{\sigma}\sigma^2+\\
        &C_{\delta}\delta^2+C_{\rho}\rho^2
    \end{split}
\end{equation}
\subsection{Proof for Corollary 2} \label{apex:proof_corol}
To find the convergence rate with a learning rate $\eta=\mathcal{O}\bigg(\sqrt{\frac{n}{T}}\bigg)$ and perturbation radius $\rho=\mathcal{O}{\bigg(\sqrt{\frac{1}{T}}\bigg)}$, we find the order of all the terms in equation \ref{eq:final_pre}:
\begin{itemize}
    \item $\frac{4}{\tilde\eta T}(f(\bar \xx^0)-f^*)= \mathcal{O}\bigg(\frac{1}{\sqrt{nT}}\bigg)$
    \item $C_{\sigma}\sigma^2= \mathcal{O}\bigg(\frac{\eta}{n}+\eta^2\bigg)= \mathcal{O}\bigg( \frac{1}{\sqrt{nT}}+\frac{n}{T}\bigg)$
    \item $C_{\delta}\delta^2=\mathcal{O}\bigg(\eta^2\bigg)=\mathcal{O}\bigg(\frac{n}{T}\bigg)$
    \item $C_{\rho}\rho^2= \mathcal{O}\bigg(\rho^2+\frac{\eta\rho^2}{n}+\eta^2\rho^2\bigg)=\mathcal{O}\bigg(\frac{1}{T}+\frac{1}{n^{1/2}T^{3/2}}+\frac{n}{T^2}\bigg)$
\end{itemize}

Adding all the terms and ignoring $n$ in higher order terms:
\begin{equation}
    \begin{split}
        &\frac{1}{T}\sum_{t=0}^{T-1}\E\norm{\nabla f(\bar\xx^{(t)})}^2 \leq \mathcal{O}\bigg(\frac{1}{\sqrt{nT}}+\frac{1}{T}+\frac{1}{T^{3/2}}+\frac{1}{T^2}\bigg)
    \end{split}
\end{equation}

This implies that when $T$ is sufficiently large, Q-SADDLe converges at the rate of $\mathcal{O}\bigg(\frac{1}{\sqrt{nT}}\bigg)$.

\subsection{Condition on Learning Rate $\eta$ and Momentum Coefficient $\beta$}
%add all previously used conditions here!
In Lemma~\ref{lem:consensus-change}, we assume $\eta\leq \frac{\lambda}{10L}$ and in Lemma~\ref{lem:momentum}, we assume $\eta\leq \frac{1}{7L}$. Combining both bounds results in $\eta\leq \min(\frac{1}{7L}, \frac{\lambda}{10L}) \leq \min(\frac{1}{7L}, \frac{\lambda}{7L}) \leq \frac{\lambda}{7L}$.In Theorem 1 proof, we assume $\eta \leq \frac{(1-\beta)}{4L}$ to simplify equation~\ref{eq:etacondition}. Further to simplify equation~\ref{eq:etacond2}, we have the following upper bound on $\eta$:
\begin{equation}
\begin{split}
    &\frac{12L^2\tilde\eta^2 \beta^2}{(1-\beta)^2 (1-\mu)^2}-1 \leq 0\\
    &12L^2\tilde\eta^2 \beta^2-(1-\beta)^2 (1-\mu)^2 \leq 0\\
    &\eta \leq \frac{(1-\beta)^2(1-\mu)}{\sqrt{12}L\beta}
\end{split}
\end{equation}

Combining all the above mentioned bounds, we can describe $\eta \leq \min \left(\frac{\lambda}{7 L}, \frac{1-\beta}{4L} ,  \frac{(1-\beta)^2(1-\mu)}{\sqrt{12}L\beta}\right)$.

Similarly, for momentum coefficient $\beta$, we assume $\frac{\beta}{1-\beta} \leq \frac{\lambda}{21}$ in Lemma~\ref{lem:momentum}. Note that we don't abide by these constraints and still achieve competitive performance for our results in Section 6 (main paper) and Section \ref{addresults} (Supplementary).

\section{Algorithmic Details}
\subsection{Background}\label{apex:bg}
To highlight that SADDLe can improve the generalization and communication efficiency of existing decentralized algorithms, we choose two state-of-the-art techniques for our evaluation: Quasi-Global Momentum (QGM) \cite{qgm} and Neighborhood Gradient Mean (NGM) \cite{ngm}. QGM improves the performance of D-PSGD \cite{dpsgd} without introducing any extra communication. However, as shown in our results in Section 6, it performs poorly with extreme data heterogeneity. To achieve competitive performance with higher degrees of non-IIDness, NGM proposes to boost the performance through cross-gradients, which require 2x communication (i.e., an extra round of communication) as compared to D-PSGD \cite{dpsgd}.

\textbf{Quasi-Global Momentum (QGM)}: The authors in QGM \cite{qgm} show that local momentum acceleration is hindered by data heterogeneity. Inspired by this, QGM updates the momentum buffer by computing the difference between two consecutive models $\mathbf{x}_i^{t+1}$ and $\mathbf{x}_i^{t}$ to approximate the global optimization direction locally. The following equation illustrates the update rule for QGM:
\begin{equation}
\label{eq:qgm}
\begin{split}
   &\text{QGM:} \hspace{2mm}\mathbf{x}_i^{t+1} = \sum_{j \in \mathcal{N}(i)} w_{ij} [\mathbf{x}_j^t - \eta (\mathbf{g}_j^t+\beta \mathbf{m}_j^{t-1} )]\\
   &\text{where,} \hspace{2mm} \mathbf{m}_i^t=\mu \mathbf{m}_i^{t-1}+ (1-\mu)\frac{\mathbf{x}_i^{t}-\mathbf{x}_i^{(t+1)}}{\eta}.
\end{split}
\end{equation}

\textbf{Neighborhood Gradient Mean (NGM)}: NGM \cite{ngm} modifies the local gradient update with the aid of self and cross-gradients. The self-gradients are computed at each agent through its model parameters and the local dataset. The data variant cross-gradients are derivatives of the local model with respect to the dataset of neighbors. These gradients are obtained through an additional round of communication. The update rule for NGM is shown in equation \ref{eq:ngm}, where each gradient update $\mathbf{g}_j^t$ is a weighted average of the self and received cross-gradients.
% \vspace{-1mm}
\begin{equation}
\label{eq:ngm}
\begin{split}
   &\text{NGM:} \hspace{2mm} \mathbf{x}_i^{t+1} = \sum_{j \in \mathcal{N}(i)}w_{ij}\mathbf{x}_j^t - \eta \mathbf{g}_j^t;\\
   &\mathbf{g}_j^t= \sum_{j\in \mathcal{N}(i)} w_{ij}\nabla F_j(\mathbf{x}_i^{t}; d_j^t).  
\end{split}
\end{equation}

\begin{algorithm}[ht]
\textbf{Input:} Each agent $i \in [1,n]$ initializes model weights $\mathbf{x}_i$, step size $\eta$, momentum coefficient $\beta$, averaging rate $\gamma$, mixing matrix $\mathbf{W}=[w_{ij}]_{i,j \in [1,n]}$, and $I_{ij}$ are elements of $n\times n$ identity matrix, $\mathcal{N}(i)$ represents neighbors of $i$ including itself.\\
 
  \textbf{procedure} T\text{\scriptsize RAIN}( ) $\forall i$\\
1.  \hspace{4mm}\textbf{for} t = $1,2,\hdots,T$ \textbf{do}\\
2.  \hspace*{8mm} $d^i_{t} \sim D^i$\\
3.  \hspace*{8mm}$\mathbf{g}_{ii}^{t}=\nabla F_i(\mathbf{x}_i^{t}; d_i^t)$\\
4.  \hspace*{8mm}\colorbox{lightmintbg}{$\tilde{\mathbf{g}}_{ii}^{t}=\nabla F_i(\mathbf{x}_i^{t}+ \xi(\mathbf{x}_i^{t}); d_i^t)$, where $\xi(\mathbf{x}_i^{t})= \rho \frac{\mathbf{g}_{ii}^t}{\|\mathbf{g}_{ii}^t\|}$} \\
5.  \hspace*{8mm}S\text{\scriptsize END}R\text{\scriptsize ECEIVE}($\mathbf{x}_i^{t}$)\\
6.  \hspace*{8mm}\textbf{for} each neighbor $j \in \{\mathcal{N}(i)-i\}$ \textbf{do}\\
7.  \hspace*{12mm}$\mathbf{g}_{ji}^{t}=\nabla F_i(\mathbf{x}_j^{t}; d_i^t)$\\
8.  \hspace*{12mm}\colorbox{lightmintbg}{$\tilde{\mathbf{g}}_{ji}^{t}=\nabla F_i(\mathbf{x}_j^{t}+ \xi(\mathbf{x}_j^{t}); d_i^t)$, where $\xi(\mathbf{x}_j^{t})= \rho \frac{\mathbf{g}_{ji}^t}{\|\mathbf{g}_{ji}^t\|}$} \\
9.  \hspace*{12mm} S\text{\scriptsize END}R\text{\scriptsize ECEIVE}\colorbox{lightapricot}{($\mathbf{g}_{ji}^{t}$)} \colorbox{lightmintbg}{($\tilde{\mathbf{g}}_{ji}^{t}$)}\\
10. \hspace*{8mm}\textbf{end}\\
11. \hspace*{8mm}\colorbox{lightapricot}{$\mathbf{g}_i^{t}=\sum_{j\in \mathcal{N}(i)} w_{ij}\mathbf{g}_{ij}^{t}$}\\
12. \hspace*{8mm}\colorbox{lightapricot}{$\mathbf{m}_i^{t}= \beta \mathbf{m}_i^{(t-1)} + \mathbf{g}_i^{t}$}\\
13. \hspace*{8mm}\colorbox{lightmintbg}{$\tilde{\mathbf{g}}_i^{t}=\sum_{j\in \mathcal{N}(i)} w_{ij}\tilde{\mathbf{g}}_{ij}^{t}$}\\

14. \hspace*{8mm}\colorbox{lightmintbg}{$\mathbf{m}_i^{t}= \beta \mathbf{m}_i^{(t-1)} + \tilde{\mathbf{g}}_i^{t}$}\\
15. \hspace*{8mm}$\mathbf{x}_i^{(t+1/2)}=\mathbf{x}_i^{t}- \eta \mathbf{m}_i^{t}$\\
16. \hspace*{8mm}$\mathbf{x}_i^{(t+1)}=\mathbf{x}_i^{(t+1/2)}+\gamma\sum_{j\in \mathcal{N}(i)}(w_{ij}-I_{ij})\mathbf{x}_j^{t}$\\
17. \hspace{4mm}\textbf{end}\\
 \textbf{return $\mathbf{x}_i^{T}$}
\caption{\colorbox{lightapricot}{NGM} vs \colorbox{lightmintbg}{N-SADDLe}}
\label{alg:NGM}
\end{algorithm}

\begin{algorithm}[ht]
\textbf{Input:} Each agent $i$ initializes model weights $\mathbf{x}_i$, step size $\eta$, averaging rate $\gamma$, mixing matrix $\mathbf{W}=[w_{ij}]_{i,j \in [1,n]}$, $Q(.)$ is the compression operator, $\mathcal{N}(i)$ represents neighbors of $i$.\\
 
% Each agent simultaneously implements the 
% T\text{\scriptsize RAIN}( ) procedure\\
  \textbf{procedure} T\text{\scriptsize RAIN}( ) $\forall i$\\
1.  \hspace{4mm}\textbf{for} t=$1,2,\hdots,T$ \textbf{do}\\
2.  \hspace*{8mm}$d_i^{t} \sim D_i$\\
3.  \hspace*{8mm}$\mathbf{g}_{ii}^{t}=\nabla F_i(\mathbf{x}_i^{t}; d_i^t)$\\
4.  \hspace*{8mm}\colorbox{lightmintbg}{$\widetilde{\mathbf{g}}_{ii}^{t}=\nabla F_i(\mathbf{x}_i^{t}+ \xi(\mathbf{x}_i^{t}); d_i^t)$, where $\xi(\mathbf{x}_i^{t})= \rho \frac{\mathbf{g}_{ii}^t}{\|\mathbf{g}_{ii}^t\|}$}\\
5.  \hspace*{8mm}\colorbox{lightapricot}{$\mathbf{p}_{ii}^{t}=\mathbf{g}_{ii}^{t}+\mathbf{e}_{ii}^{t}$}\\
6.  \hspace*{8mm}\colorbox{lightmintbg}{$\mathbf{p}_{ii}^{t}=\widetilde{\mathbf{g}}_{ii}^{t}+\mathbf{e}_{ii}^{t}$}\\
% 8.  \hspace*{8mm}$\mathbf{\delta}_{ii}^{t}=(||\mathbf{p}_{ii}^{t}||_1/d)sgn(\mathbf{p}_{ii}^t)$\\
7.    \hspace*{8mm} $\mathbf{\delta}_{ii}^{t}= Q(\mathbf{p}_{ii}^t)$\\
8.  \hspace*{8mm}$\mathbf{e}_{ii}^{t+1}=\mathbf{p}_{ii}^{t}-\mathbf{\delta}_{ii}^{t}$\\
9.  \hspace*{8mm}S\text{\scriptsize END}R\text{\scriptsize ECEIVE}($\mathbf{x}_i^{t}$)\\
10.  \hspace*{8mm}\textbf{for} each neighbor $j \in \{N(i)-i\}$ \textbf{do}\\
11.  \hspace*{12mm}$\mathbf{g}_{ji}^{t}=\nabla F_i(\mathbf{x}_j^{t}; d_i^t)$\\
12.  \hspace*{11mm}\colorbox{lightmintbg}{$\widetilde{\mathbf{g}}_{ji}^{t}=\nabla F_i(\mathbf{x}_j^{t}+ \xi(\mathbf{x}_j^{t}); d_i^t)$, where $\xi(\mathbf{x}_j^{t})= \rho \frac{\mathbf{g}_{ji}^t}{\|\mathbf{g}_{ji}^t\|}$} \\
13. \hspace*{12mm}\colorbox{lightapricot}{$\mathbf{p}_{ji}^{t}=\mathbf{g}_{ji}^{t}+\mathbf{e}_{ji}^{t}$}\\
14.  \hspace*{12mm}\colorbox{lightmintbg}{$\mathbf{p}_{ji}^{t}=\widetilde{\mathbf{g}}_{ji}^{t}+\mathbf{e}_{ji}^{t}$}\\
% 13. \hspace*{12mm}$\mathbf{\delta}_{ji}^{t}=(||\mathbf{p}_{ji}^{t}||_1/d)sgn(\mathbf{p}_{ji}^t)$\\
15. \hspace*{12mm}$\mathbf{\delta}_{ji}^{t}= Q(\mathbf{p}_{ji}^{t})$\\
16. \hspace*{12mm}$\mathbf{e}_{ji}^{t+1}=\mathbf{p}_{ji}^{t}-\mathbf{\delta}_{ji}^{t}$\\
17. \hspace*{12mm}S\text{\scriptsize END}R\text{\scriptsize ECEIVE}($\mathbf{\delta}_{ji}^{t}$)\\
18. \hspace*{12mm}\textbf{end}\\
19. \hspace*{8mm}\textbf{end}\\
20. \hspace*{8mm}$\mathbf{g}_i^{t}=\sum_{j\in \mathcal{N}(i)} w_{ij}\mathbf{\delta}_{ij}^{t}$\\
21.  \hspace*{8mm}$\mathbf{m}_i^{t}= \beta \mathbf{m}_i^{(t-1)}+\mathbf{g}_i^{t}$\\
22. \hspace*{8mm}$\mathbf{x}_i^{(t+1/2)}=\mathbf{x}_i^{t}- \eta \mathbf{m}_i^{t}$\\
23. \hspace*{8mm}$\mathbf{x}_i^{(t+1)}=\mathbf{x}_i^{(t+1/2)}+\gamma\sum_{j\in \mathcal{N}(i)}(w_{ij}-I_{ij}) \mathbf{x}_j^{t}$\\
24. \hspace*{4mm}\textbf{end}\\
 \textbf{return $\mathbf{x}_i^{T}$}
\caption{\colorbox{lightapricot}{Comp NGM} vs \colorbox{lightmintbg}{Comp N-SADDLe}}
\label{apx_alg:compNGM}
\end{algorithm}
\subsection{N-SADDLe and Comp N-SADDLe}
Algorithm \ref{alg:NGM} highlights the difference between NGM and N-SADDLe. Specifically, N-SADDLe computes SAM-based gradient updates for self and cross gradients (lines 4 and 8). Similarly, please refer to Algorithm \ref{apx_alg:compNGM} to understand the difference between the compressed versions of NGM and N-SADDLe (i.e., Comp NGM and Comp N-SADDLe). The error between the original gradients and their compressed version is added as feedback to the gradients before compressing them in the next iteration (lines 5, 6, 13, and 14 in Algorithm \ref{apx_alg:compNGM}).

\begin{algorithm}[h!]
\textbf{Input:} Each agent $i \in [1,n]$ initializes model weights $\mathbf{x}_i^{(0)}$, learning rate $\eta$, perturbation radius $\rho$, and  mixing matrix $\mathbf{W}=[w_{ij}]_{i,j \in [1,n]}$, $\mathcal{N}(i)$ represents neighbors of $i$.\\
  \textbf{procedure} T\text{\scriptsize RAIN}( ) $\forall i$\\
1.  \hspace{4mm}\textbf{for} t=$0,1,\hdots,T-1$ \textbf{do}\\
2.  \hspace*{8mm}$d_i^{k} \sim D_i$\\
3.  \hspace*{8mm}$\mathbf{g}_{i}^{t}=\nabla F_i(d_i^{t}; \mathbf{x}_i^{t}) $ \\
4.  \hspace*{8mm}\colorbox{lightmintbg}{$\widetilde{\mathbf{g}}_{i}^{t}=\nabla F_i(\mathbf{x}_i^{t}+ \xi(\mathbf{x}_i^{t}); d_i^t)$, where $\xi(\mathbf{x}_i^{t})= \rho \frac{\mathbf{g}_{i}^t}{\|\mathbf{g}_{i}^t\|}$}\\
5.  \hspace*{8mm}\colorbox{lightapricot}{$\mathbf{x}_{i}^{(t+\frac{1}{2})}=\mathbf{x}_i^{t}-\eta \mathbf{g}_{i}^{t}$}\\
6. \hspace*{8mm}\colorbox{lightmintbg}{$\mathbf{x}_{i}^{(t+\frac{1}{2})}=\mathbf{x}_i^{t}-\eta \widetilde{\mathbf{g}}_{i}^{t}$}\\
7.  \hspace*{8mm}S\text{\scriptsize END}R\text{\scriptsize ECEIVE}($\mathbf{x}_{i}^{(t+\frac{1}{2})}$)\\
8.  \hspace*{8mm}$\mathbf{x}_i^{(t+1)}=\sum_{j\in \mathcal{N}_i} w_{ij} \mathbf{x}_{j}^{t+\frac{1}{2}}$\\
\textbf{return}
\caption{\colorbox{lightapricot}{DPSGD} vs \colorbox{lightmintbg}{D-SADDLe}}
\label{alg:dl}
\end{algorithm}
\section{Additional Results}\label{addresults}

\subsection{SADDLe with DPSGD} 
A natural question that arises is, does SADDLe improve the performance of DPSGD\cite{dpsgd} in the presence of data heterogeneity? Note that DPSGD assumes the data distribution to be IID and has been shown to incur significant performance drop with non-IID data \cite{qgm}. Algorithm \ref{alg:dl} shows the difference between DPSGD and D-SADDLe, a version incorporating SAM-based updates within DPSGD. D-SADDLe leads to an average improvement of 10\% and 5.4\% over DPSGD for CIFAR-10 and CIFAR-100 datasets, respectively, as shown in Table \ref{tab:dsamcf10_100}. 
%However, the accuracy achieved by D-SADDLe is lower than QGM (results in Table \ref{tab:qgmcf10_100}) and significantly lesser than the upper bound, i.e., DPSGD results on IID data.
\begin{table*}[ht!]
\caption{Test accuracy of DPSGD and D-SADDLe evaluated on CIFAR-10 and CIFAR-100 over ResNet-20, distributed with different degrees of heterogeneity over ring topologies.}
% \vspace{-3mm}
\label{tab:dsamcf10_100}
\begin{center}
\small
%\resizebox{1.0\columnwidth}{!}{
\begin{tabular*} {\textwidth}{cl @{\extracolsep{\fill}}*{4}{c}}
\hline
\multirow{ 2}{*}{Agents} &\multirow{ 2}{*}{Method}& \multicolumn{2}{c}{CIFAR-10} & \multicolumn{2}{c}{CIFAR-100}\\
\cline{3-4}  
\cline{5-6}
& & $\alpha=0.01$ & $\alpha=0.001$ & $\alpha=0.01$ & $\alpha=0.001$\\
 \hline
 \multirow{3}{*}{$5$} & DPSGD (IID) & \multicolumn{2}{c}{91.05 $\pm$ 0.06}  & \multicolumn{2}{c}{64.47 $\pm$ 0.48}\\
 & DPSGD &  82.15 $\pm$ 3.25 & 80.54 $\pm$ 4.36 & 47.30 $\pm$ 4.92 & 45.54 $\pm$ 0.71\\
 & \textit {D-SADDLe (ours)} & \textbf{85.38 $\pm$ 0.84} & \textbf{84.94 $\pm$ 0.31} & \textbf{54.35 $\pm$ 0.48} & \textbf{54.30 $\pm$ 0.50}\\
 \hline
\multirow{3}{*}{$10$} & DPSGD (IID) & \multicolumn{2}{c}{90.46 $\pm$ 0.33} & \multicolumn{2}{c}{62.73 $\pm$ 1.03} \\
  & DPSGD & 49.17 $\pm$ 17.38 & 40.74 $\pm$ 2.62 & 31.66 $\pm$ 0.84 & 29.79 $\pm$ 1.30 \\
 & \textit {D-SADDLe (ours)} & \textbf{64.18 $\pm$ 5.63} & \textbf{61.30 $\pm$ 0.79} & \textbf{37.49 $\pm$ 0.59} & \textbf{35.31 $\pm$ 0.77}\\  \hline
 \multirow{3}{*}{$20$} & DPSGD (IID) & \multicolumn{2}{c}{89.46 $\pm$ 0.02} & \multicolumn{2}{c}{59.61 $\pm$ 1.15} \\
 & DPSGD &  40.49 $\pm$ 3.06 & 36.13 $\pm$ 5.67 & 24.45 $\pm$ 0.51 & 21.58 $\pm$ 1.00 \\
 & \textit {D-SADDLe (ours)} & \textbf{52.14 $\pm$ 2.02} & \textbf{47.06 $\pm$ 2.35} & \textbf{26.39 $\pm$ 0.17} & \textbf{24.92 $\pm$ 0.62} \\
 %%%%%%%%%%%%%%%%%%%
 \hline
 % \vspace{-4mm}
\end{tabular*}
\end{center}
\end{table*}

\subsection{Results with Top-k Sparsification}
% \vspace{-2mm}
We present results for QGM and Q-SADDLe with Top-30\% Sparsification-based compressor in Table \ref{tab:qgmcf10_100_topk}. Note that Top-30\% implies that only the top 30\% of model updates for each layer are communicated to the peers. As shown in Table \ref{tab:qgmcf10_100_topk}, QGM performs poorly in the presence of compression, with a significant drop of $\sim5-57\%$, and the training even diverges for some cases. In contrast, Q-SADDLe is much more stable, with an accuracy drop of $\sim0.6-18.5\%$ with compression.
\begin{table*}[ht]
% \vspace{-3mm}
\caption{Test accuracy (Acc) and accuracy drop (Drop) of QGM and Q-SADDLe with Sparsification (top-30\%) based compression evaluated on CIFAR-10 distributed over ring topologies. $\star$ indicates 1 out of 3 runs diverged.}
% \vspace{-3mm}
\label{tab:qgmcf10_100_topk}
\small
\begin{center}
\begin{tabular*} {\textwidth}{ccl @{\extracolsep{\fill}}*{4}{c}}
\hline
\multirow{2}{*}{Agents} &\multirow{2}{*}{Comp} & \multirow{ 2}{*}{Method}& \multicolumn{4}{c}{CIFAR-10}\\
\cline{4-7}  
& & & \multicolumn{2}{c}{$\alpha=0.01$} & \multicolumn{2}{c}{$\alpha=0.001$}\\
\cline{4-5}
\cline{6-7}
& & & Acc (\%) & Drop(\%) & Acc (\%) & Drop(\%)\\
 \hline
 \multirow{2}{*}{$5$} & \checkmark & QGM & 83.58 $\pm$ 2.96 & 4.86 & 67.04 $\pm$ 9.76 & 21.68\\
 & \checkmark & \textit{Q-SADDLe (ours)} & \textbf{90.01 $\pm$ 0.38} & 0.65 & \textbf{89.49 $\pm$ 0.38} & 1.18\\
 \hline
  \multirow{2}{*}{$10$} & \checkmark & QGM & 52.23 $\star$ & 25.18 & 23.00 $\pm$ 1.96 & 56.48\\
 & \checkmark & \textit{Q-SADDLe (ours)} & \textbf{80.34 $\pm$ 5.56} & 7.38 & \textbf{71.01 $\pm$ 3.75} & 15.32\\
  \hline
 \multirow{2}{*}{$20$} & \checkmark & QGM & 62.90 $\pm$ 5.89 & 9.3 & 32.92 $\pm$ 9.25 & 29.56\\
 & \checkmark & \textit{Q-SADDLe (ours)} & \textbf{71.96 $\pm$ 2.51} & 6.45 & \textbf{64.31 $\pm$ 2.14} & 18.50\\
 %%%%%%%%%%%%%%%%%%%
 \hline
\end{tabular*}
\end{center}
% \vspace{-4mm}
\end{table*}
% \subsection{Results for Torus Topology}
% We present additional results on torus topology with 20 and 40 agents in Table \ref{tab:torus}. Each agent has 4 peers/neighbors; hence, the torus has higher connectivity than the ring. Q-SADDLe and N-SADDLe achieve 16.5\% and 0.8\% better accuracy than QGM and NGM, respectively. In the presence of communication compression, Q-SADDLe incurs a $\sim$3\% drop, while QGM suffers a significant accuracy drop of 9.6\%. Interestingly, both NGM and N-SADDLe nearly retain their accuracy even in the presence of compression. 

\subsection{Compression Error and Gradient Norms for N-SADDLe}
Recall that the expectation of compression error for a compression operator $Q(.)$ has the following upper bound:
\begin{equation}\label{apx:comperrorbound}
    \mathbb{E}_Q \|Q(\mathbf{\theta})- \theta\|^2 \leq (1-\zeta) \|\theta\|^2, \hspace{1mm} \text{where} \hspace{1mm} \zeta>0
\end{equation}

\begin{figure*}[ht]
\centering
	\begin{subfigure}{0.4\linewidth}
		\includegraphics[width=\textwidth]{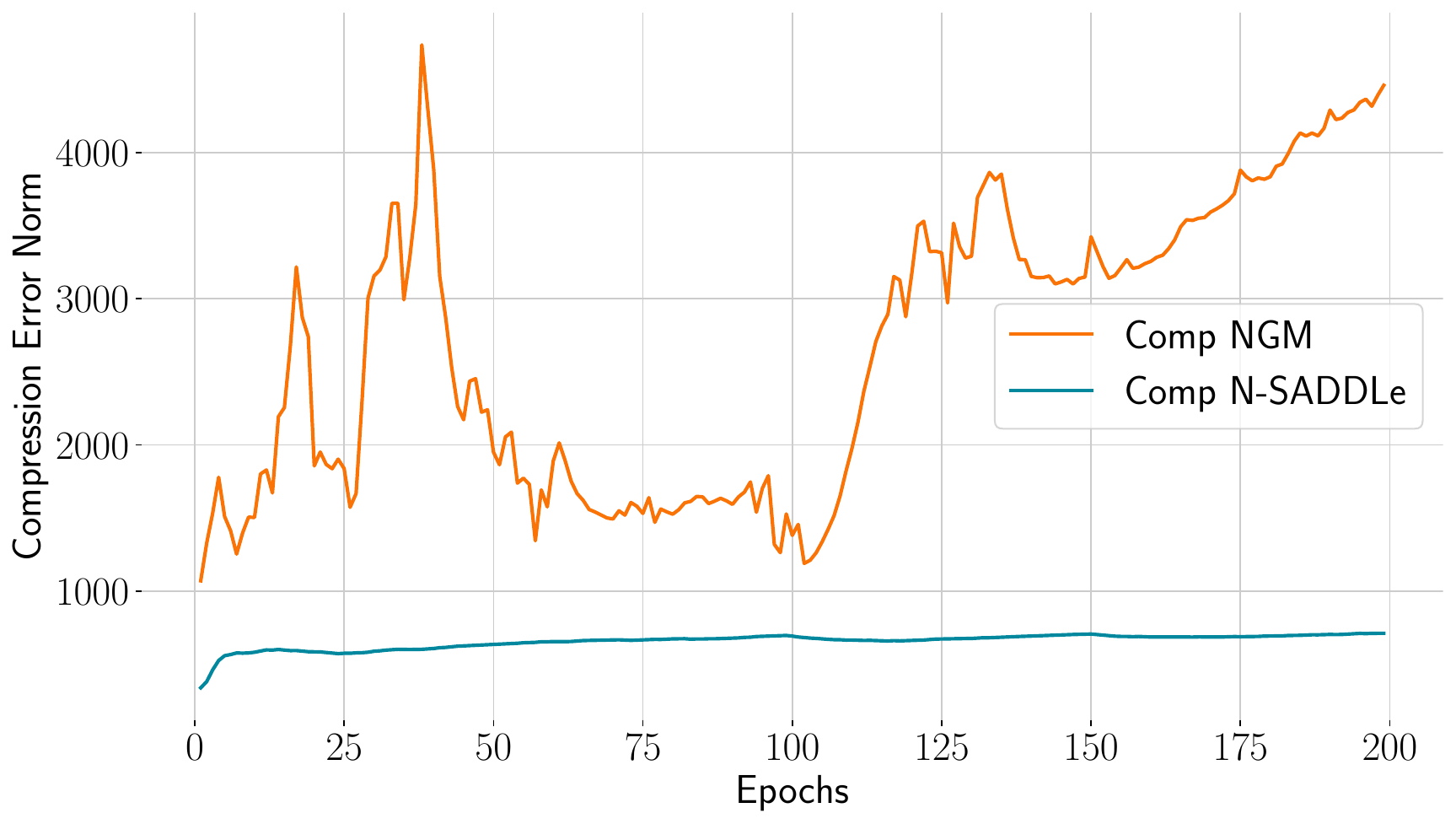}
		\caption{Compression Error}
	\end{subfigure}
	\begin{subfigure}{0.4\linewidth}
		\includegraphics[width=\textwidth]{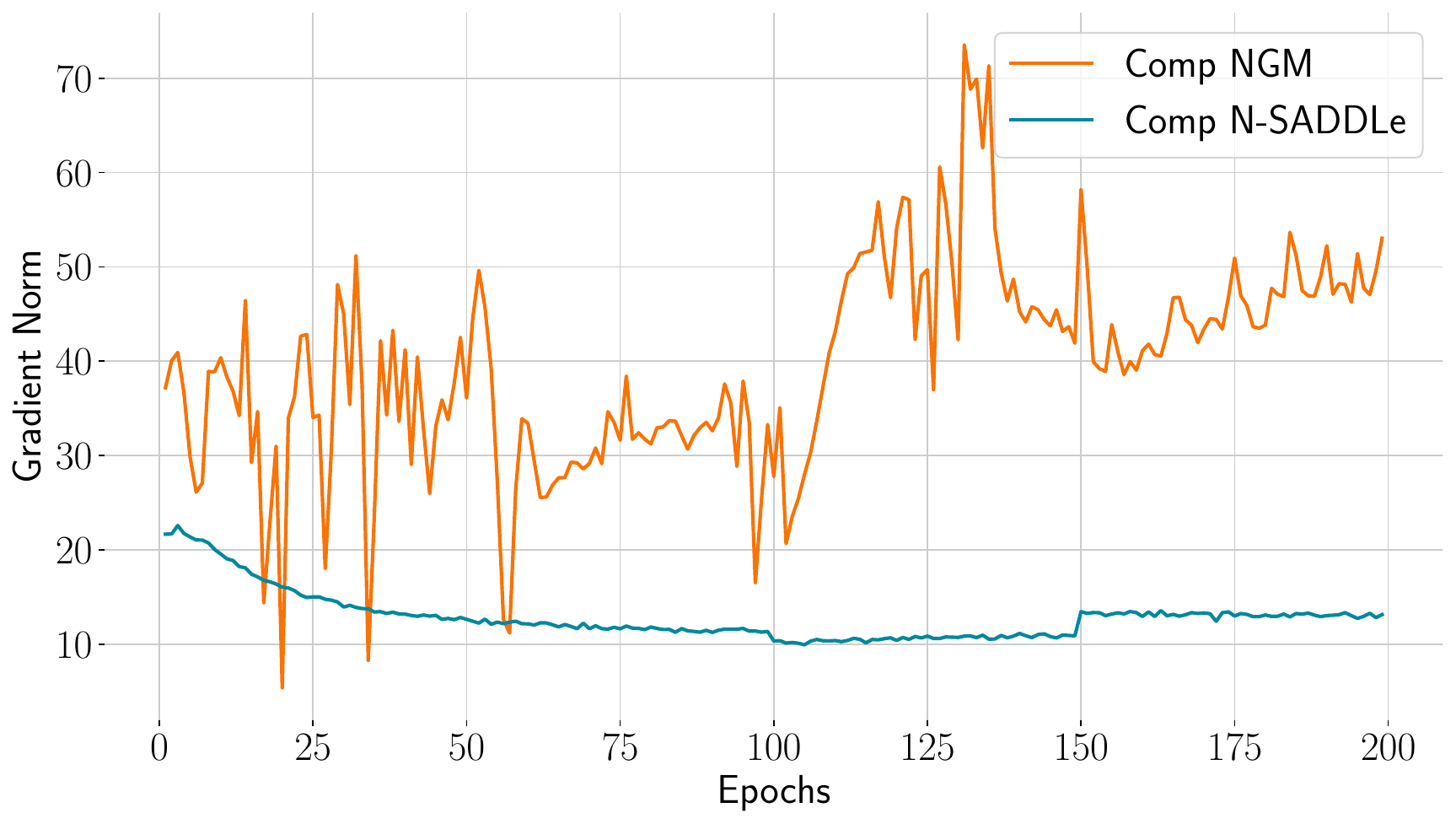}
		\caption{Gradient Norm}
	\end{subfigure}
 % \vspace{-2mm}
	\caption{Impact of flatness on (a) Compression Error and (b) Gradient Norm for ResNet-20 trained on CIFAR-10 distributed in a non-IID manner across a 10 agent ring topology.}
	\label{fig:comperror_ngm}
 % \vspace{-2mm}
\end{figure*}

For NGM and N-SADDLe, $\theta$ corresponds to the gradients $\mathbf{g}_i$ and $\widetilde{\mathbf{g}}_i$ respectively.  In Figure \ref{fig:comperror_ngm}, we compare the compression error ($\|Q(\mathbf{\theta})- \theta\|$) and gradient norms for NGM and N-SADDLe with a 1-bit Sign SGD-based compression scheme. Clearly, N-SADDLe leads to a lower compression error, as well as lower gradient norms throughout the training. Here, we plot the sum of layer-wise compression errors and the sum of gradient norms for each layer in the ResNet-20 model. Like Q-SADDLe, the bound in Equation \ref{apx:comperrorbound} is tighter for N-SADDLe than NGM. 

% \begin{table}[ht]
% \caption{Test accuracy of different decentralized algorithms on CIFAR-10, distributed with $\alpha=0.001$ over torus topology.}
% \vspace{-3mm}
% \label{tab:torus}
% \small
% \begin{center}
% \begin{tabular}{cccc}
% \hline
% Agents & Comp & Method & Accuracy(\%)\\ 
%  \hline
% \multirow{8}{*}{$20$} & & QGM & 56.07 $\pm$ 3.88 \\
% & \checkmark & QGM & 47.62 $\pm$ 6.75 \\
% & & \textit{Q-SADDLe (ours)} & \textbf{77.10 $\pm$ 1.18} \\
% & \checkmark & \textit{Q-SADDLe (ours)} & \textbf{74.50 $\pm$ 1.31} \\
% \cline{2-4}
% & & NGM & 88.18 $\pm$ 0.17 \\
% & \checkmark & NGM & 87.71 $\pm$ 0.54 \\
% & & \textit{N-SADDLe (ours)} & \textbf{89.21 $\pm$ 0.26} \\
% & \checkmark & \textit{N-SADDLe (ours)} & \textbf{89.30 $\pm$ 0.36} \\
%  \hline

% \multirow{8}{*}{$40$} & & QGM & 57.96 $\pm$ 3.90 \\
% & \checkmark & QGM & 47.08 $\pm$ 7.72 \\
% & & \textit{Q-SADDLe (ours)} & \textbf{70.05 $\pm$ 3.35} \\
% & \checkmark & \textit{Q-SADDLe (ours)} & \textbf{64.84 $\pm$ 2.46} \\
% \cline{2-4}
% & & NGM & 86.00 $\pm$ 0.34 \\
% & \checkmark & NGM & 86.30 $\pm$ 0.52 \\
% & & \textit{N-SADDLe (ours)} & \textbf{86.67 $\pm$ 0.32} \\
% & \checkmark & \textit{N-SADDLe (ours)} & \textbf{87.00 $\pm$ 0.18} \\
% \hline
%  %%%%%%%%%%%%%%%%%%%
% \end{tabular}
% \end{center}
% \end{table}

\begin{figure*}[htbp!]
\centering
	\begin{subfigure}{0.37\linewidth}
		\includegraphics[width=\textwidth]{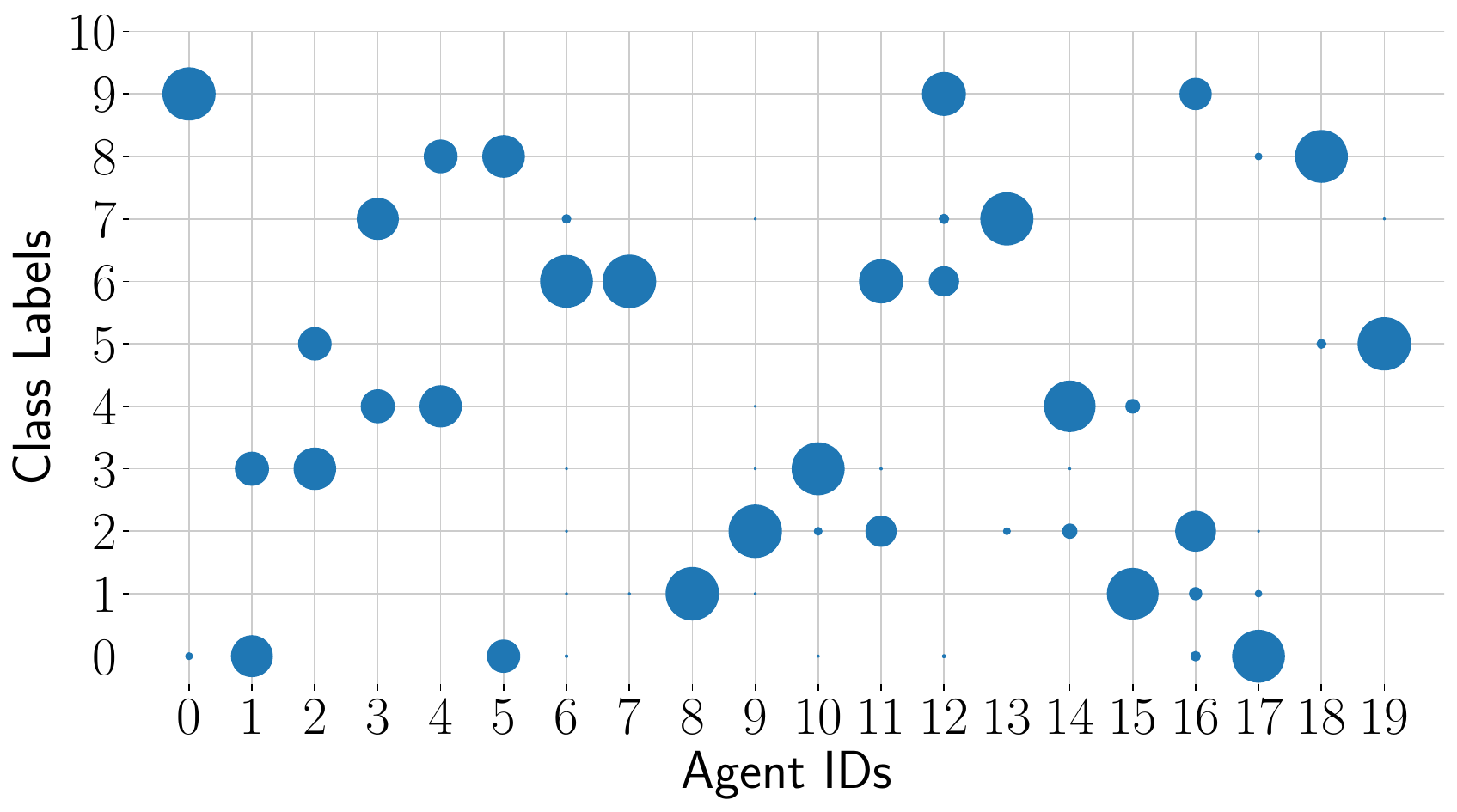}
		\caption{$\alpha=0.01$}
	\end{subfigure}
 \bigskip
	\begin{subfigure}{0.37\linewidth}
		\includegraphics[width=\textwidth]{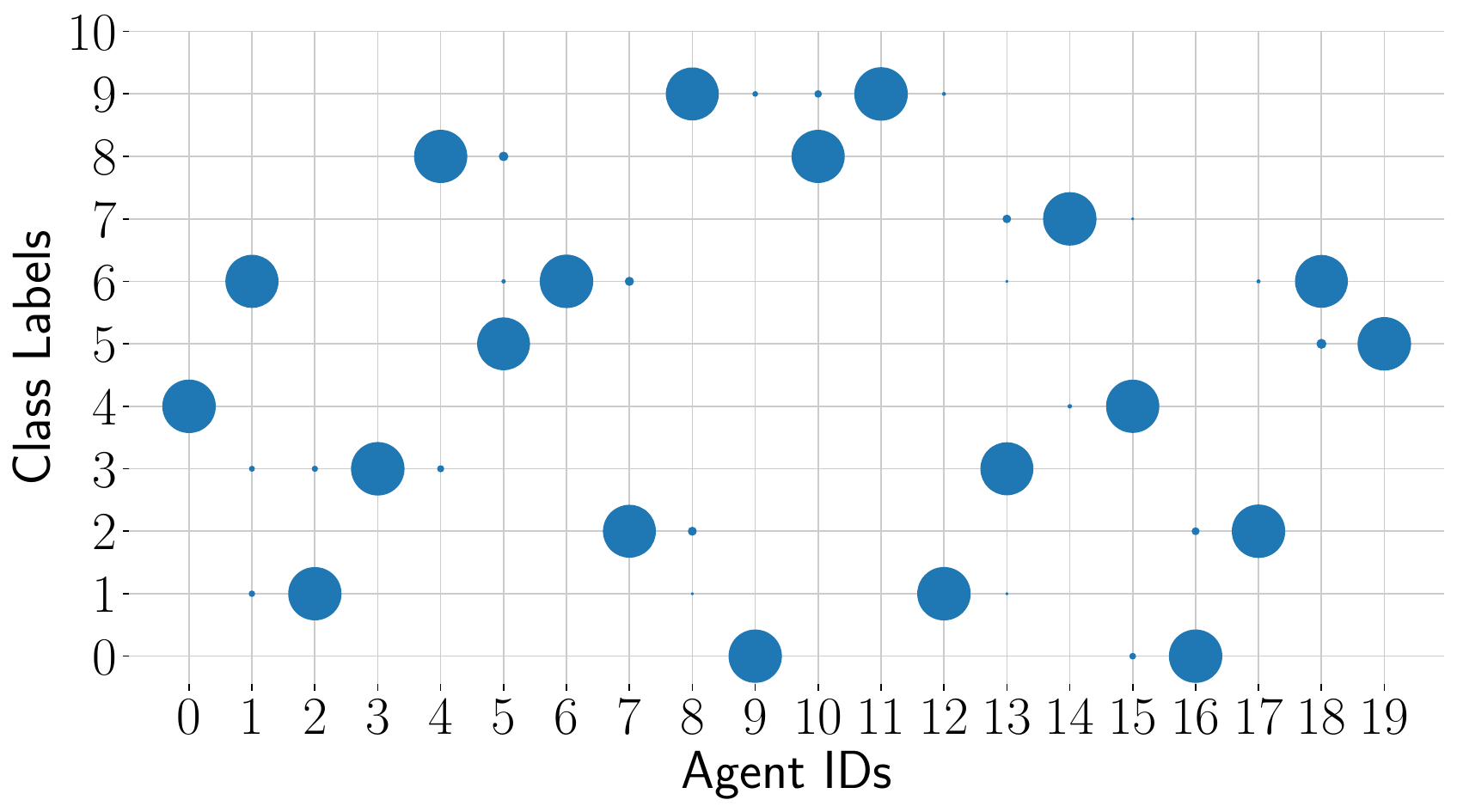}
		\caption{$\alpha=0.001$}
	\end{subfigure}
 % \vspace{-2mm}
	\caption{Visualization of the number of samples from each class allocated to each agent for different Dirichlet distribution $\alpha$ values on the CIFAR-10 dataset.}
	\label{fig:alphavisual}
 % \vspace{-1mm}
\end{figure*}

\begin{table*}[htbp!]
\caption{Test accuracy (Acc) and accuracy drop (Drop) of NGM and N-SADDLe with 2-bit quantization compression scheme \cite{qsgd} evaluated on CIFAR-10, with $\alpha=0.01, 0.001$.}
\vspace{-3mm}
\label{tab:ngm_2bitcf10_100}
\small
\begin{center}
%\resizebox{1.0\columnwidth}{!}{
\begin{tabular*} {\textwidth}{ccl @{\extracolsep{\fill}}*{4}{c}}
\hline
\multirow{ 3}{*}{Agents} &\multirow{3}{*}{Comp} & \multirow{ 3}{*}{Method}& \multicolumn{4}{c}{CIFAR-10 (ResNet-20)}\\
\cline{4-7}  
& & & \multicolumn{2}{c}{$\alpha=0.01$} & \multicolumn{2}{c}{$\alpha=0.001$}\\
\cline{4-5}
\cline{6-7}
& & & Acc (\%) & Drop(\%) & Acc (\%) & Drop(\%)\\
 \hline
%\multirow{5}{*}{$5$} & DSGDm (IID) & \multicolumn{2}{c}{-}  \\
 % \multirow{4}{*}{$5$} & NGM &  90.87 $\pm$ 0.39 & 90.73 $\pm$ 0.46 & 59.00 $\pm$ 4.26 & 54.78 $\pm$ 4.68\\
 \multirow{2}{*}{$5$} & \checkmark & NGM & 87.38 $\pm$ 2.01 & 3.49 & 87.27 $\pm$ 0.56 & 3.46\\
 % & \textit {NGM-SAM (ours)} & 91.96 $\pm$ 0.19 & 91.69 $\pm$ 0.15 & 63.87 $\pm$ 0.45 & 64.10 $\pm$ 0.48\\
 & \checkmark & \textit{N-SADDLe (ours)} & \textbf{91.35 $\pm$ 0.17} & 0.61 & \textbf{91.18 $\pm$ 0.25} & 0.51 \\
 \hline
%\multirow{5}{*}{$10$} & DSGDm (IID) & \multicolumn{1}{c}{-}  \\
 % \multirow{4}{*}{$10$} & NGM &  85.08 $\pm$ 2.73 & 83.43 $\pm$ 0.95 & 55.2 $\pm$ 1.41 & 54.70 $\pm$ 1.36\\
 \multirow{2}{*}{$10$} & \checkmark & NGM & 79.89 $\pm$ 8.74 & 5.19 & 79.20 $\pm$ 3.05 & 4.23\\
 % & \textit {NGM-SAM (ours)} & 90.03 & 87.29 $\pm$ 1.23 & 59.31 $\pm$ 0.61 & 58.37 $\pm$ 0.30 \\
 & \checkmark & \textit{N-SADDLe (ours)} & \textbf{87.25 $\pm$ 1.65} & 1.18 & \textbf{85.70 $\pm$ 1.15} & 1.59\\
  \hline
%\multirow{5}{*}{$15$} & DSGDm (IID) & \multicolumn{2}{c}{-}  \\
 % \multirow{4}{*}{$20$} & NGM &  84.84 $\pm$ 0.43 & 83.58 $\pm$ 0.89 & 53.98 $\pm$ 0.31 & 53.37 $\pm$ 0.53 \\
 \multirow{2}{*}{$20$} & \checkmark & NGM & 81.87 $\pm$ 1.17 & 2.97 & 76.68 $\pm$ 0.95 & 6.90\\
 & \checkmark & \textit {N-SADDLe (ours)} & \textbf{84.25 $\pm$ 0.17} & 2.01 & \textbf{85.09 $\pm$ 0.31} & 1.52\\
 % & \textit{Comp NGM-SAM (ours)} & 86.34 $\pm$ 0.24 & 87.41 $\pm$ 0.52 & 56.65 $\pm$ 0.17 & 54.39 \\
  \hline
 %%%%%%%%%%%%%%%%%%%
\end{tabular*}
%}
\end{center}
\vspace{-4mm}
\end{table*}

\subsection{Stochastic Quantization for NGM and N-SADDLe}
The main paper uses Sign SGD \cite{efsgd} compression scheme with NGM and N-SADDLe since it has been shown to perform better than stochastic quantization for extreme compression \cite{choco-sgd, efsgd}. However, to demonstrate the generalizability of our approach, we present results on 2-bit stochastic quantization in Table \ref{tab:ngm_2bitcf10_100}. NGM incurs an average drop of 4.4\%, while N-SADDLe incurs only a 1.2\% average accuracy drop in the presence of this compression scheme.

\subsection{Loss Landscape Visualization}
\begin{figure}[h!]
\centering
	\begin{subfigure}{0.7\linewidth}
		\includegraphics[width=\textwidth]{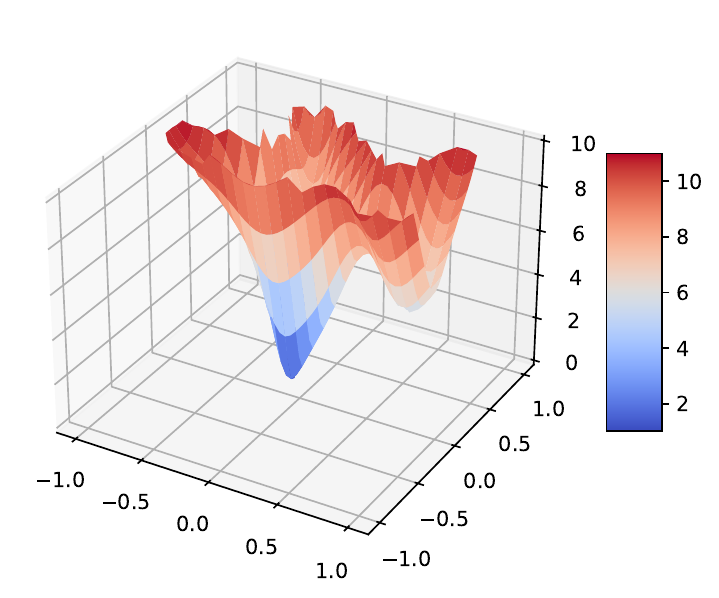}
		\caption{QGM}
	\end{subfigure}
	\begin{subfigure}{0.7\linewidth}
		\includegraphics[width=\textwidth]{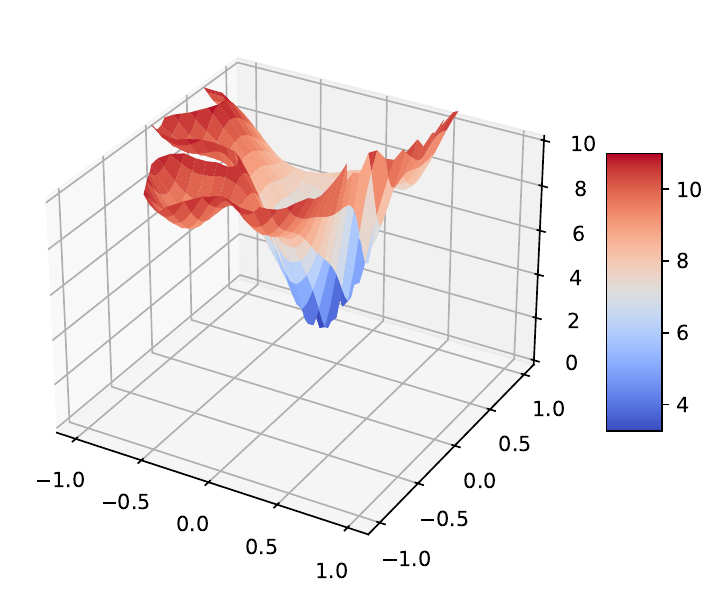}
		\caption{Comp QGM}
	\end{subfigure}
 	\begin{subfigure}{0.7\linewidth}
		\includegraphics[width=\textwidth]{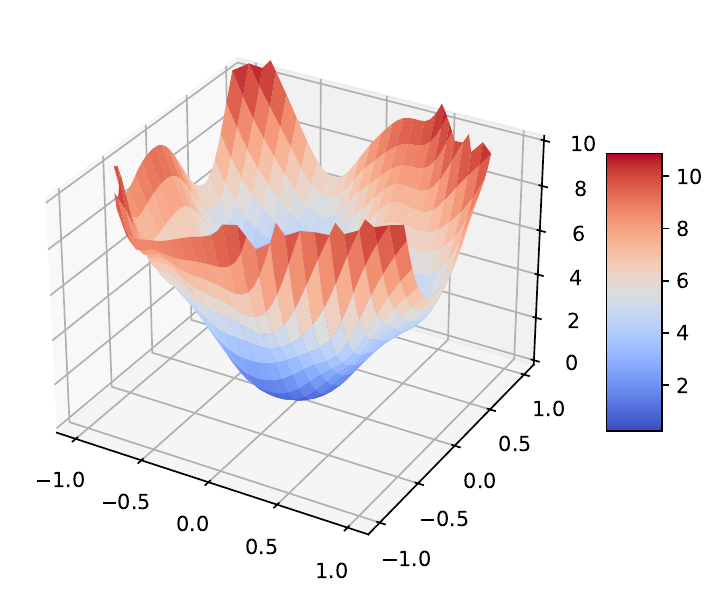}
		\caption{Q-SADDLe}
	\end{subfigure}
	\begin{subfigure}{0.7\linewidth}
		\includegraphics[width=\textwidth]{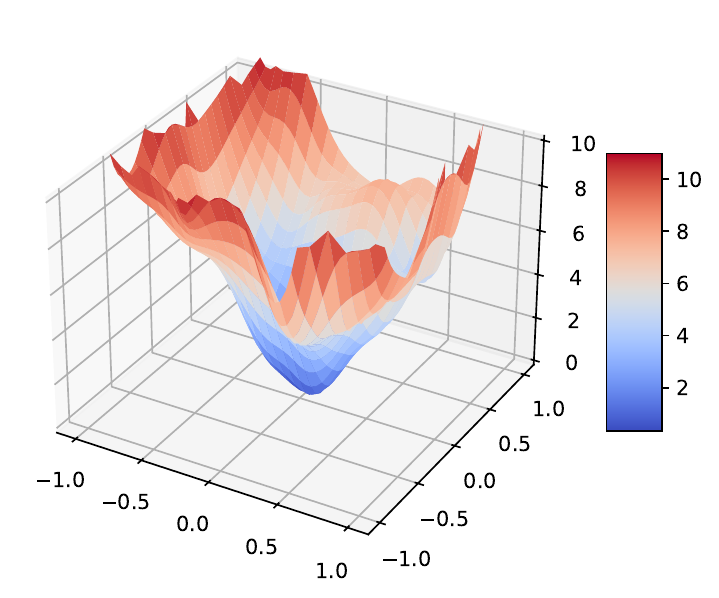}
		\caption{Comp Q-SADDLe}
	\end{subfigure}
	\caption{Visualization of the loss landscape for ResNet-20 trained on the CIFAR-10 dataset distributed across a 10 agent ring topology with $\alpha=0.001$.}
	\label{fig:apexlossvisualqgm}
\end{figure}

\begin{figure*}[h!]
\centering
	\begin{subfigure}{0.37\linewidth}
		\includegraphics[width=\textwidth]{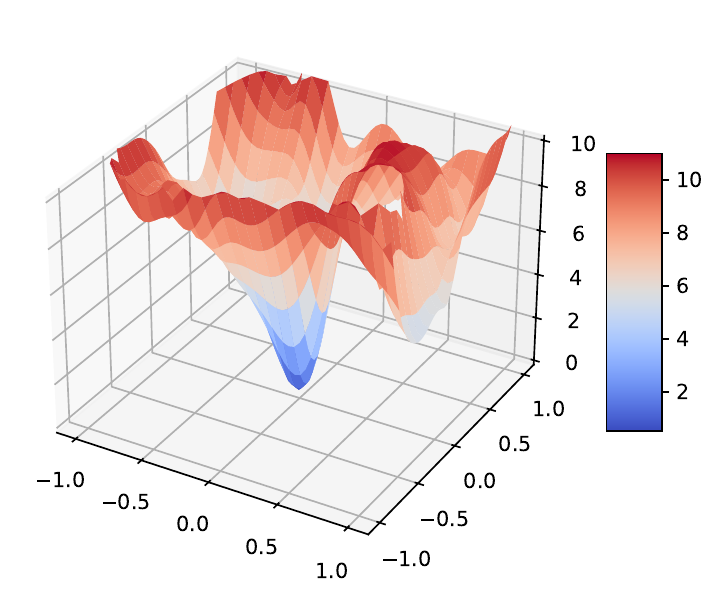}
		\caption{NGM}
	\end{subfigure}
	\begin{subfigure}{0.37\linewidth}
		\includegraphics[width=\textwidth]{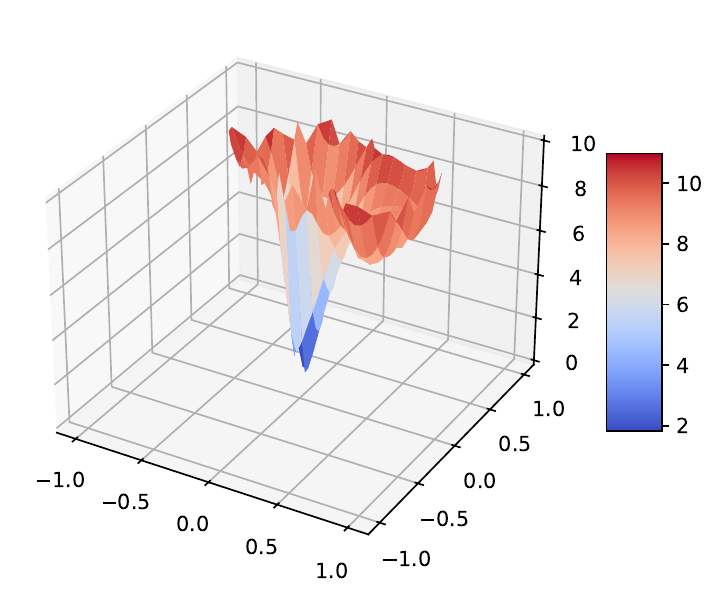}
		\caption{Comp NGM}
	\end{subfigure}
 	\begin{subfigure}{0.37\linewidth}
		\includegraphics[width=\textwidth]{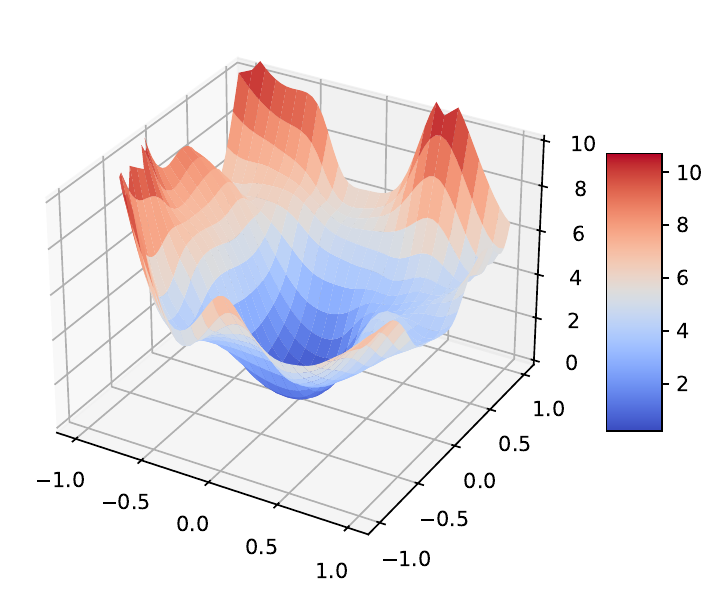}
		\caption{N-SADDLe}
	\end{subfigure}
	\begin{subfigure}{0.37\linewidth}
		\includegraphics[width=\textwidth]{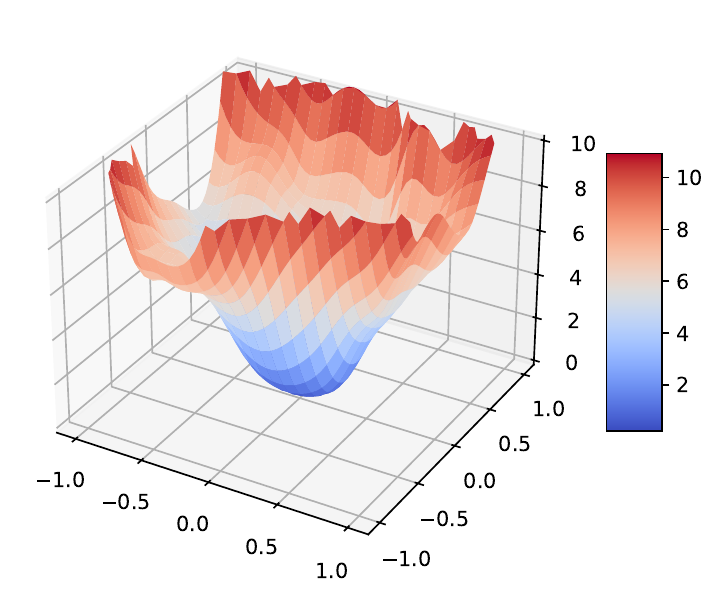}
		\caption{Comp N-SADDLe}
	\end{subfigure}
	\caption{Visualization of the loss landscape for ResNet-20 trained on the CIFAR-10 dataset distributed across a 10 agent ring topology with $\alpha=0.001$.}
	\label{fig:apexlossvisualngm}
\end{figure*}

To visualize the loss landscape, we randomly sample two directions through orthogonal Gaussian perturbations \cite{visualizeloss} and plot the loss for ResNet-20 trained with CIFAR-10 distributed across 10 nodes with $\alpha=0.001$. As shown in Figure \ref{fig:apexlossvisualqgm}, we observe that Q-SADDLe and Comp Q-SADDLe have much smoother loss landscapes than QGM and Comp QGM. The compressed counterparts of QGM and Q-SADDLe are relatively sharper than their respective full communication versions. This is intuitively expected since communication compression leads agents to receive less information from their neighbors, resulting in more reliance on local updates. This can exacerbate over-fitting in the presence of data heterogeneity. We observe similar trends for NGM, N-SADDLe, and their compressed versions as shown in Figure \ref{fig:apexlossvisualngm}.

\subsection{Communication Cost}\label{commcost}
This section presents the exact amount of data transmitted (in Gigabytes) during training (Tables \ref{tab:qgmcf10_100_comm}-\ref{tab:ngmimnet_comm}).

\begin{table}[ht!]
\vspace{-2.5mm}
\caption{Communication costs per agent (in GBs) for experiments in Table 1 (main paper) for QGM and \textit{Q-SADDLe} with a stochastic quantization-based compression scheme with 8 bits, leading to a 4$\times$ reduction in communication cost.}
\vspace{-3mm}
\label{tab:qgmcf10_100_comm}
\small
\begin{center}
\begin{tabular}{cccc}
\hline
Agents & Comp & CIFAR-10 & CIFAR-100\\
 \hline
 \multirow{2}{*}{$5$} & & 136.45 & 111.32 \\
 & \checkmark & 34.11 & 27.83 \\
 \hline
 \multirow{2}{*}{$10$} & & 68.44 & 55.66 \\
 & \checkmark & 17.11 & 13.91 \\
  \hline
 \multirow{2}{*}{$20$} & & 34.43 & 27.83 \\
 & \checkmark & 8.60 & 6.95 \\
  \hline
 \multirow{2}{*}{$40$} & & 17.43 & 14.02 \\
 & \checkmark & 4.35 & 3.50\\
 %%%%%%%%%%%%%%%%%%%
 \hline
\end{tabular}
\end{center}
\vspace{-3mm}
\end{table}

\begin{table}[ht!]
\vspace{-2.5mm}
\caption{Communication costs per agent (in GBs) for experiments in Table \ref{tab:qgmcf10_100_topk} for QGM and \textit{Q-SADDLe} with a top-30\% compression scheme, leading to a 2.2$\times$ reduction in communication cost.}
\vspace{-3mm}
\label{tab:qgmcf10_100_topk_comm}
\small
\begin{center}
\begin{tabular} {ccc}
\hline
Agents & Comp & CIFAR-10\\
 \hline
 $5$ & \checkmark & 61.38 \\
 \hline
 $10$ & \checkmark & 30.78 \\
  \hline
 $20$ & \checkmark & 15.49 \\
 %%%%%%%%%%%%%%%%%%%
 \hline
\end{tabular}
\end{center}
\vspace{-3mm}
\end{table}

\begin{table}[ht!]
\vspace{-2.5mm}
\caption{Communication costs per agent (in GBs) for experiments in Table 2 (main paper) for  QGM and \textit{Q-SADDLe} with a stochastic quantization-based compression scheme with 10 bits , leading to a 3.2$\times$ reduction in communication cost.}
\vspace{-3mm}
\label{tab:qgmimnette_comm}
\small
\begin{center}
\begin{tabular}{ccc}
\hline
Agents & Comp & Imagenette\\
 \hline
 \multirow{2}{*}{$5$} & & 110.23\\
 & \checkmark & 34.44 \\
 \hline
 \multirow{2}{*}{$10$} & & 55.10\\
 & \checkmark & 17.21 \\
 %%%%%%%%%%%%%%%%%%%
 \hline
\end{tabular}
\end{center}
\vspace{-3mm}
\end{table}

\begin{table}[ht!]
\vspace{-2.5mm}
\caption{Communication costs per agent (in GBs) for experiments in Table 3 (main paper) for NGM and \textit{N-SADDLe} with 1-bit Sign SGD, leading to a 32$\times$ reduction in the cost for the second round and a total of 1.94$\times$ reduction in the entire communication cost.}
\vspace{-3mm}
\label{tab:ngmcf10_100_comm}
\small
\begin{center}
\begin{tabular}{cccc}
\hline
Agents & Comp & CIFAR-10 & CIFAR-100\\
 \hline
 \multirow{2}{*}{$5$} & & 272.91 & 222.65 \\
 & \checkmark & 140.67 & 114.76 \\
 \hline
 \multirow{2}{*}{$10$} & & 136.89 & 111.32 \\
 & \checkmark & 70.56 & 57.38 \\
  \hline
 \multirow{2}{*}{$20$} & & 68.88 & 55.66 \\
 & \checkmark & 35.50 & 28.69 \\
 %%%%%%%%%%%%%%%%%%%
 \hline
\end{tabular}
\end{center}
\vspace{-3mm}
\end{table}

\begin{table}[ht!]
% \vspace{-2.5mm}
\caption{Communication costs per agent (in GBs) for experiments in Table 4 (main paper) for NGM and \textit{N-SADDLe} with 1-bit Sign SGD, leading to a 32$\times$ reduction in the cost for the second round and a total of 1.94$\times$ reduction in the entire communication cost.}
\vspace{-3mm}
\label{tab:ngmimnet_comm}
\small
\begin{center}
\begin{tabular}{cccc}
\hline
Agents & Comp & Imagenette & ImageNet\\
 \hline
 \multirow{2}{*}{$10$} & & 110.25 & 22466.30 \\
 & \checkmark & 56.82 & 11580.56 \\
  \hline
 %%%%%%%%%%%%%%%%%%%
\end{tabular}
\end{center}
\vspace{-3mm}
\end{table}
\section{Decentralized Learning Setup}\label{setupdetails}
All our experiments were conducted on a system with 4 NVIDIA A40 GPUs, each with 48GB GDDR6. We report the test accuracy of the consensus model averaged over three randomly chosen seeds.

\subsection{Visualization of Non-IID Data}
Figure \ref{fig:alphavisual} illustrates the number of samples from each class allocated to each agent for the 2 different Dirichlet distribution $\alpha$ values used in our work. $\alpha=0.001$ corresponds to the most extreme form of data heterogeneity, i.e. samples from only 1 class per agent. Note that this level of non-IIDness has been used in CGA \cite{cga} and NGM \cite{ngm} to evaluate the performance. $\alpha=0.01$ has been used in QGM \cite{qgm} and is relatively mild, with most agents accessing samples from 2 different classes (some even from 4 classes).

% \subsection{Compression Operators} \label{apx:compoperators}
% \textbf{Stochastic Quantization:}

% \textbf{1-bit Sign SGD}:

% \textbf{Top-k Sparsification:}
\subsection{Hyper-parameters} \label{apx:hyperparams}
This section presents the hyper-parameters for results presented in Section 6 (main paper) and Section \ref{addresults}. All our experiments were run for three randomly chosen seeds. We decay the learning rate by 10$\times$ after 50\% and 75\% of the training for all experiments except for ImageNet results in Table 4 and Figure 2. For ImageNet, we decay the learning rate by 10$\times$ after 33\%, 67\%, and 90\% of the training. For Figure 2, we use the StepLR scheduler, where the learning rate decays by 0.981 after every epoch. We use a Nesterov momentum of 0.9 for all our experiments, and keep  $\mu=\beta$, similar to QGM \cite{qgm}. We also use a weight decay of 1e-4 for all the presented experiments. Please refer to Table \ref{tab:hyper} for the learning rate, perturbation radius, number of epochs, and batch size per agent for all the experiments in this paper. For a fair comparison, we ensure that all the techniques utilize the same set of hyper-parameters.

We tune the global averaging rate $\gamma$ through a grid search over $\gamma=\{0.01,0.1,0.2,...,1.0\}$ and present the fine-tuned $\gamma$ used for experiments in Tables 3, 4 from the main paper and Table \ref{tab:ngm_2bitcf10_100} in Table \ref{tab:hyper_gamma}. For results in Tables 1, 2 (main paper), and \ref{tab:dsamcf10_100}, we use $\gamma=1.0$ for all the experiments. For Top-30\% Sparsification results shown in Table \ref{tab:qgmcf10_100_topk}, we use $\gamma=0.4$. For our experiments on torus topology in Table 5 (main paper), we use an averaging rate of $0.5$.

% \textbf{Hyper-parameters for CIFAR-10:} All the experiments for CIFAR-10 (trained on ResNet-20) have a stopping criteria for 200 epochs, with step size decayed by 10$\times$ in multiple steps at $100^{th}$ and $150^{th}$ epoch. We use a mini-batch size of 32 per agent.

% \textbf{Hyper-parameters for CIFAR-100:} All the experiments for CIFAR-100 (trained on ResNet-20) have a stopping criteria for 100 epochs, with step size decayed by 10$\times$ in multiple steps at $50^{th}$ and $75^{th}$ epoch. We use a mini-batch size of 20 per agent.

% \textbf{Hyper-parameters for Imagenette:} All the experiments for Imagenette (trained on MobileNet-V2) have stopping criteria for 100 epochs, with step size decayed by 10$\times$ in multiple steps at $50^{th}$ and $75^{th}$ epoch. We use a mini-batch size of 32 per agent.

% \textbf{Hyper-parameters for ImageNet:}

\begin{table*}[htbp!]
\vspace{-2.5mm}
\caption{Learning rate ($\eta$), the perturbation radius ($\rho$) (where applicable), batch size per agent, and the number of epochs for all the experiments for QGM, Q-SADDLe, NGM, N-SADDLe, and their compressed versions across various datasets.}
\vspace{-3mm}
\label{tab:hyper}
\small
\begin{center}
\begin{tabular}{ccccc}
\hline
Dataset & CIFAR-10 & CIFAR-100 & Imagenette & ImageNet\\
\hline
\hline
Learning Rate ($\eta$) & 0.1 & 0.1 & 0.01 & 0.01 \\
\hline
Perturbation Radius ($\rho$) & 0.1 & 0.05 & 0.01 & 0.05 \\
\hline
Epochs & 200 & 100 & 100 & 60 \\
\hline
Batch-Size/Agent & 32 & 20 & 32 & 64 \\
 \hline
\end{tabular}
\end{center}
\vspace{-3mm}
\end{table*}

\begin{table*}[h!]
\caption{Global averaging rate ($\gamma$) for our experiments in Table 3, 4 (main paper) and \ref{tab:ngm_2bitcf10_100}.}
\vspace{-4mm}
\label{tab:hyper_gamma}
\small
\begin{center}
\begin{tabular}{cccccc}
\hline
Method & Non-IID Level ($\alpha$) & CIFAR-10 & CIFAR-100 & Imagenette & ImageNet\\
\hline
\hline
\multirow{2}{*}{NGM} & 0.01 & 1.0 & 1.0 & 0.5 & 1.0\\
& 0.001 & 1.0 & 1.0 & 0.5 & 1.0\\
\hline
\multirow{2}{*}{Comp NGM} & 0.01 & 0.5 & 0.5 & 0.1 & 0.5\\
& 0.001 & 0.5 & 0.5 & 0.5 & 0.5\\
\hline
\multirow{2}{*}{N-SADDLe} & 0.01 & 1.0 & 1.0 & 0.5 & 1.0\\
& 0.001 & 1.0 & 1.0 & 0.5 & 1.0\\
\hline
\multirow{2}{*}{Comp N-SADDLe} & 0.01 & 0.5 & 0.5 & 0.1 & 1.0\\
& 0.001 & 0.5 & 0.5 & 0.5 & 1.0\\
 \hline
\end{tabular}
\end{center}
% \vspace{-3mm}
\end{table*}

% \clearpage
% {\small
% \bibliographystyle{ieee_fullname}
% \bibliography{egbib}
% }
% \end{document}
\end{document}